\documentclass[unnumsec,webpdf,contemporary,large]{oup-authoring-template}%
\usepackage{float}
\usepackage{tabularx}
\usepackage{booktabs}
\usepackage{array}
\usepackage{xcolor}
\makeatletter
\AtBeginDocument{\setcitestyle{square}}%
\makeatother

\theoremstyle{thmstyleone}%
\theoremstyle{thmstyletwo}%
\theoremstyle{thmstylethree}%

\newcommand{\cdash}{---}

\makeatletter
\long\def\abstract#1{\let\subheading\abstsubheading\global\long\def\@abstract{#1}}
\makeatother

\begin{document}

\journaltitle{Briefings in Bioinformatics}
\DOI{}
\copyrightyear{2026}
\pubyear{2026}
\vol{}
\issue{}
\access{}
\appnotes{}

\firstpage{1}

%\subtitle{Subject Section}

\title[HEDGEHOG molecular-generator benchmark]{HEDGEHOG: Hierarchical Evaluation of Drug Generators Through Rigorous Filtration}

\author[1,2,$\ast$]{Daria A. Ryabchenko\ORCID{0009-0000-3304-418X}}
\author[1,2]{Pavel Gurevich\ORCID{0009-0001-2810-8123}}
\author[1]{Shamil Kadyrov\ORCID{0009-0009-9455-6125}}
\author[1,2]{Daria Frolova\ORCID{0000-0001-8368-9015}}
\author[1]{Kseniia Fedisheva\ORCID{0009-0002-1164-5253}}
\author[1]{Sergei A. Nikolenko\ORCID{0000-0003-1150-9390}}
\author[1,2]{Alexander Shapeev\ORCID{0000-0002-7497-5594}}
\author[1]{Marina A. Pak\ORCID{0000-0003-1075-6509}}

\address[1]{\orgname{Ligand Pro}, \orgaddress{\state{Moscow}, \country{Russia}}}
\address[2]{\orgname{Applied AI Institute}, \orgdiv{Artificial Intelligence Center}, \orgaddress{\state{Moscow}, \country{Russia}}}

\corresp[$\ast$]{Corresponding author. \href{podariya12@gmail.com}{podariya12@gmail.com}}

% arXiv: omit production date placeholders
%\received{Date}{0}{Year}
%\revised{Date}{0}{Year}
%\accepted{Date}{0}{Year}

\abstract{%
\noindent\hspace*{1.4em}%
\begin{minipage}{\dimexpr\linewidth-1.4em\relax}%
\noindent
\begin{minipage}[t]{0.50\linewidth}%
Generative molecular models can support early drug discovery by proposing new candidate compounds \textit{de novo}. In practice, useful candidates must balance target-relevant activity, physicochemical properties, and other multiparameter design constraints. However, standard metrics commonly used to evaluate molecular generators only weakly reflect whether the generated compounds are medicinally plausible and suitable for downstream computation. This can produce an incomplete view of model performance and inefficient use of computational resources. We introduce \textsc{HEDGEHOG}, a unified \textit{six-stage} filtration benchmark that is constructed as a hit identification workflow: (i)~preprocessing; (ii)~physicochemical descriptor screening; (iii)~structural alerts and graph-sanity checks; (iv)~synthesis feasibility; (v)~docking; and (vi)~three-dimensional pose and interaction checks. We evaluated $22$ generative models in a KRAS~G12D case study, using three runs of 1,000 requested generation attempts per model. The models showed different patterns of attrition, and final survival ranged from $0$ to $127.0 \pm 18.1$ molecules per run. None of the standard metrics showed a significant association with final survival after correction for multiple testing. \textsc{HEDGEHOG} provides a reproducible benchmark for evaluating molecular generative models by measuring molecule survival
\end{minipage}\hfill
\begin{minipage}[t]{0.48\linewidth}%
\centering\vspace{0pt}%
\includegraphics[width=\linewidth]{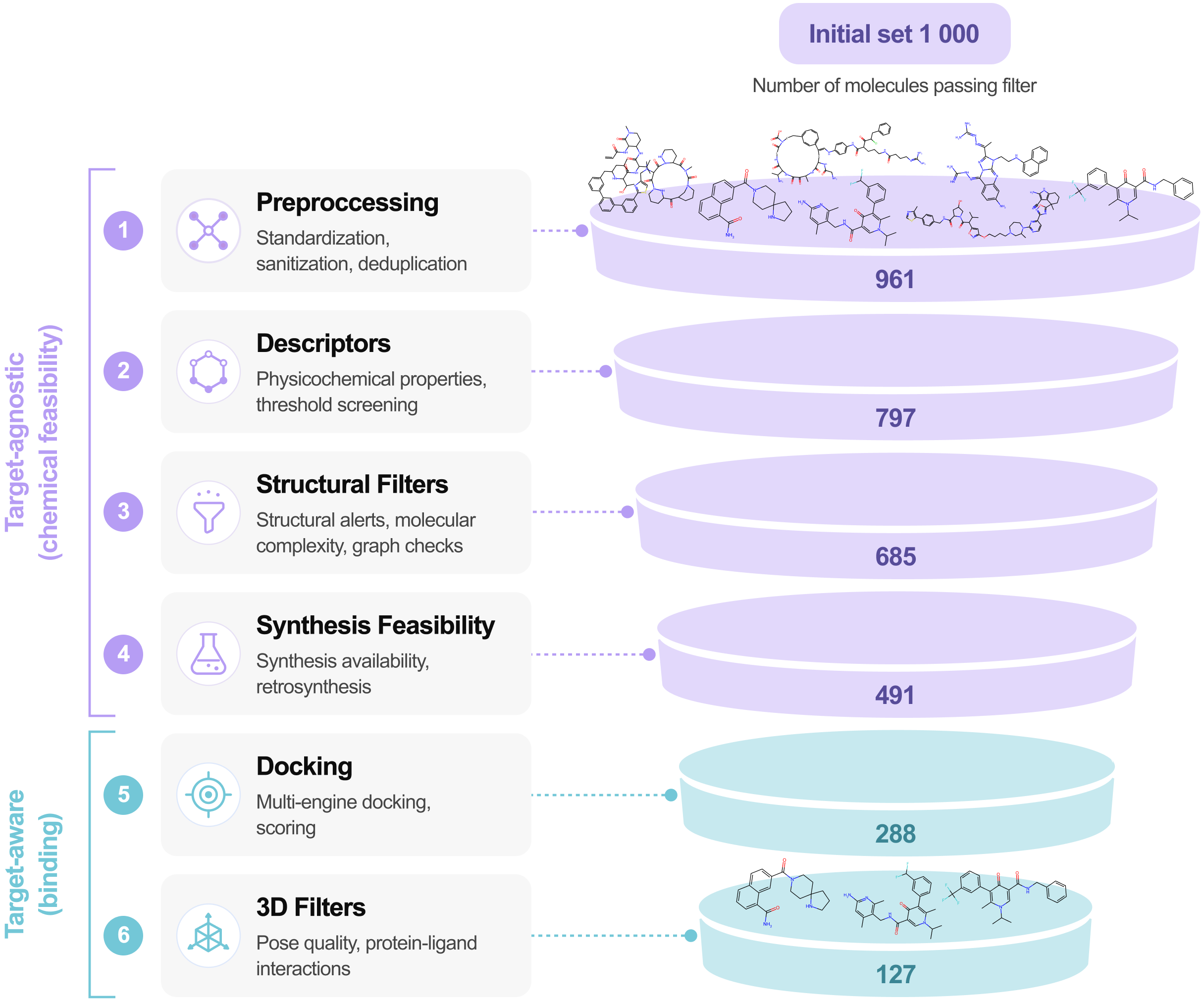}%
\end{minipage}\\[0.3em]
\noindent through chemical filters. The framework identifies stage-wise failure modes across generator classes and provides a practical basis for developing molecular generators better aligned with early drug discovery.%
\end{minipage}\\[0.3em]
}

\makeatletter
\gdef\@keywords{\par\noindent\hspace*{1.4em}{\sffamily\fontseries{b}\selectfont\itshape Keywords}\enspace
molecular generative models, molecular generation benchmarking, \textit{de novo} drug discovery, virtual screening}
\makeatother

% arXiv: suppress first-page journal banner (title/DOI/date placeholders + gray logo box).
\makeatletter
\def\ps@opening{%
  \let\@oddhead\@empty
  \let\@evenhead\@empty
  \let\@oddfoot\@empty
  \let\@evenfoot\@empty
}%
\makeatother

\maketitle

% arXiv: drop journal running header on later pages; keep page numbers only.
\makeatletter
\def\@evenhead{\hbox to \textwidth{\sffamily\fontsize{9bp}{10bp}\selectfont\thepage\hfill}}%
\def\@oddhead{\hbox to \textwidth{\hfill\sffamily\fontsize{9bp}{10bp}\selectfont\thepage}}%
\makeatother

\section{Introduction}
\label{sec1}
Generative models are widely used in molecular design for drug discovery, enabling the generation of large numbers of candidate compounds through efficient exploration and exploitation of chemical space~\citep{tropsha2024integrating}. However, the ability to generate chemically valid molecules alone is insufficient for practical drug development~\citep{ivanenkov2023hitchhiker, du2024machine}. To serve as meaningful starting points for medicinal chemistry, generated molecules must satisfy multiple criteria related to drug-likeness, developability, and biological relevance~\citep{atz2024prospective}.

Standard evaluations of molecular generators emphasize intrinsic metrics such as validity, novelty, uniqueness, and distributional similarity~\citep{brown2019guacamol}. These metrics are useful for assessing whether a model produces chemically plausible and nontrivial outputs. However, they are not sufficient on their own to establish practical value in early-stage drug discovery. For those applications, compounds are prioritized using multiple medicinal chemistry and drug discovery criteria beyond potency alone, including physicochemical properties~\citep{veber2002molecular}, structural liabilities~\citep{bruns2012rules}, synthetic tractability~\citep{du2024machine,swanson2024generative,segler2018planning}, and structure-based screening procedures~\citep{zhou2024artificial,zhao2024science}.

This gap between intrinsic generation quality and practical utility is especially important in hit identification. Early discovery campaigns do not operate on raw strings or abstract graph distributions. Instead, they use molecules that balance multiple properties~\citep{duffy2012early}. Later analysis stages are more time- and resource-intensive and often constrained by limited downstream compute budgets. Therefore, these stages are applied selectively to a prescreened set of molecules rather than exhaustively on the full generated set~\citep{dahlin2014essential,bedart2024emerging}. Molecules with implausible properties, reactive or unstable motifs, poor synthetic tractability, or incompatible binding poses consume resources without improving the quality of the process. As a result, evaluating generators only with broad distributional or structural metrics can substantially overestimate their usefulness for medicinal chemistry.

Several task-specific benchmarks have incorporated docking scores, property objectives, or synthetic accessibility proxies~\citep{brown2019guacamol,garcia2022dockstring,nigam2023tartarus,baillif2024benchmarking}. However, these objectives are often evaluated individually or in simplified combinations, and they rarely reproduce the sequential attrition imposed by practical hit identification workflows. As a result, a model may appear successful by optimizing a property, pharmacophore, or docking proxy while still producing molecules that fail basic medicinal chemistry, retrosynthesis, or pose plausibility checks.

To address these limitations, we developed \textsc{HEDGEHOG} to evaluate generated molecules under a computational triage workflow that more closely reflects hit identification practice. The benchmark applies a fixed sequence of stages covering preprocessing, descriptor screening, structural alerts, synthesis feasibility, docking, and three-dimensional (3D) pose and interaction checks.

The same protocol can be used for unconditional, ligand-based, and protein-based models. Here, we apply it to molecules generated for the KRAS~G12D, a therapeutically important and structurally challenging drug target~\citep{mao2022kras,vasta2022kras}. We report cumulative survival and survival at each stage, and examine whether conventional intrinsic metrics are associated with the final number of molecules retained. This framing makes it possible to separate strong performance on conventional intrinsic metrics from performance under a more demanding computational triage setting. By identifying where generated molecules fail during a realistic triage workflow, \textsc{HEDGEHOG} provides a more informative basis for developing molecular generative models that are better aligned with early drug discovery.

\section{Materials and Methods}\label{sec2}
% ================================================ KNOWN BENCHMARKS COMPARISON ================================================
\subsection{Benchmark systems comparison}
\textbf{Distribution-learning benchmarks.} Existing benchmark suites for molecular generation largely assess two-dimensional (2D) molecular quality and how closely generated sets reproduce reference distributions. GuacaMol~\citep{brown2019guacamol} provides standardized tasks for both distribution learning and goal-directed optimization, reporting metrics such as validity, uniqueness, novelty, KL divergence, and FCD~\citep{preuer2018frechet}. It incorporates compound quality metrics using rule-based filters derived from SureChEMBL, Glaxo~\citep{hann1999strategic}, PAINS~\citep{baell2010new}, and in-house rule sets. MOSES~\citep{polykovskiy2020molecular} standardizes dataset splits and intrinsic distributional metrics. It reports the fraction of molecules passing medicinal chemistry filters such as MCF and PAINS, and evaluates SA score, although this score is a heuristic proxy for synthesis feasibility. These benchmarks do not include comprehensive synthetic feasibility estimation, docking and binding pose assessment, or 3D structure-dependent tests.

\textbf{Structure-based benchmarks.} Recent benchmarks move closer to practical structure-based evaluation. DOCKSTRING~\citep{garcia2022dockstring} packages docking into a reproducible benchmark with a large target panel and tasks such as virtual screening and \textit{de novo} docking score optimization. DOCKSTRING also reports a small set of molecular descriptors, including HBDs, HBAs, rotatable bond count, and logP. GenBench3D~\citep{baillif2024benchmarking} evaluates 3D molecular generators inside binding pockets, introduces geometry-aware Validity3D criteria for conformational plausibility, and reports docking-based scores from Vina~\citep{trott2010autodock}, Glide~\citep{friesner2004glide}, and Gold~PLP~\citep{verdonk2003improved}. It computes a narrow panel of molecular descriptors, including molecular weight, logP, and ring proportion, and uses SA score as a proxy for synthesizability. Durian~\citep{nie2024durian} further expands structure-based 3D evaluation by incorporating protein–ligand complexes with experimental affinity annotations. It uses a broader panel of physicochemical and geometric metrics, and evaluates docking-based affinity with QuickVina2~\citep{alhossary2015fast}, Surflex~\citep{jain2003surflex}, and GNINA~\citep{mcnutt2021gnina}. However, these benchmarks do not natively integrate synthesizability estimation, structural alert filtering, or a representative descriptor panel.

\textbf{Synthesis-aware benchmarks.} Synthesis-aware evaluation has also received increasing attention. SDDBench~\citep{liu2024sddbench} argues that standard synthetic-accessibility scores are insufficient and instead evaluates molecules using a retrosynthesis-based criterion together with search success rate. TARTARUS~\citep{nigam2023tartarus} introduces practical objectives based on physical simulation and explicit computational budgets. Its drug design tasks combine docking objectives with structural constraints and medicinal chemistry filters, while also incorporating metrics such as QED, TPSA, and SA score. These works move evaluation closer to downstream utility, but they are not designed for efficient reduction to actionable chemical space.

\textbf{General-purpose evaluation framework.} Frameworks such as MolScore~\citep{thomas2024molscore} and TDC~\citep{huang2021therapeutics} provide flexible infrastructure for composing objectives, datasets, and evaluation modules. They are valuable for building and testing molecular design workflows, and in some cases contain many of the components that a practical pipeline would use. Our aim is complementary. Rather than proposing another general-purpose scoring framework, we define a highly configurable benchmarking protocol with a fixed staged structure and stage-wise reporting, motivated by hit identification triage. \textsc{HEDGEHOG} evaluates whether generated molecules continue to survive under a sequence of increasingly restrictive and practically motivated filters. Table~\ref{tab:benchmarks-landscape} summarizes the relationship between prior work and \textsc{HEDGEHOG}.

\textbf{HEDGEHOG benchmark protocol.} \textsc{HEDGEHOG} measures molecular survival through sequential stages including molecular standardization and deduplication, $21$ descriptor filters, selected structural liability filters drawn from a database of $2,458$ SMARTS patterns from $23$ rule sets, synthesis scores and retrosynthetic route search with AiZynthFinder~\citep{genheden2020aizynthfinder}, docking with \textit{smina}~\citep{koes2013lessons}, GNINA~\citep{mcnutt2021gnina}, and Matcha~\citep{frolova2025matcha}, and final 3D pose checks. Supplementary~Table~S1 contains software versions for each tool and lists the generic and target-adapted settings.

\begin{table*}[!ht]
  \small
  \caption{Comparison of \textsc{HEDGEHOG} with prior benchmarks and evaluation frameworks for molecular generation. Each column shows whether a benchmark covers one stage of the \textsc{HEDGEHOG} workflow. 
    “Yes” means that the stage is evaluated directly or supported out of the box, 
    “Partial” means that only part of the stage is covered through a proxy metric or narrower components, and 
    “No” means that the stage is not covered. “Code” indicates whether an official implementation or public repository is available}
    \label{tab:benchmarks-landscape}
    \begin{tabular*}{\textwidth}{@{\extracolsep{\fill}}lcccccc}
      \toprule
      \textbf{Benchmark / Framework} & \textbf{Code} & \textbf{Descriptors} & \textbf{Structural Filters} & \textbf{Synthesis Feasibility} & \textbf{Docking} & \textbf{3D Filters} \\
      \midrule
      GuacaMo                     & Yes & No      & Yes & No      & No  & No  \\
      MOSES                       & Yes & Partial & Yes & Partial & No  & No  \\
      DOCKSTRING                  & Yes & Partial & No  & No      & Yes & No  \\
      GenBench3D                  & Yes & Partial & No  & Partial & Yes & Yes \\
      Durian                      & No  & Yes     & No  & Partial & Yes & Yes \\
      SDDBench                    & No  & No      & No  & Yes     & No  & No  \\
      TARTARUS                    & Yes & Partial & Yes & Partial & Yes & No  \\
      MolScore                    & Yes & Yes     & Yes & Yes     & Yes & No  \\
      TDC                         & Yes & No      & Yes & Yes     & Yes & No  \\
      \textbf{\textsc{HEDGEHOG} } & Yes & Yes     & Yes & Yes     & Yes & Yes \\
      \botrule
  \end{tabular*}
\end{table*}

% ================================================ HEDGEHOG PIPELINE DESCRIPTION ================================================
\subsection{HEDGEHOG pipeline}\label{subsec:hedgehog-pipeline}\label{sec:hedgehog_pipeline_description}
\begin{figure*}[!ht]
  \centering
   \includegraphics[width=\linewidth]{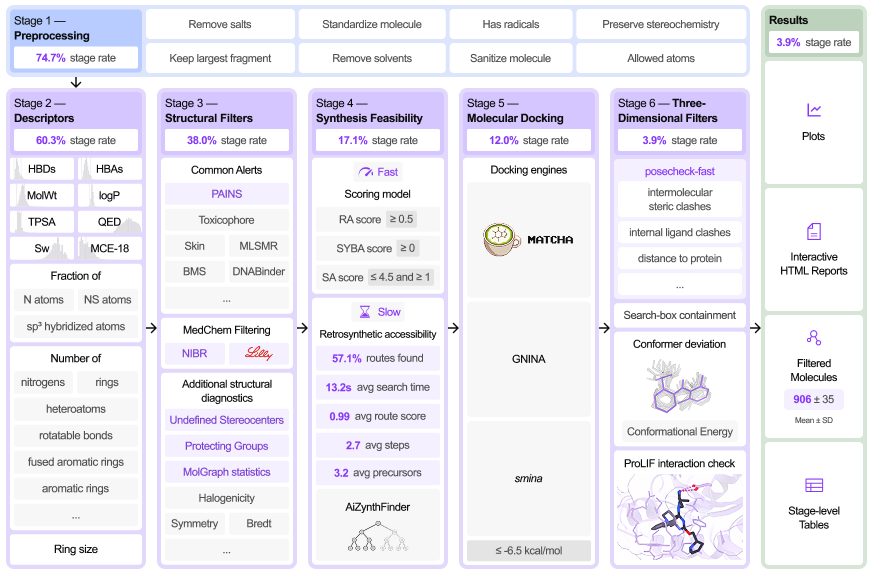}
   \caption{Overview of the HEDGEHOG molecular evaluation workflow. Generated molecules pass sequentially through six configurable stages: (i) molecular preprocessing, (ii) descriptor filtering, (iii) structural filtering, (iv) synthesis feasibility assessment, (v) molecular docking, and (vi) three-dimensional pose filtering. Only molecules satisfying all enabled criteria proceed to the next stage. HEDGEHOG produces diagnostic plots, interactive HTML reports, stage-level tables, and the final set of filtered molecules. Percentages indicate the cumulative fraction of generated molecules retained after each stage, and the final count is reported as the mean $\pm$ sample standard deviation across three seeds.}
   \label{fig:hedgehog-main}
\end{figure*}

The staged design of \textsc{HEDGEHOG} is motivated by computational efficiency: retrosynthetic route finding and molecular docking are computationally more demanding than early descriptor- and structure-based checks. \textsc{HEDGEHOG} therefore uses a coarse-to-fine filtration cascade that removes implausible molecules before passing a smaller candidate set to the downstream stages. The pipeline comprises six stages (Fig.~\ref{fig:hedgehog-main}), namely molecular preprocessing, physicochemical filtering, structural filtering, synthesis feasibility evaluation, docking, and 3D pose analysis. It provides a generic configuration and a mechanism for adapting descriptor and docking criteria to a specific target.

\subsubsection{Stage 1: Molecular preprocessing}\label{subsec:preprocessing}
RDKit~\citep{landrum2013rdkit} and DataMol~\citep{hadrien_mary_2024_10535844} are used to preprocess all generated molecules. The procedure parses each structure, removes salts and solvents, keeps the largest fragment, disconnects metals, and normalizes and sanitizes the molecule while preserving stereochemistry. Molecules with unsupported elements, radicals, invalid valence states, or unresolved multiple fragments are removed. After standardization, duplicate molecules are identified from canonical isomeric SMILES and removed within each model. For 3D generators, native coordinates remain linked to the standardized molecules. Details are provided in the Supplementary~Section~1.1.

\subsubsection{Stage 2: Physicochemical descriptors}\label{subsec:descriptors}
After preprocessing, \textsc{HEDGEHOG} calculates $21$ physicochemical descriptors (Supplementary~Table~S3). This stage filters molecules using configurable threshold ranges defined in the generic configuration (Section~\ref{subsec:sholds}). The panel includes size- and composition-related descriptors, lipophilicity and polarity descriptors, hydrogen bond features, ring statistics, fraction of sp\textsuperscript{3}-hybridized carbon atoms, and additional medicinal chemistry checks.

\subsubsection{Stage 3: Structural filters}\label{subsec:structfilters}
Stage~3 evaluates structural liabilities using a set of $2,458$ explicit SMARTS-based alert rules organized into $23$ Common~Alerts rule sets, two published medicinal chemistry filtering systems, including Lilly~MedChem~Rules~\citep{bruns2012rules} and NIBR~\citep{schuffenhauer2020evolution}, and eight additional structural diagnostic modules. The relevance of individual structural alerts depends on the intended application~\citep{dahlin2014essential}, so a module can be evaluated and reported without being used to reject molecules. The alert layer covers medicinal chemistry and assay-interference substructures such as Dundee~\citep{brenk2008lessons}, BMS~\citep{pearce2006empirical}, Glaxo~\citep{hann1999strategic}, PAINS~\citep{baell2010new}, and related alert sets (Supplementary~Section~1.3). The additional structural layer includes graph-level checks for ring infractions, topological anomalies, protecting groups, halogen-, symmetry-, and stereochemistry-related constraints, Bredt's-rule violations~\citep{fawcett1950bredt}, and molecular complexity metrics.

\subsubsection{Stage 4: Synthesis feasibility}\label{subsec:synthesis}
Stage~4 combines synthesis-related scores with a retrosynthesis search. By default, \textsc{HEDGEHOG} calculates the SA~score~\citep{ertl2009estimation}, RA score~\citep{thakkar2021retrosynthetic}, and SYBA~score~\citep{vorvsilak2020syba}. Molecules satisfying these thresholds are evaluated with AiZynthFinder~\citep{genheden2020aizynthfinder} to test whether at least one synthetic route can be found under the chosen retrosynthesis setup. AiZynthFinder results depend on the template library, stock, search policy, and compute budget. We used a ZINC-derived building-block stock~\citep{sterling2015zinc}, a maximum search depth of $5$, a time limit of $300$ seconds per molecule, and a maximum of $300$ search iterations, terminating after the first solved route. Supplementary~Fig.~S1 shows a route example found by AiZynthFinder.

\subsubsection{Stage 5: Docking}\label{subsec:docking}
Molecules that pass the first four stages are docked with \textit{smina}, GNINA, and Matcha, a recent neural docking method based on multi-stage Riemannian flow matching. These tools were selected based on their strong performance in independent evaluations on diverse docking datasets~\citep{pak2025bento}. The generic cutoff is $-6.5$~kcal/mol for each method. A target-specific cutoff can also be used with desired reference-retention level. In the default configuration, \textsc{HEDGEHOG} uses all three docking tools together. Method-specific cutoffs are then derived from a shared calibration cohort containing $\lceil pN/100\rceil$ reference molecules selected according to their joint performance across the three docking methods, where $p$ is the specified retention percentage. We used $p=95\%$ in this study. Molecules that reach docking have already passed validity, descriptor, structural, and synthesis feasibility checks. These early stages are essential because docking scores alone can be exploited by unrealistic molecules (Supplementary~Fig.~S2). A detailed explanation and the corresponding thresholds are provided in Supplementary~Section~1.5.

\subsubsection{Stage 6: Three-dimensional filtration}\label{subsec:3dfilters}
Stage~6 tests whether each surviving molecule has a geometrically plausible and target-compatible 3D~pose. For SMILES and other 2D generators, Stage~5 supplies a top-ranked docked pose. For 3D generators, the native generated coordinates remain linked to the standardized molecular identity and are evaluated for molecules that pass Stage~5. First, all ligand atoms must remain inside the docking search box. PoseCheck Fast~\citep{nikolenko2026posecheckfast} then evaluates intermolecular clashes, ligand-protein volume overlap and minimum distance, and internal ligand geometry. Conformational plausibility is assessed as the minimum symmetry-corrected heavy atom RMSD between the evaluated pose and an ensemble of $50$ RDKit ETKDGv3 conformers with the default cutoff of 3.0~\AA{}. ProLIF~\citep{bouysset2021prolif} calculates protein-ligand interaction fingerprints and can enforce required or forbidden residue contacts, a minimum number of hydrogen bonds, or similarity to a reference interaction pattern. In the KRAS~G12D benchmark, the target-specific interaction profile requires a ProLIF-detected interaction with Asp12, illustrated in Supplementary~Fig.~S3.

\subsection{Benchmark evaluation setup}\label{subsec:benchmark_evaluation_setup}
We evaluated $22$ generators from unconditional (UN), ligand-based (LB), and protein-based (PB) model classes. Unconditional class included eight models, namely EDM~\citep{hoogeboom2022equivariant}, HierGraphVAE~\citep{jin2020hierarchical}, JT-VAE~\citep{jin2018junction}, MoLeR~\citep{maziarz2021learning}, MolGPT~\citep{bagal2021molgpt}, REINVENT4~\citep{loeffler2024reinvent}, ShEPhERD~\citep{adams2025shepherd}, and TGM-DLM~\citep{gong2024text}. Ligand-based class included six models, namely GCPG~\citep{zou2025structure}, GENTRL~\citep{zhavoronkov2019deep}, MolFinder~\citep{kwon2021molfinder}, PGMG~\citep{zhu2023pharmacophore}, and REINVENT4 prior (P) and REINVENT4 transfer-learned (TL)~\citep{loeffler2024reinvent}. Protein-based class included eight models, namely DiffSBDD~\citep{schneuing2024structure}, Dragonfly and Dragonfly biased (b)~\citep{atz2024prospective}, DrugFlow~\citep{schneuing2025multi}, Pocket2Mol~\citep{peng2022pocket2mol}, ProtoBind-Diff~\citep{mistryukova2025protobind}, ResGen~\citep{zhang2023resgen}, and TargetDiff~\citep{guan20233d}.

Evaluation was under a fixed-budget of $1,000$ requested raw generation attempts in three independent runs.  When supported by the model, seeds $0$, $1$, and $2$ were used for generation. ProtoBind-Diff~\citep{mistryukova2025protobind} does not support deterministic seed control, threrfore it was evaluated in three independent stochastic runs. All generated outputs were submitted to Stage~1, which standardized structures and removed invalid and duplicate molecular graphs. The molecules followed the generic configuration defined above for each stage. Subsequent stages used the frozen generic configuration except for the explicitly defined KRAS-specific settings. For the KRAS~G12D comparison, the descriptor limits were expanded only as needed to retain $95\%$ of the KRAS reference set, producing the frozen KRAS-95 descriptor profile. The reference-derived docking cutoffs at the same $95\%$ coverage were more stringent than the generic threshold. Therefore, the cutoff remained $-6.5$~kcal/mol for each docking method in the KRAS-95 profile. Stage~6 used the default $3.0$~\AA{} conformer RMSD cutoff, and the KRAS-specific Asp12 interaction requirement. 

\subsection{Selection and calibration of HEDGEHOG filtering criteria}\label{subsec:sholds}
\textbf{Generic descriptor calibration.} The generic descriptor profile was calibrated using fixed samples of $50,000$ ZINC-250k and $50,000$ ChEMBL~34 molecules. We predefined the calibration objective as retaining at least $95\%$ of each reference sample after applying all $21$ descriptors together. We tested symmetric pooled quantile intervals on an $\alpha=0.0001$ grid. The largest value satisfying the retention objective was $\alpha=0.0029$, corresponding to the $0.29_{th}-99.71_{th}$ percentile interval. The descriptor specific profile obtained from this interval was rounded outward and frozen before validation.

\textbf{KRAS-specific descriptor calibration.} We next adapted the frozen generic descriptor profile to KRAS~G12D for use in the subsequent KRAS-based experiments. The KRAS~G12D set contained $645$ entries. After removing $13$ invalid entries and three duplicate molecular graphs, $629$ unique ligands remained. These ligands were randomly divided into an $80\%$ calibration subset and a $20\%$ holdout. Starting from the frozen generic profile, individual descriptor bounds were expanded to achieve predefined calibration set retention levels of $50\%$, $90\%$, $95\%$, and $98\%$, then fixed and validated (Table~\ref{tab:kras_results}). KRAS-95 was selected as the primary descriptor profile. It retained most target ligands on the descriptor bounds, $124$ of $126$, while limiting the influence of extreme reference compounds.

\begin{table}[ht!]
  \caption{Stage-wise retention of KRAS~G12D reference ligands under different configuration profiles. Default column reports the $629$-compound reference set, whereas KRAS adapted configurations are evaluated on the $126$ validation subset}
  \label{tab:kras_results}
  \centering
  \resizebox{\columnwidth}{!}{
      \begin{tabular}{lccccc}
          \toprule 
          \textbf{Stage}        & \textbf{Default} & \textbf{KRAS-50} & \textbf{KRAS-90} & \textbf{KRAS-95} & \textbf{KRAS-98} \\
          \midrule
          Initial               & $629$ ($100\%$)  & $126$ ($100\%$)  & $126$ ($100\%$)  & $126$ ($100\%$)  & $126$ ($100\%$) \\
          Preprocessing         & $629$ ($100\%$)  & $126$ ($100\%$)  & $126$ ($100\%$)  & $126$ ($100\%$)  & $126$ ($100\%$) \\
          Descriptors           & $282$ ($44.8\%$) & $109$ ($86.5\%$) & $123$ ($97.6\%$) & $124$ ($98.4\%$) & $126$ ($100\%$) \\
          Structural Filters    & $268$ ($42.6\%$) & $89$ ($70.6\%$)  & $90$ ($71.4\%$)  & $90$ ($71.4\%$)  & $90$ ($71.4\%$)  \\
          Synthesis Feasibility & $2$ ($0.3\%$)    & $2$ ($1.6\%$)    & $2$ ($1.6\%$)    & $2$ ($1.6\%$)    & $2$ ($1.6\%$)    \\
          Docking               & $1$ ($0.2\%$)    & $2$ ($1.6\%$)    & $2$ ($1.6\%$)    & $2$ ($1.6\%$)    & $2$ ($1.6\%$)    \\
          3D Filters            & $1$ ($0.2\%$)    & $2$ ($1.6\%$)    & $2$ ($1.6\%$)    & $2$ ($1.6\%$)    & $2$ ($1.6\%$)    \\
          \botrule
      \end{tabular}
  }
\end{table}

\textbf{Structural filters configuration.} In the frozen generic configuration, molecular survival requires passing the PAINS subset~\citep{baell2010new} of Common~Alerts which includes $481$ SMARTS criteria, the native NIBR policy with accumulated severity below 10, the native Lilly hard-reject criteria and a soft demerit score below 160, MolGraph severity below 5, the protecting-group filter, and a maximum of two undefined stereocenters (Supplementary~Table~S4 and Supplementary~File~1). The remaining structural modules are reported as diagnostics by default. This configuration was applied unchanged to KRAS~G12D. Task-specific profiles can be manually defined by selecting the enforced alert sets, individual SMARTS patterns, graph checks, and numerical thresholds. However, flagged motifs occur in established drugs and no alert collection provides a universal set of exclusion rules~\citep{bruns2012rules,capuzzi2017phantom}. We evaluated all $23$ Common~Alerts rule sets diagnostically across different data sources. The results are provided in Supplementary~Section~1.3.

\textbf{Synthesis feasibility configuration.} The frozen generic configuration applies thresholds of SA~score~$\leq 4.5$~\citep{ertl2009estimation,vorvsilak2020syba}, RA~score~$\geq~0.5$~\citep{thakkar2021retrosynthetic}, and SYBA~score~$\geq~0$~\citep{vorvsilak2020syba}. Sensitivity to SA scorecutoffs was evaluated in Supplementary~Table~S6. Synthesis score thresholds were not adapted to KRAS. 

AiZynthFinder was evaluated separately on the complete set of $629$ reference ligands. Under generic \textsc{HEDGEHOG} settings of maximum search depth $5$, time limit $300$~seconds, and $300$ iterations, at least one route was identified for $8$ compounds ($1.27\%$). For unresolved compounds, we progressively increased the maximum depth to $15$, time to $1,200$~seconds, and iterations to $4,000$. Across these searches, $230$ compounds ($36.57\%$) had at least one completed route. The improvement incurred greater computational cost, and many unresolved searches terminated at chemically simple precursors that were absent from the exact-match ZINC stock. Details of the synthesis feasibility analysis are provided in Supplementary~Section~1.4.

% AiZynthFinder used an exact-match ZINC stock with a default maximum search depth of $5$, a time limit of $300$~s per molecule, and $300$ iterations. Search-budget sensitivity was evaluated by progressively increasing these settings to a maximum depth of $15$, a time limit of $1,200$~s per molecule, and $4,000$ iterations. Molecular representation, result aggregation, and route-recovery results are described in Supplementary~Section~1.4.

\textbf{Docking threshold selection.}
HEDGEHOG applies a generic docking score cutoff of $-6.5$~kcal/mol to \textit{smina}, GNINA, and Matcha. This value was selected as a practical literature-supported cutoff commonly used in structure based virtual screening~\citep{jang2021drugs,cai2025structure,fischer2020potential}. When reference ligands are available, method-specific cutoffs are estimated jointly from the corresponding docking score distributions to achieve a predefined joint retention level.

\textbf{KRAS-specific docking threshold calibration.} Docking thresholds were calibrated on the same $80\%$~KRAS subset. Molecules lacking a score for any method were classified as failures. At the $95\%$ setting, the resulting cutoffs were $-9.01$~kcal/mol for \textit{smina}, $-10.14$~kcal/mol for GNINA, and $-9.25$~kcal/mol for Matcha. Together, these cutoffs retained $478$ of $503$ calibration and $122$ of $126$ holdout ligands. The KRAS-based cutoffs were more negative than the generic value and the cutoff remained $-6.5$~kcal/mol for \textit{smina}, GNINA, and Matcha.

\textbf{Three-dimensional filter settings.}
Stage~6 thresholds and target-specific interaction requirements are specified explicitly in the configuration and are not adapted automatically to a new target. The generic configuration uses a conformer-RMSD cutoff of $3.0$~\AA, while required or forbidden residue interactions can be defined manually for a particular target. Complete Stage~6 settings are provided in Supplementary~Section~1.6. For the KRAS~G12D case study, we manually specified an interaction with Asp12 as a target-specific requirement. 

\section{Results and discussion}\label{sec3}
\textsc{HEDGEHOG} evaluates molecular generators by measuring the sequential retention of their outputs through configurable chemical, synthesis, target-binding, and 3D filters. We first validate the frozen generic benchmark configuration using reference molecular subsets. We then compare generators under a fixed raw generation budget and examine whether end-to-end survival is reproduced by conventional generative metrics and remains robust under alternative benchmark settings.

\subsection{Evaluation of the frozen generic configuration}
To evaluate the frozen generic configuration, we processed independent molecular subsets sequentially through Stages~1-4 (Table~\ref{tab:desecr_val}). We used $50,000$-molecule subsets of ZINC-250k, ChEMBL~34, MOSES, PubChem, and GuacaMol, together with $673$ orally administered FDA-approved drugs. Preprocessing and descriptor filtering retained $97.53\%$ of ZINC-250k molecules and $95.16\%$ of ChEMBL~34 molecules. Descriptor stage retention in the other reference collections ranged from $72.21\%$ for $673$ FDA-approved oral drugs to $98.32\%$ for MOSES. After the structural and synthesis score gates, cumulative retention ranged from $44.28\%$ for the oral FDA drugs to $90.63\%$ for MOSES. The synthesis feasibility stage for this large-scale control included SA, RA, and SYBA scores criteria and omitted AiZynthFinder.

\begin{table*}[ht!]
  \caption{Sequential retention of reference molecular subsets through the frozen generic configuration. Percentages are cumulative relative to the initial subset size}
  \label{tab:desecr_val}
  \centering
  \resizebox{\textwidth}{!}{
      \begin{tabular}{lcccccc}
          \toprule
          \textbf{Dataset}      & \textbf{ZINC-250k}   & \textbf{ChEMBL~34}   & \textbf{MOSES}       & \textbf{PubChem}     & \textbf{GuacaMol}    & \textbf{FDA-approved} \\
          \midrule
          Initial               & $50,000$ ($100.0\%$) & $50,000$ ($100.0\%$) & $50,000$ ($100.0\%$) & $50,000$ ($100.0\%$) & $50,000$ ($100.0\%$) & $673$ ($100.0\%$) \\
          Preprocessing         & $49,991$ ($99.98\%$) & $49,995$ ($99.99\%$) & $50,000$ ($100.0\%$) & $49,656$ ($99.31\%$) & $49,867$ ($99.73\%$) & $670$ ($99.55\%$) \\
          Descriptors           & $48,765$ ($97.53\%$) & $47,582$ ($95.16\%$) & $49,159$ ($98.32\%$) & $39,799$ ($79.60\%$) & $38,688$ ($77.38\%$) & $486$ ($72.21\%$) \\
          Structural Filters    & $39,791$ ($79.58\%$) & $33,545$ ($67.09\%$) & $45,827$ ($91.65\%$) & $27,713$ ($55.43\%$) & $26,666$ ($53.33\%$) & $368$ ($54.68\%$) \\
          Synthesis Feasibility & $36,369$ ($72.74\%$) & $31,255$ ($62.51\%$) & $45,317$ ($90.63\%$) & $25,858$ ($51.72\%$) & $24,548$ ($49.10\%$) & $298$ ($44.28\%$) \\
          \botrule
      \end{tabular}%
  }
\end{table*}

We independently evaluated SA score thresholds on the same molecular subsets of ZINC-250k, ChEMBL~34, MOSES, and the oral FDA-approved drugs. At the selected cutoff of $4.5$, the corresponding retention rates were $93.65\%$, $97.08\%$, $99.37\%$, and $88.56\%$, respectively (Table~\ref{tab:analysis_sascore}).

\begin{table}[ht!]
\caption{Independent retention of molecular subsets at SA score cutoffs}
\label{tab:analysis_sascore}
\centering
\resizebox{\columnwidth}{!}{
  \begin{tabular}{lcccc}
      \toprule
      \textbf{SA max} & \textbf{ZINC-250k}  & \textbf{ChEMBL~34}   & \textbf{MOSES}       & \textbf{FDA-approved} \\
      \midrule
      $4.0$          & $42,676$ ($85.35\%$) & $46,880$ ($93.76\%$) & $49,684$ ($99.37\%$) & $542$ ($80.53\%$) \\ 
      $4.5$          & $46,825$ ($93.65\%$) & $48,541$ ($97.08\%$) & $49,941$ ($99.88\%$) & $596$ ($88.56\%$) \\ 
      $5.0$          & $48,687$ ($97.37\%$) & $49,337$ ($98.67\%$) & $49,987$ ($99.97\%$) & $637$ ($94.65\%$) \\
      \botrule
  \end{tabular}%
}
\end{table}

These controls show that the frozen descriptor profile achieved its predefined retention objective in independent ZINC-250k and ChEMBL~34 subsets. The structural and synthesis score gates produced greater and dataset-dependent attrition. Accordingly, the frozen generic configuration is used below as a transparent and reproducible baseline for cross-generator comparison. The effects of alternative thresholds and target-specific settings are evaluated in Section~\ref{subsec:benchmark_evaluation_setup}.

\subsection{Controlled fixed-budget generator comparison}
The main comparison used three independent runs with a fixed-budget of $1,000$ requested generation attempts per model (Table~\ref{tab:fixed-budget-final}). Within each group the survivor range remained $0.3-102$ for unconditional, $0-61$ for ligand-based, and $0-127$ for protein-based, with class-level retention means of $5.52\%$, $1.74\%$, and $2.91\%$ respectively. 

\begin{table}[!ht]
\centering
\caption{Generation output and final Stage~6 survival under a fixed-budget of $1,000$ requested raw generation attempts. Valid and unique denotes the number of distinct canonical molecular graphs among valid returned outputs. Values are mean $\pm$ sample standard deviation across three runs}
\label{tab:fixed-budget-final}
\resizebox{\columnwidth}{!}{
  \begin{tabular}{llcccc}
    \toprule
    \textbf{Rank} & \textbf{Model} & \textbf{Class} & \textbf{Returned} & \textbf{Unique valid} & \textbf{Final survivors} \\
    \midrule
    1             & Dragonfly      & PB             & $994.3 \pm 2.9$   & $959.7 \pm 5.7$       & $127.0 \pm 18.1$ \\
    2             & REINVENT4      & UN             & $1000 \pm 0$      & $985.0 \pm 2.6$       & $102.0 \pm 5.0$ \\
    3             & MoLeR          & UN             & $1000 \pm 0$      & $1000 \pm 0$          & $92.7 \pm 9.7$ \\
    4             & HierGraphVAE   & UN             & $1000 \pm 0$      & $996.3 \pm 0.6$       & $91.7 \pm 9.5$ \\
    5             & JT-VAE         & UN             & $1000 \pm 0$      & $997.7 \pm 1.2$       & $74.7 \pm 12.0$ \\
    6             & MolGPT         & UN             & $1000 \pm 0$      & $966.7 \pm 9.2$       & $73.0 \pm 6.6$ \\
    7             & REINVENT4 (P)  & LB             & $1000 \pm 0$      & $784.3 \pm 8.4$       & $61.3 \pm 11.9$ \\
    8             & DrugFlow       & PB             & $1000 \pm 0$      & $832.7 \pm 13.6$      & $58.0 \pm 8.5$ \\
    9             & Pocket2Mol     & PB             & $556.0 \pm 18.2$  & $544.7 \pm 17.1$      & $31.0 \pm 11.5$ \\
    10            & PGMG           & LB             & $1000 \pm 0$      & $925.0 \pm 15.4$      & $24.3 \pm 2.3$ \\
    11            & MolFinder      & LB             & $1000 \pm 0$      & $941.7 \pm 0.6$       & $12.0 \pm 0.0$ \\
    12            & Dragonfly (b)  & PB             & $589.0 \pm 21.7$  & $512.7 \pm 27.6$      & $9.7 \pm 3.1$ \\
    13            & ProtoBind-Diff & PB             & $266.0 \pm 8.2$   & $266.0 \pm 8.2$       & $6.0 \pm 2.6$ \\
    14            & ShEPhERD       & UN             & $641.3 \pm 8.1$   & $641.3 \pm 8.1$       & $4.0 \pm 1.0$ \\
    15            & GCPG           & LB             & $1000 \pm 0$      & $949.3 \pm 9.9$       & $3.3 \pm 1.5$ \\
    15            & GENTRL         & LB             & $1000 \pm 0$      & $380.0 \pm 5.2$       & $3.3 \pm 1.5$ \\
    16            & TGM-DLM        & UN             & $1000 \pm 0$      & $392.3 \pm 3.8$       & $3.0 \pm 0.0$ \\
    17            & DiffSBDD       & PB             & $1000 \pm 0$      & $980.3 \pm 7.1$       & $0.7 \pm 1.2$ \\
    18            & EDM            & UN             & $1000 \pm 0$      & $437.3 \pm 11.9$      & $0.3 \pm 0.6$ \\
    18            & TargetDiff     & PB             & $1000 \pm 0$      & $896.7 \pm 9.5$       & $0.3 \pm 0.6$ \\
    19            & REINVENT4 (TL) & LB             & $1000 \pm 0$      & $34.0 \pm 2.6$        & $0.0 \pm 0.0$ \\
    19            & ResGen         & PB             & $237.0 \pm 111.3$ & $237.0 \pm 111.3$     & $0.0 \pm 0.0$ \\
    \botrule
  \end{tabular}
  }
\end{table}

Several models, including ResGen, ProtoBind-Diff, Pocket2Mol, ShEPhERD, did not fill the requested molecule budget. Valid unique counts spanned $34.0\pm2.6$ for REINVENT4~(TL) to $1000\pm0$ for MoLeR. However, high graph uniqueness did not imply high \textsc{HEDGEHOG} Stage~6 survival. Precisely, DiffSBDD, TargetDiff, and GCPG each retained more than 900 unique graphs on average but fewer than four final survivors, and DiffSBDD and TargetDiff fewer than one. EDM likewise retained fewer than one survivor despite a much smaller unique set. ResGen and REINVENT4~(TL) had no final survivors. Dragonfly retained a comparable unique set to these high-uniqueness models and the largest survivor count of $127.0\pm18.1$ molecules, followed by REINVENT4 with $102.0\pm5.0$, MoLeR with $92.7\pm9.7$, and HierGraphVAE with $91.7\pm9.5$ molecules. Models differed in both end-to-end survival and the stages responsible for attrition. EDM, TGM-DLM, and ShEPhERD lost substantial fractions of their outputs during preprocessing and early chemical filtering. GCPG retained many molecules through the descriptor stage, but few survived structural filters and synthesis feasibility evaluation. REINVENT4~(P) remained large until synthesis, then dropped sharply. Detailed per stage output results for each model are provided in Supplementary~Tables~S12-S14. Complete per run generation statistics, including raw validity, graph uniqueness, and generation time are provided in Supplementary~Table~S15.

\subsection{Relationship to conventional generative metrics}
\textbf{Initial KRAS-relative novelty.} Before filtration, we characterized KRAS~G12D relative novelty of the target-conditioned generators on RDKit-valid outputs from the same fixed-budget runs (Table~\ref{tab:lb_pb_initial_novelty}). Novelty was assessed using standardized molecular graphs without stereochemistry, and scaffold novelty using achiral Bemis-Murcko scaffolds. $T_c^{\mathrm{NN}}$ is the Tanimoto similarity of a generated molecule's ECFP4 fingerprint (Morgan radius~$2$, $2048$~bits) to its nearest neighbor in the KRAS reference set. Values near zero indicate little fingerprint overlap with known ligands, whereas values near one indicate near-identical chemotypes.

Most ligand- and protein-based models produced almost entirely novel molecular graphs and scaffolds relative to the KRAS collection. The exceptions were REINVENT4~(P), for which novel-scaffold molecules fell to $87.17\pm0.77\%$, and REINVENT4~(TL) which largely reproduced reference chemistry with graph novelty $3.1\pm0.2\%$, novel-scaffold molecules $0.03\pm0.06\%$, and mean Tanimoto similarity of $0.995\pm0.001$. This is expected, as REINVENT4~(TL) was transfer-learned on a KRAS~G12D inhibitor subset, so low novelty relative to that chemistry follows directly from its training setup. Among the remaining ligand-based models, PGMG and GCPG stayed distant from KRAS ligands, whereas MolFinder and REINVENT4~(P) were substantially closer. 

Protein-based generators remained far from the KRAS inhibitor collection with mean Tanimoto about $0.14-0.20$, and with essentially no molecules at Tanimoto greater than $0.4$. This is consistent with their conditioning, that most protein-based models generate into the KRAS~G12D pocket defined from PDB~7EW9. A separate check against the crystallographic ligand 05C likewise showed low similarity of $0.085-0.124$ and no exact graph or scaffold matches (Supplementary~Table~S17). Thus, prior to \textsc{HEDGEHOG}, the generators differed strongly in how closely their outputs matched known KRAS chemistry.

\begin{table*}[htbp]
  \centering
  \caption{KRAS~G12D-relative novelty of RDKit-valid ligand-based and protein-based outputs obtained within the first $1,000$ requested raw generation attempts per seed. Values are mean $\pm$ sample standard deviation across three runs}
  \label{tab:lb_pb_initial_novelty}
  \resizebox{\textwidth}{!}{%
  \begin{tabular}{lcccccc}
      \toprule
      \textbf{Model} & \textbf{Mean $T_c^{\mathrm{NN}}$} & \textbf{$T_c^{\mathrm{NN}}\geq0.3$, \%} & \textbf{$T_c^{\mathrm{NN}}\geq0.4$, \%} & \textbf{Novelty, \%} & \textbf{Novel-scaffold, \%} & \textbf{Novel unique scaffolds, \%} \\
      \midrule
      GCPG           & $0.183 \pm 0.007$                 & $1.17 \pm 1.14$                         & $0.0 \pm 0.0$                           & $100.0 \pm 0.0$    & $100.0 \pm 0.0$               & $100.0 \pm 0.0$ \\
      GENTRL         & $0.233 \pm 0.006$                 & $14.38 \pm 2.31$                        & $2.20 \pm 0.77$                         & $100.0 \pm 0.0$    & $100.0 \pm 0.0$               & $100.0 \pm 0.0$ \\
      MolFinder      & $0.277 \pm 0.005$                 & $27.62 \pm 1.78$                        & $17.17 \pm 1.65$                        & $100.0 \pm 0.0$    & $99.50 \pm 0.41$              & $99.61 \pm 0.22$ \\
      PGMG           & $0.182 \pm 0.005$                 & $0.14 \pm 0.16$                         & $0.0 \pm 0.0$                           & $100.0 \pm 0.0$    & $100.0 \pm 0.0$               & $100.0 \pm 0.0$ \\
      REINVENT4 (P)  & $0.464 \pm 0.002$                 & $97.94 \pm 0.94$                        & $74.10 \pm 0.91$                        & $100.0 \pm 0.0$    & $87.17\pm 0.77$               & $99.45 \pm 0.21$ \\
      REINVENT4 (TL) & $0.995 \pm 0.001$                 & $100.0 \pm 0.0$                         & $100.0 \pm 0.0$                         & $3.1 \pm 0.2$      & $0.03 \pm 0.06$               & $2.78 \pm 4.81$ \\
      \midrule
      DiffSBDD       & $0.142 \pm 0.001$                 & $0.0 \pm 0.0$                           & $0.0 \pm 0.0$                           & $100.0 \pm 0.0$    & $100.0 \pm 0.0$               & $100.0 \pm 0.0$ \\
      Dragonfly      & $0.176 \pm 0.000$                 & $0.03 \pm 0.06$                         & $0.0 \pm 0.0$                           & $100.0 \pm 0.0$    & $100.0 \pm 0.0$               & $100.0 \pm 0.0$ \\
      Dragonfly (b)  & $0.167 \pm 0.001$                 & $0.0 \pm 0.0$                           & $0.0 \pm 0.0$                           & $100.0 \pm 0.0$    & $100.0 \pm 0.0$               & $100.0 \pm 0.0$ \\
      DrugFlow       & $0.186 \pm 0.001$                 & $0.28 \pm 0.14$                         & $0.0 \pm 0.0$                           & $100.0 \pm 0.0$    & $100.0 \pm 0.0$               & $100.0 \pm 0.0$ \\
      Pocket2Mol     & $0.167 \pm 0.001$                 & $0.25 \pm 0.29$                         & $0.0 \pm 0.0$                           & $100.0 \pm 0.0$    & $100.0 \pm 0.0$               & $100.0 \pm 0.0$ \\
      ProtoBind-Diff & $0.195 \pm 0.002$                 & $1.51 \pm 0.75$                         & $0.12 \pm 0.22$                         & $100.0 \pm 0.0$    & $100.0 \pm 0.0$               & $100.0 \pm 0.0$ \\
      ResGen         & $0.165 \pm 0.004$                 & $0.0 \pm 0.0$                           & $0.0 \pm 0.0$                           & $100.0 \pm 0.0$    & $100.0 \pm 0.0$               & $100.0 \pm 0.0$ \\
      TargetDiff     & $0.143 \pm 0.001$                 & $0.0 \pm 0.0$                           & $0.0 \pm 0.0$                           & $100.0 \pm 0.0$    & $100.0 \pm 0.0$               & $100.0 \pm 0.0$ \\
      \botrule
  \end{tabular}%
}
\end{table*}

\textbf{Novelty after HEDGEHOG filtration.} To determine whether \textsc{HEDGEHOG} preferentially retains molecules that resemble KRAS~G12D ligands, we compared the same $T_c^{\mathrm{NN}}$ and novelty metrics after Stage~1 preprocessing and among Stage~6 survivors from the fixed-budget runs (Table~\ref{tab:novelty_before_after_hedgehog}). For each run, the pooled budgets were $8,000$, $6,000$, and $8,000$ attempts for the unconditional, ligand-based, and protein-based groups, respectively, and total of $22$ models $\times$ $3$ runs. 
 
Across all $2,335$ final survivors, every molecule was novel at the standardized graph level. All $2,334$ survivors with a Bemis-Murcko scaffold also had a novel scaffold. Mean nearest-neighbor similarity changed little during filtering, from $0.1888\pm0.0008$ after Stage~1 to $0.1951\pm0.0021$ after Stage~6. The pooled fraction with $T_c^{\mathrm{NN}}\geq0.4$ decreased from $5.01 \pm 0.10\%$ to $2.78 \pm 0.37\%$. This overall decrease was driven by changing group composition. Precisely, within ligand-based models the $\geq0.4$ fraction remained high and slightly increased from $19.25\pm0.23\%$ to $20.77\pm2.77\%$, whereas unconditional and protein-based survivors stayed near zero on this threshold.

Model-level inspection of the ligand-based survivors shows that this residual signal is not a general preference for known KRAS chemotypes. All ligand-based Stage~6 molecules with $T_c^{\mathrm{NN}}\geq0.4$ came from REINVENT4~(P). Within that model the $T_c^{\mathrm{NN}}\geq0.4$ fraction itself fell from $71.6\pm0.5\%$ after Stage~1 to $35.5\pm3.3\%$ among survivors. Full ligand-based per-model Stage~1 compared to Stage~6 statistics are provided in Supplementary~Table~S18.

Thus \textsc{HEDGEHOG} retained neither known KRAS graphs nor known scaffolds. The absence of enrichment for close ECFP4 analogs indicates that survival is not explained by recapitulating the reference chemotype library.

\begin{table}[htbp]
  \centering
  \caption{KRAS~G12D-reference novelty among initial and final filters survivors in the controlled fixed-budget experiment. Counts and metrics are mean $\pm$ sample standard deviation across three runs. The raw-attempt budgets per run were $8,000$, $6,000$, and $8,000$ for the unconditional, ligand-based, and protein-based models, respectively}
  \label{tab:novelty_before_after_hedgehog}
  \resizebox{\columnwidth}{!}{%
  \begin{tabular}{lcccc}
    \toprule
    Metric                                       & Unconditional       & Ligand-based        & Protein-based       & All models         \\
    \midrule
    Stage~1 molecules, $N$                       & $6418.0 \pm 21.5$   & $4095.3 \pm 44.0$   & $5230.7 \pm 132.4$  & $15744.0 \pm 153.2$ \\
    Stage~1 retention, \%                        & $80.23 \pm 0.27$    & $68.26 \pm 0.73$    & $65.38 \pm 1.66$    & $71.56 \pm 0.70$ \\
    Stage~6 survivors, $N$                       & $441.3 \pm 24.4$    & $104.3 \pm 9.7$     & $232.7 \pm 31.2$    & $778.3 \pm 23.9$ \\
    Stage~6 relative to initial, \%              & $5.52 \pm 0.30$     & $1.74 \pm 0.16$     & $2.91 \pm 0.39$     & $3.54 \pm 0.11$ \\
    Mean $T_c^{\mathrm{NN}}$ after Stage~1       & $0.1571 \pm 0.0001$ & $0.2692 \pm 0.0036$ & $0.1648 \pm 0.0004$ & $0.1888 \pm 0.0008$ \\
    Mean $T_c^{\mathrm{NN}}$ after Stage~6       & $0.1739 \pm 0.0013$ & $0.3117 \pm 0.0087$ & $0.1830 \pm 0.0020$ & $0.1951 \pm 0.0021$ \\
    $T_c^{\mathrm{NN}}\geq0.4$ after Stage~1, \% & $0.0 \pm 0.0$       & $19.25 \pm 0.23$    & $0.01 \pm 0.01$     & $5.01 \pm 0.10$ \\
    $T_c^{\mathrm{NN}}\geq0.4$ after Stage~6, \% & $0.0 \pm 0.0$       & $20.77 \pm 2.77$    & $0.0 \pm 0.0$       & $2.78 \pm 0.37$ \\
    Final molecular novelty, \%                  & $100.0 \pm 0.0$     & $100.0 \pm 0.0$     & $100.0 \pm 0.0$     & $100.0 \pm 0.0$ \\
    Final scaffold novelty, \%                   & $100.0 \pm 0.0$     & $100.0 \pm 0.0$     & $100.0 \pm 0.0$     & $100.0 \pm 0.0$ \\
    \botrule
  \end{tabular}%
  }
\end{table}

\textbf{Conventional metrics and Stage~6 survival.} We next compared Stage~6 survival with conventional generative metrics for the $14$ target-conditioned models (Supplementary~Table~S16). The metrics show that strong conventional scores need not coincide with downstream survival. DiffSBDD combined high uniqueness and high IntDiv$_1$ with almost no survivors, whereas Dragonfly retained a large unique set and the largest final count. REINVENT4~(TL) had the lowest novelty and diversity and no survivors, but REINVENT4~(P) combined comparatively favorable KL and FCD values with substantial survival.

To test whether conventional generative metrics are associated with downstream \textsc{HEDGEHOG} survival, we examined model-level Spearman $\rho$ and Kendall $\tau_b$ between each metric mean and the mean number of Stage~6 survivors across three runs of $14$ target-conditioned models (Table~\ref{tab:conventional_metric_correlations}). Benjamini-Hochberg correction~\citep{benjamini1995controlling} was applied across the six metrics within each correlation family to control the false-discovery rate. Spearman correlations ranged from $-0.323$ to $0.414$, and Kendall $\tau_b$ from $-0.292$ to $0.353$. All nominal $p$-values exceeded $0.05$. None remained after Benjamini-Hochberg correction. In this sample, the examined metrics showed no monotonic association with \textsc{HEDGEHOG} survival.

\begin{table}[ht!]
    \centering
    \caption{Model-level associations between conventional generative metrics and final HEDGEHOG survival across $14$ target-conditioned models. Each observation represents the mean metric value and mean number of Stage~6 survivors across three fixed-budget runs}
    \label{tab:conventional_metric_correlations}
    \resizebox{\columnwidth}{!}{%
    \begin{tabular}{lcccccc}
        \toprule
        \textbf{Metric}  & \textbf{Spearman $\rho$} & \textbf{$p$} & \textbf{$q_{\mathrm{BH}}$} & \textbf{Kendall $\tau_b$} & \textbf{$p$} & \textbf{$q_{\mathrm{BH}}$} \\
        \midrule
        Raw validity     & $0.106$                  & $0.719$      & $0.719$                    & $0.100$                   & $0.621$      & $0.701$           \\
        Graph uniqueness & $-0.323$                 & $0.260$      & $0.519$                    & $-0.292$                  & $0.168$      & $0.505$           \\  
        KRAS novelty     & $0.414$                  & $0.141$      & $0.519$                    & $0.353$                   & $0.136$      & $0.505$           \\
        IntDiv$_1$       & $0.159$                  & $0.588$      & $0.706$                    & $0.078$                   & $0.701$      & $0.701$           \\
        KL similarity    & $0.330$                  & $0.249$      & $0.519$                    & $0.211$                   & $0.297$      & $0.594$           \\
        FCD              & $0.207$                  & $0.478$      & $0.706$                    & $0.167$                   & $0.410$      & $0.615$           \\
        \botrule
    \end{tabular}%
    }
\end{table}

\subsection{Sensitivity analysis}
For the conformer-RMSD and Asp12 sensitivity analyses, we selected five protein-based generators with native coordinate outputs. These models are DiffSBDD, DrugFlow, Pocket2Mol, ResGen, and TargetDiff, and sampled $100$ molecules from each, resulting in a total of $500$ molecules. We also included KRAS~G12D inhibitors as a target-specific control set.

\textbf{Conformer RMSD cutoff.} The conformer RMSD filter is a different quantity from crystal-pose recovery. In \textsc{HEDGEHOG} it is defined as 
\[
\min\mathrm{RMSD}(\text{docked pose},\ \text{ETKDG conformer ensemble}),
\] 
using a symmetry-corrected heavy atom RMSD. We recalculated RMSD over the complete ETKDG ensemble of $50$ conformers using the stored top-ranked Matcha poses. For generated molecules, the reported counts additionally require that the same Matcha pose passed the original search-box, pose-quality, and interaction criteria.

\begin{table}[ht!]
  \caption{Retention as a function of the \textsc{HEDGEHOG} conformer RMSD cutoff. The KRAS~G12D column reports the conformer RMSD criterion alone. Generated model columns additionally require the original search-box, pose-quality, and interaction criteria on the same Matcha poses}
  \label{tab:conformer_rmsd_sensitivity}
  \centering
  \resizebox{\columnwidth}{!}{%
      \begin{tabular}{lcccccc}
          \toprule
          \textbf{Cutoff (\AA)} & \textbf{KRAS inhibitors} & \textbf{DiffSBDD} & \textbf{DrugFlow} & \textbf{Pocket2Mol} & \textbf{ResGen} & \textbf{TargetDiff} \\
          \midrule
          $2.0$                & $420$ ($66.8\%$)          & $59$ ($59\%$)     & $60$ ($60\%$)     & $57$ ($57\%$)       & $15$ ($15\%$)   & $78$ ($78\%$) \\
          $2.5$                & $529$ ($84.1\%$)          & $66$ ($66\%$)     & $71$ ($71\%$)     & $61$ ($61\%$)       & $15$ ($15\%$)   & $81$ ($81\%$) \\
          $3.0$                & $587$ ($93.3\%$)          & $67$ ($67\%$)     & $72$ ($72\%$)     & $61$ ($61\%$)       & $15$ ($15\%$)   & $82$ ($82\%$) \\
          $3.5$                & $617$ ($98.1\%$)          & $67$ ($67\%$)     & $73$ ($73\%$)     & $62$ ($62\%$)       & $15$ ($15\%$)   & $82$ ($82\%$) \\
          $4.0$                & $625$ ($99.4\%$)          & $67$ ($67\%$)     & $73$ ($73\%$)     & $63$ ($63\%$)       & $15$ ($15\%$)   & $82$ ($82\%$) \\
          \botrule
      \end{tabular}%
  }
\end{table}

At the default cutoff of $3.0$~\AA, the conformer-RMSD criterion alone retained $587$ of $629$ ($93.3\%$) KRAS~G12D inhibitors (Table~\ref{tab:conformer_rmsd_sensitivity}). Tightening the cutoff to $2.0$~\AA\ reduced retention to $66.8\%$ of the reference set, whereas relaxing it to $3.5$~\AA\ and $4.0$~\AA\ added only a small number of extra KRAS~G12D compounds with increasing the retention to $98.1\%$ and $99.4\%$, respectively. We therefore set $3.0$~\AA\ as a practical intermediate setting. Absolute Stage~6 pass counts for the generated models changed with the cutoff, but the ranking TargetDiff $>$ DrugFlow $>$ DiffSBDD $>$ Pocket2Mol $>$ ResGen was unchanged across all five cutoffs.

\textbf{Asp12 contact requirement.} The Asp12 contact requirement was evaluated alone (Table~\ref{tab:asp12-control}) and no other Stage~6 filter was included. It retained $608$ out of $629$ known KRAS~G12D reference ligands and only $287$ out of $500$ generated protein-based candidates. Asp12 is a KRAS-specific interaction criterion that preserves the large majority of known reference ligands while selecting only a subset of generated poses with the intended local contact.

\begin{table}[!ht]
  \centering
  \caption{Effect of the Asp12 contact rule alone in 100 molecules from each native-3D generator and in 629 KRAS~G12D reference ligands}
  \label{tab:asp12-control}
  \resizebox{\columnwidth}{!}{%
    \begin{tabular}{lcccccc}
      \toprule
      \textbf{Setting} & \textbf{DiffSBDD} & \textbf{DrugFlow} & \textbf{Pocket2Mol} & \textbf{ResGen} & \textbf{TargetDiff} & \textbf{KRAS ligands} \\
      \midrule
      Asp12, OFF       & 100 & 100 & 100 & 100 & 100 & 629 \\
      Asp12, ON        & 46 & 65 & 63 & 61 & 52 & 608 \\
      \botrule
    \end{tabular}
  }
\end{table}

\textbf{Self-docking pose recovery.} Pose recovery was evaluated in a separate PDB self-docking experiment (Table~\ref{tab:kras_pose_recovery}). We assembled $15$ KRAS~G12D X-ray co-crystal complexes with organic Switch-II ligands (7EW9, 7EWA, 7EWB, 7RPZ, 7RT1-7RT5, 8ONV, 8TXE, 8TXG, 8TXH, 8WB1, and 9BL0) from RCSB~Protein~Data~Bank. For each complex, the receptor was taken from the corresponding holo structure. 

\begin{table}[ht!]
  \caption{Self docking pose recovery for $15$ KRAS~G12D co-crystal ligands using the top ranked scored pose of each method}
  \label{tab:kras_pose_recovery}
  \resizebox{\columnwidth}{!}{%
    \begin{tabular}{lccc}
      \toprule
      \textbf{Docking Tool} & \textbf{$n$} & \textbf{RMSD $< 2$~\AA} & \textbf{RMSD $< 3$~\AA} \\
      \midrule
      \textit{smina}        & $15$         & $5$ ($33.3\%$)          & $15$ ($100\%$) \\
      GNINA                 & $15$         & $5$ ($33.3\%$)          & $15$ ($100\%$) \\
      Matcha                & $15$         & $5$ ($33.3\%$)          & $12$ ($80.0\%$) \\
      \botrule
    \end{tabular}
  }
\end{table}

The crystallographic ligand defined the ground truth pose and the docking box. The ligand was then redocked from an independently generated ETKDG conformer, and the reported pose was the top ranked output of each method according to its minimized affinity. Recovery was quantified as symmetry-corrected heavy atom RMSD of that scored pose to the crystallographic ligand. At RMSD $<2$~\AA, \textit{smina}, GNINA, and Matcha each recovered $5$ of $15$ complexes. At RMSD $<3$~\AA, \textit{smina} and GNINA recovered all $15$ complexes, whereas Matcha recovered $12$ of $15$ complexes.

\textbf{Ranking robustness under alternative settings.} Beyond these Stage~6 controls, we examined ranking robustness for the $14$ target-conditioned models under alternative descriptor, structural-filter, SA, docking, AiZynthFinder, conformer-RMSD, and Asp12 settings. Rankings at Stages~2-5 use the corresponding intermediate retention counts, and Stage~6 rankings are reconstructed from final survivors under each alternative setting (Table~\ref{tab:ranking_robustness}). Across all endpoints, Spearman $\rho$ ranged from $0.802$ to $1.000$, Kendall $\tau_b$ from $0.648$ to $1.000$, and Top-5 overlap was $3/5$, $4/5$ or $5/5$. Absolute survivor counts responded strongly to some settings, but comparative orderings usually did not. 

At Stage~6, relaxing the conformer RMSD cutoff to $3.5$ or $4.0$~\AA\ increased the mean pooled survivor count by $0.9\%$ and $1.2\%$, respectively, without changing the ranking. Tightening the docking cutoff to $-7.0$~kcal/mol, switching to generic descriptor bounds, and reducing the AiZynthFinder budget to $100$ iterations reduced Stage~6 throughput by $10.1\%$, $11.3\%$, and $14.4\%$, respectively,  while preserving all five baseline Top-5 models. Disabling the Asp12 requirement increased  the mean pooled survivor count from $337.0$ to $624.3$ ($85.3\%$) and was the only Stage~6 perturbation that changed Top-5 membership. There PGMG left and MolFinder entered, while REINVENT4~(P) fell from second to fourth after larger gains by DrugFlow and Pocket2Mol.

\begin{table}[!htbp]
\centering
\caption{Stability of model rankings under alternative \textsc{HEDGEHOG} settings across $14$ target-conditioned models. Rankings are based on the mean number of retained molecules across three fixed-budget runs}
\label{tab:ranking_robustness}
\resizebox{\columnwidth}{!}{%
  \begin{tabular}{lllccc}
    \toprule
    Endpoint & Alternative configuration & Baseline              & Spearman $\rho$ & Kendall $\tau_b$ & Top-5 overlap \\
    \midrule
    Stage~2 & Generic descriptor bounds  & KRAS-95               & $0.943$         & $0.846$          & $5/5$ \\
    Stage~2 & KRAS-90 descriptor bounds  & KRAS-95               & $0.996$         & $0.978$          & $5/5$ \\
    Stage~3 & Lilly disabled             & Default               & $0.842$         & $0.714$          & $4/5$ \\
    Stage~3 & NIBR disabled              & Default               & $1.000$         & $1.000$          & $5/5$ \\
    Stage~3 & Lilly and NIBR disabled    & Default               & $0.802$         & $0.648$          & $3/5$ \\
    Stage~4 & SA score $\leq 4.0$        & SA score $\leq 4.5$   & $0.996$         & $0.978$          & $4/5$ \\
    Stage~4 & SA score $\leq 5.0$        & SA score $\leq 4.5$   & $0.991$         & $0.956$          & $5/5$ \\
    Stage~4 & AiZynthFinder: $100$ iters & $300$ iters           & $0.991$         & $0.956$          & $5/5$ \\
    Stage~4 & AiZynthFinder: $200$ iters & $300$ iters           & $0.992$         & $0.972$          & $5/5$ \\
    Stage~5 & $\leq -6.0$ kcal/mol       & $\leq -6.5$ kcal/mol  & $1.000$         & $1.000$          & $5/5$ \\
    Stage~5 & $\leq -7.0$ kcal/mol       & $\leq -6.5$ kcal/mol  & $0.996$         & $0.978$          & $5/5$ \\
    Stage~6 & RMSD $\leq 3.5$~\AA        & RMSD $\leq 3.0$~\AA   & $1.000$         & $1.000$          & $5/5$ \\
    Stage~6 & RMSD $\leq 4.0$~\AA        & RMSD $\leq 3.0$~\AA   & $1.000$         & $1.000$          & $5/5$ \\
    Stage~6 & Asp12 disabled             & Asp12 enabled         & $0.968$         & $0.883$          & $4/5$ \\
    Stage~6 & Generic descriptor bounds  & KRAS-95               & $0.994$         & $0.972$          & $5/5$ \\
    Stage~6 & KRAS-90 descriptor bounds  & KRAS-95               & $1.000$         & $1.000$          & $5/5$ \\
    Stage~6 & $\leq -7.0$ kcal/mol       & $\leq -6.5$ kcal/mol  & $0.999$         & $0.994$          & $5/5$ \\
    Stage~6 & AiZynthFinder: $100$ iters & $300$ iters           & $0.999$         & $0.994$          & $5/5$ \\
    Stage~6 & AiZynthFinder: $200$ iters & $300$ iters           & $0.999$         & $0.994$          & $5/5$ \\
  \botrule
  \end{tabular}%
  }
\end{table}

Intermediate Stage~3 rankings were more fragile. Disabling Lilly and NIBR together moved DiffSBDD from ninth to fourth and brought MolFinder into the Top-5 in place of GCPG and PGMG, and overlap $3/5$. TargetDiff also gained strongly but remained outside the Top-5. NIBR alone left the Stage~3 ordering unchanged, indicating that Lilly drives most of this sensitivity. 

Overall, \textsc{HEDGEHOG} survival counts are configuration-dependent. Final Stage~6 rankings remained stable under most tested numerical perturbations, whereas disabling the Asp12 requirement changed one Top-5 position and changing the structural-filter profile produced larger ranking shifts at Stage~3.

\section{Conclusion}\label{sec4-revised}
Molecular generators are still often ranked by validity, uniqueness, novelty, and distributional similarity, yet these scores do not tell whether outputs survive the filters used in early discovery. \textsc{HEDGEHOG} makes that attrition explicit. In three controlled runs, models differed in output efficiency, graph uniqueness, losses at each stage, and final survival. Dragonfly had the largest mean final set with $127.0\pm18.1$ molecules per $1,000$ attempts. This KRAS~G12D result does not show a general advantage of protein conditioning. Variation within each model class was large, and the models differ in architecture, training data, representation, and sampling. 

Conventional generative metrics were not monotonically associated with Stage~6 survival, and survivors were not enriched for known KRAS chemotypes. Passing \textsc{HEDGEHOG} is therefore not a proxy for looking like the reference set, nor a redundant restatement of uniqueness or FCD. Moreover, absolute survivor counts moved with threshold choices, but relative model rankings were largely stable. KRAS~G12D specific interaction criterion was the main Stage~6 exception to the ranking stability.

The following limits define the next work. Failed AiZynthFinder searches often reflect stock coverage, charge matching, and search policy rather than intrinsic synthetic impossibility. Alternative stocks and matching rules need systematic comparison before Stage~4 is treated as a hard synthesizability verdict. Multi-target controls support reuse of docking and pose gates across targets, but generator rankings here remain a KRAS~G12D case study, with no prospective assay validation and no ADMET or pharmacokinetic endpoints. Within that scope, \textsc{HEDGEHOG} complements standard generative scores by evaluating where the chemistry of a generative framework fails under early medicinal chemistry and structure based constraints.

\noindent\fbox{\parbox{\dimexpr\linewidth-2\fboxsep-2\fboxrule\relax}{%
\textbf{Key Points}
\begin{itemize}
  \item We present \textsc{HEDGEHOG}, a multi-stage benchmark for molecular generators testing, inspired by hit identification.
  \item Under a fixed budget of $1,000$ generation attempts, Stage~6 survival ranged from $0$ to $127.0\pm18.1$ molecules, with large within-class variation.
  \item  Validity, uniqueness, novelty, diversity, KL similarity, and FCD were not significantly associated with Stage~6 survival. Survivors were not enriched for known KRAS chemotypes. Absolute survivor counts depend on threshold settings, but model rankings were largely stable in sensitivity tests.
  \item \textsc{HEDGEHOG} outcomes are computational filter results and do not prove experimental activity.
\end{itemize}%
}
}
\vspace{1em}

\section{Conflicts of interest}
The authors declare that they have no competing interests.

\section{Data availability}
The \textsc{HEDGEHOG} source code is available at \url{https://github.com/LigandPro/hedgehog}.  The version used in this study is version~\texttt{1.1.42}, corresponding to commit \texttt{10792e9}. The generic and target-adapted configuration files, input sets, receptor files, random seeds, and stage-level outputs are deposited in a repository and Supplementary~Information.

\section{Author contributions statement}

Conceptualization and methodology: D.R., P.G., S.N. and M.P. Software: D.R., with contributions from P.G. and S.N. Investigation: S.K. (preparation of the KRAS G12D receptor). Software optimization: S.N. Writing—original manuscript: D.R. Writing—review and editing: M.P., P.G., S.N., D.F. and S.K. Visualization: K.F. and M.P. Supervision: A.S. All authors read and approved the final version of the manuscript and agree to be accountable for all aspects of the work.

\bibliographystyle{unsrt}
\bibliography{references}

@article{ho2020denoising,
  title={Denoising diffusion probabilistic models},
  author={Ho, Jonathan and Jain, Ajay and Abbeel, Pieter},
  journal={Advances in neural information processing systems},
  volume={33},
  pages={6840--6851},
  year={2020}
}

@article{kwon2021molfinder,
  title={MolFinder: an evolutionary algorithm for the global optimization of molecular properties and the extensive exploration of chemical space using SMILES},
  author={Kwon, Yongbeom and Lee, Juyong},
  journal={Journal of cheminformatics},
  volume={13},
  number={1},
  pages={24},
  year={2021},
  publisher={Springer}
}

@article{jin2018junction,
  title={Junction tree variational autoencoder for molecular graph generation},
  author={Jin, Wengong and Barzilay, Regina and Jaakkola, Tommi},
  journal={Proceedings of the International Conference on Machine Learning},
  pages={2323--2332},
  year={2018}
}

@article{jin2020hierarchical,
  title={Hierarchical generation of molecular graphs using structural motifs},
  author={Jin, Wengong and Barzilay, Regina and Jaakkola, Tommi},
  journal={Proceedings of the International Conference on Machine Learning},
  pages={4839--4848},
  year={2020}
}

@article{maziarz2021learning,
  title={Learning to extend molecular scaffolds with structural motifs},
  author={Maziarz, Krzysztof and Jackson-Flux, Henry and Cameron, Pashmina and Sirockin, Finton and Schneider, Nadine and Stiefl, Nikolaus and Segler, Marwin and Brockschmidt, Marc},
  journal={arXiv preprint arXiv:2103.03864},
  year={2021}
}

@article{hoogeboom2022equivariant,
  title={Equivariant diffusion for molecule generation in 3d},
  author={Hoogeboom, Emiel and Satorras, V{\i}ctor Garcia and Vignac, Cl{\'e}ment and Welling, Max},
  journal={Proceedings of the International Conference on Machine Learning},
  pages={8867--8887},
  year={2022}
}

@article{guan20233d,
  title={3d equivariant diffusion for target-aware molecule generation and affinity prediction},
  author={Guan, Jiaqi and Qian, Wesley Wei and Peng, Xingang and Su, Yufeng and Peng, Jian and Ma, Jianzhu},
  journal={arXiv preprint arXiv:2303.03543},
  year={2023}
}

@article{zhang2023resgen,
  title={ResGen is a pocket-aware 3D molecular generation model based on parallel multiscale modelling},
  author={Zhang, Odin and Zhang, Jintu and Jin, Jieyu and Zhang, Xujun and Hu, RenLing and Shen, Chao and Cao, Hanqun and Du, Hongyan and Kang, Yu and Deng, Yafeng and others},
  journal={Nature Machine Intelligence},
  volume={5},
  number={9},
  pages={1020--1030},
  year={2023},
  publisher={Nature Publishing Group UK London}
}

@article{gong2024text,
  title={Text-guided molecule generation with diffusion language model},
  author={Gong, Haisong and Liu, Qiang and Wu, Shu and Wang, Liang},
  journal={Proceedings of the AAAI Conference on Artificial Intelligence},
  volume={38},
  pages={109--117},
  year={2024}
}

@article{bagal2021molgpt,
  title={MolGPT: molecular generation using a transformer-decoder model},
  author={Bagal, Viraj and Aggarwal, Rishal and Vinod, PK and Priyakumar, U Deva},
  journal={Journal of chemical information and modeling},
  volume={62},
  number={9},
  pages={2064--2076},
  year={2021},
  publisher={ACS Publications}
}

@article{peng2022pocket2mol,
  title={Pocket2mol: Efficient molecular sampling based on 3d protein pockets},
  author={Peng, Xingang and Luo, Shitong and Guan, Jiaqi and Xie, Qi and Peng, Jian and Ma, Jianzhu},
  journal={Proceedings of the International Conference on Machine Learning},
  pages={17644--17655},

  year={2022}
}

@article{loeffler2024reinvent,
  title={Reinvent 4: modern AI--driven generative molecule design},
  author={Loeffler, Hannes H and He, Jiazhen and Tibo, Alessandro and Janet, Jon Paul and Voronov, Alexey and Mervin, Lewis H and Engkvist, Ola},
  journal={Journal of Cheminformatics},
  volume={16},
  number={1},
  pages={20},
  year={2024},
  publisher={Springer}
}

@article{atz2024prospective,
  title={Prospective de novo drug design with deep interactome learning},
  author={Atz, Kenneth and Cotos, Leandro and Isert, Clemens and H{\aa}kansson, Maria and Focht, Dorota and Hilleke, Mattis and Nippa, David F and Iff, Michael and Ledergerber, Jann and Schiebroek, Carl CG and others},
  journal={Nature Communications},
  volume={15},
  number={1},
  pages={3408},
  year={2024},
  publisher={Nature Publishing Group UK London}
}

@article{kingma2013auto,
  title={Auto-encoding variational bayes},
  author={Kingma, Diederik P and Welling, Max},
  journal={arXiv preprint arXiv:1312.6114},
  year={2013}
}

@article{weininger1988smiles,
  title={SMILES, a chemical language and information system. 1. Introduction to methodology and encoding rules},
  author={Weininger, David},
  journal={Journal of chemical information and computer sciences},
  volume={28},
  number={1},
  pages={31--36},
  year={1988},
  publisher={ACS Publications}
}

@article{polykovskiy2020molecular,
  title={Molecular sets (MOSES): a benchmarking platform for molecular generation models},
  author={Polykovskiy, Daniil and Zhebrak, Alexander and Sanchez-Lengeling, Benjamin and Golovanov, Sergey and Tatanov, Oktai and Belyaev, Stanislav and Kurbanov, Rauf and Artamonov, Aleksey and Aladinskiy, Vladimir and Veselov, Mark and others},
  journal={Frontiers in pharmacology},
  volume={11},
  pages={565644},
  year={2020},
  publisher={Frontiers Media SA}
}

@article{thomas2024molscore,
  title={MolScore: a scoring, evaluation and benchmarking framework for generative models in de novo drug design},
  author={Thomas, Morgan and O’Boyle, Noel M and Bender, Andreas and De Graaf, Chris},
  journal={Journal of Cheminformatics},
  volume={16},
  number={1},
  pages={64},
  year={2024},
  publisher={Springer}
}

@article{brown2019guacamol,
  title={GuacaMol: benchmarking models for de novo molecular design},
  author={Brown, Nathan and Fiscato, Marco and Segler, Marwin HS and Vaucher, Alain C},
  journal={Journal of chemical information and modeling},
  volume={59},
  number={3},
  pages={1096--1108},
  year={2019},
  publisher={ACS Publications}
}

@article{sterling2015zinc,
  title={ZINC 15--ligand discovery for everyone},
  author={Sterling, Teague and Irwin, John J},
  journal={Journal of chemical information and modeling},
  volume={55},
  number={11},
  pages={2324--2337},
  year={2015},
  publisher={ACS Publications}
}

@article{landrum2013rdkit,
  title={Rdkit documentation},
  author={Landrum, Greg},
  journal={Release},
  volume={1},
  number={1-79},
  pages={4},
  year={2013}
}

@article{genheden2020aizynthfinder,
  title={AiZynthFinder: a fast, robust and flexible open-source software for retrosynthetic planning},
  author={Genheden, Samuel and Thakkar, Amol and Chadimov{\'a}, Veronika and Reymond, Jean-Louis and Engkvist, Ola and Bjerrum, Esben},
  journal={Journal of cheminformatics},
  volume={12},
  number={1},
  pages={70},
  year={2020},
  publisher={Springer}
}

@article{thakkar2021retrosynthetic,
  title={Retrosynthetic accessibility score (RAscore)--rapid machine learned synthesizability classification from AI driven retrosynthetic planning},
  author={Thakkar, Amol and Chadimov{\'a}, Veronika and Bjerrum, Esben Jannik and Engkvist, Ola and Reymond, Jean-Louis},
  journal={Chemical science},
  volume={12},
  number={9},
  pages={3339--3349},
  year={2021},
  publisher={Royal Society of Chemistry}
}

@article{zhu2023pharmacophore,
  title={A pharmacophore-guided deep learning approach for bioactive molecular generation},
  author={Zhu, Huimin and Zhou, Renyi and Cao, Dongsheng and Tang, Jing and Li, Min},
  journal={Nature Communications},
  volume={14},
  number={1},
  pages={6234},
  year={2023},
  publisher={Nature Publishing Group UK London}
}

@article{zou2025structure,
  title={A structure-based framework for selective inhibitor design and optimization},
  author={Zou, Yurong and Guo, Tao and Fu, Zhiyuan and Guo, Zhongning and Bo, Weichen and Yan, Dengjie and Wang, Qiantao and Zeng, Jun and Xu, Dingguo and Wang, Taijin and others},
  journal={Communications Biology},
  volume={8},
  number={1},
  pages={422},
  year={2025},
  publisher={Nature Publishing Group UK London}
}

@article{baell2010new,
  title={New substructure filters for removal of pan assay interference compounds (PAINS) from screening libraries and for their exclusion in bioassays},
  author={Baell, Jonathan B and Holloway, Georgina A},
  journal={Journal of medicinal chemistry},
  volume={53},
  number={7},
  pages={2719--2740},
  year={2010},
  publisher={ACS Publications}
}

@article{hann1999strategic,
  title={Strategic pooling of compounds for high-throughput screening},
  author={Hann, Mike and Hudson, Brian and Lewell, Xiao and Lifely, Rob and Miller, Luke and Ramsden, Nigel},
  journal={Journal of chemical information and computer sciences},
  volume={39},
  number={5},
  pages={897--902},
  year={1999},
  publisher={ACS Publications}
}

@article{ertl2009estimation,
  title={Estimation of synthetic accessibility score of drug-like molecules based on molecular complexity and fragment contributions},
  author={Ertl, Peter and Schuffenhauer, Ansgar},
  journal={Journal of cheminformatics},
  volume={1},
  number={1},
  pages={8},
  year={2009},
  publisher={Springer}
}

@article{vorvsilak2020syba,
  title={SYBA: Bayesian estimation of synthetic accessibility of organic compounds},
  author={Vor{\v{s}}il{\'a}k, Milan and Kol{\'a}{\v{r}}, Michal and {\v{C}}melo, Ivan and Svozil, Daniel},
  journal={Journal of cheminformatics},
  volume={12},
  number={1},
  pages={35},
  year={2020},
  publisher={Springer}
}

@article{zhavoronkov2019deep,
  title={Deep learning enables rapid identification of potent DDR1 kinase inhibitors},
  author={Zhavoronkov, Alex and Ivanenkov, Yan A and Aliper, Alex and Veselov, Mark S and Aladinskiy, Vladimir A and Aladinskaya, Anastasiya V and Terentiev, Victor A and Polykovskiy, Daniil A and Kuznetsov, Maksim D and Asadulaev, Arip and others},
  journal={Nature biotechnology},
  volume={37},
  number={9},
  pages={1038--1040},
  year={2019},
  publisher={Nature Publishing Group US New York}
}

@article{schneuing2024structure,
  title={Structure-based drug design with equivariant diffusion models},
  author={Schneuing, Arne and Harris, Charles and Du, Yuanqi and Didi, Kieran and Jamasb, Arian and Igashov, Ilia and Du, Weitao and Gomes, Carla and Blundell, Tom L and Lio, Pietro and others},
  journal={Nature Computational Science},
  volume={4},
  number={12},
  pages={899--909},
  year={2024},
  publisher={Nature Publishing Group US New York}
}

@article{schneuing2025multi,
  title={Multi-domain distribution learning for de novo drug design},
  author={Schneuing, Arne and Igashov, Ilia and Dobbelstein, Adrian W and Castiglione, Thomas and Bronstein, Michael and Correia, Bruno},
  journal={arXiv preprint arXiv:2508.17815},
  year={2025}
}

@article{ghazi2025quantum,
  title={Quantum-computing-enhanced algorithm unveils potential KRAS inhibitors},
  author={Ghazi Vakili, Mohammad and Gorgulla, Christoph and Snider, Jamie and Nigam, AkshatKumar and Bezrukov, Dmitry and Varoli, Daniel and Aliper, Alex and Polykovsky, Daniil and Padmanabha Das, Krishna M and Cox Iii, Huel and others},
  journal={Nature Biotechnology},
  pages={1--6},
  year={2025},
  publisher={Nature Publishing Group US New York}
}

@article{mistryukova2025protobind,
  title={ProtoBind-Diff: A Structure-Free Diffusion Language Model for Protein Sequence-Conditioned Ligand Design},
  author={Mistryukova, Lukia and Manuilov, Vladimir and Avchaciov, Konstantin and Fedichev, Peter O},
  journal={bioRxiv},
  pages={2025--06},
  year={2025},
  publisher={Cold Spring Harbor Laboratory}
}

@article{fawcett1950bredt,
  title={Bredt's rule of double bonds in atomic-bridged-ring structures.},
  author={Fawcett, Frank S},
  journal={Chemical Reviews},
  volume={47},
  number={2},
  pages={219--274},
  year={1950},
  publisher={ACS Publications}
}

@article{mcnutt2021gnina,
  title={GNINA 1.0: molecular docking with deep learning},
  author={McNutt, Andrew T and Francoeur, Paul and Aggarwal, Rishal and Masuda, Tomohide and Meli, Rocco and Ragoza, Matthew and Sunseri, Jocelyn and Koes, David Ryan},
  journal={Journal of cheminformatics},
  volume={13},
  number={1},
  pages={43},
  year={2021},
  publisher={Springer}
}

@article{koes2013lessons,
  title={Lessons learned in empirical scoring with smina from the CSAR 2011 benchmarking exercise},
  author={Koes, David Ryan and Baumgartner, Matthew P and Camacho, Carlos J},
  journal={Journal of chemical information and modeling},
  volume={53},
  number={8},
  pages={1893--1904},
  year={2013},
  publisher={ACS Publications}
}

@article{david2020molecular,
  title={Molecular representations in AI-driven drug discovery: a review and practical guide},
  author={David, Laurianne and Thakkar, Amol and Mercado, Roc{\'\i}o and Engkvist, Ola},
  journal={Journal of cheminformatics},
  volume={12},
  number={1},
  pages={56},
  year={2020},
  publisher={Springer}
}

@article{bilodeau2022generative,
  title={Generative models for molecular discovery: Recent advances and challenges},
  author={Bilodeau, Camille and Jin, Wengong and Jaakkola, Tommi and Barzilay, Regina and Jensen, Klavs F},
  journal={Wiley Interdisciplinary Reviews: Computational Molecular Science},
  volume={12},
  number={5},
  pages={e1608},
  year={2022},
  publisher={Wiley Online Library}
}

@article{schuffenhauer2020evolution,
  title={Evolution of Novartis’ small molecule screening deck design},
  author={Schuffenhauer, Ansgar and Schneider, Nadine and Hintermann, Samuel and Auld, Douglas and Blank, Jutta and Cotesta, Simona and Engeloch, Caroline and Fechner, Nikolas and Gaul, Christoph and Giovannoni, Jerome and others},
  journal={Journal of medicinal chemistry},
  volume={63},
  number={23},
  pages={14425--14447},
  year={2020},
  publisher={ACS Publications}
}

@article{kim2023pubchem,
  title={PubChem 2023 update},
  author={Kim, Sunghwan and Chen, Jie and Cheng, Tiejun and Gindulyte, Asta and He, Jia and He, Siqian and Li, Qingliang and Shoemaker, Benjamin A and Thiessen, Paul A and Yu, Bo and others},
  journal={Nucleic acids research},
  volume={51},
  number={D1},
  pages={D1373--D1380},
  year={2023},
  publisher={Oxford University Press}
}

@article{veber2002molecular,
  title={Molecular properties that influence the oral bioavailability of drug candidates},
  author={Veber, Daniel F and Johnson, Stephen R and Cheng, Hung-Yuan and Smith, Brian R and Ward, Keith W and Kopple, Kenneth D},
  journal={Journal of medicinal chemistry},
  volume={45},
  number={12},
  pages={2615--2623},
  year={2002},
  publisher={ACS Publications}
}

@misc{hadrien_mary_2024_10535844,
  title= {datamol-io/datamol: 0.12.3},
  author = {Mary, Hadrien and Noutahi, Emmanuel and Invivo, Dom and Moreau, Michel and Pak, Steven and Gilmour, Desmond and Whitfield, Shawn and Hsu, Valence-Jonny and Hounwanou, Honoré and Kumar, Ishan and Maheshkar, Saurav and Nakata, Shuya and Kovary, Kyle M. and Wognum, Cas and Craig, Michael and Deep Source Bot},
  month= jan,
  year= 2024,
  publisher= {Zenodo},
  version= {0.12.3},
  doi= {10.5281/zenodo.10535844},
  url= {https://doi.org/10.5281/zenodo.10535844},
}

@misc{emmanuel_noutahi_2025_14588938,
  title= {datamol-io/medchem: Molecular filtering for drug discovery},
  author= {Noutahi, Emmanuel and Mary, Hadrien and Kovary, Kyle M. and Whitfield, Shawn and St-Laurent, Julien and Hounwanou, Honoré and Craig, Michael},
  month= jan,
  year= 2025,
  publisher= {Zenodo},
  version= {v2.0.5-alpha},
  doi= {10.5281/zenodo.14588938},
  url= {https://doi.org/10.5281/zenodo.14588938},
  swhid={swh:1:dir:6378513b39cc3eb02031ca48bf1667a45368aa38;origin=https://doi.org/10.5281/zenodo.14588937;visit=swh:1:snp:f0482d2f3c87fd376014d2ff90498b91a5701076;anchor=swh:1:rel:35eb8de89007aa36ab63a5ef64b6c4fdcd9d63cd;path=datamol-io-medchem-c2e0977},
}

@article{brenk2008lessons,
  title={Lessons learnt from assembling screening libraries for drug discovery for neglected diseases},
  author={Brenk, Ruth and Schipani, Alessandro and James, Daniel and Krasowski, Agata and Gilbert, Ian Hugh and Frearson, Julie and Wyatt, Paul Graham},
  journal={ChemMedChem: Chemistry Enabling Drug Discovery},
  volume={3},
  number={3},
  pages={435--444},
  year={2008},
  publisher={Wiley Online Library}
}

@article{pearce2006empirical,
  title={An empirical process for the design of high-throughput screening deck filters},
  author={Pearce, Bradley C and Sofia, Michael J and Good, Andrew C and Drexler, Dieter M and Stock, David A},
  journal={Journal of chemical information and modeling},
  volume={46},
  number={3},
  pages={1060--1068},
  year={2006},
  publisher={ACS Publications}
}

@article{loffler2025reinvent4priors,
  author = "L{\"o}ffler, Hannes",
  year = 2025,
  title = "REINVENT4 priors",
  doi = {10.5281/zenodo.15641297},
  url = {https://doi.org/10.5281/zenodo.15641297}
}

@article{bouysset2021prolif,
  title={ProLIF: a library to encode molecular interactions as fingerprints},
  author={Bouysset, C{\'e}dric and Fiorucci, S{\'e}bastien},
  journal={Journal of cheminformatics},
  volume={13},
  number={1},
  pages={72},
  year={2021},
  publisher={Springer}
}

@article{nigam2023tartarus,
  title={Tartarus: A benchmarking platform for realistic and practical inverse molecular design},
  author={Nigam, AkshatKumar and Pollice, Robert and Tom, Gary and Jorner, Kjell and Willes, John and Thiede, Luca and Kundaje, Anshul and Aspuru-Guzik, Al{\'a}n},
  journal={Advances in Neural Information Processing Systems},
  volume={36},
  pages={3263--3306},
  year={2023}
}

@article{nie2024durian,
  title={Durian: a comprehensive benchmark for structure-based 3D molecular generation},
  author={Nie, Dou and Zhao, Huifeng and Zhang, Odin and Weng, Gaoqi and Zhang, Hui and Jin, Jieyu and Lin, Haitao and Huang, Yufei and Liu, Liwei and Li, Dan and others},
  journal={Journal of Chemical Information and Modeling},
  volume={65},
  number={1},
  pages={173--186},
  year={2024},
  publisher={ACS Publications}
}

@article{garcia2022dockstring,
  title={DOCKSTRING: easy molecular docking yields better benchmarks for ligand design},
  author={Garc{\'\i}a-Orteg{\'o}n, Miguel and Simm, Gregor NC and Tripp, Austin J and Hern{\'a}ndez-Lobato, Jos{\'e} Miguel and Bender, Andreas and Bacallado, Sergio},
  journal={Journal of chemical information and modeling},
  volume={62},
  number={15},
  pages={3486--3502},
  year={2022},
  publisher={ACS Publications}
}

@article{baillif2024benchmarking,
  title={Benchmarking structure-based three-dimensional molecular generative models using GenBench3D: ligand conformation quality matters},
  author={Baillif, Benoit and Cole, Jason and McCabe, Patrick and Bender, Andreas},
  journal={arXiv preprint arXiv:2407.04424},
  year={2024}
}

@article{liu2024sddbench,
  title={SDDBench: A Benchmark for Synthesizable Drug Design},
  author={Liu, Songtao and Tu, Zhengkai and Dai, Hanjun and Liu, Peng},
  journal={arXiv preprint arXiv:2409.05822},
  year={2024}
}

@article{frolova2025matcha,
  title={Matcha: Multi-Stage Riemannian Flow Matching for Accurate and Physically Valid Molecular Docking},
  author={Frolova, Daria and Daulbaev, Talgat and Sevriugov, Egor and Nikolenko, Sergei A and Ivankov, Dmitry N and Oseledets, Ivan and Pak, Marina A},
  journal={arXiv preprint arXiv:2510.14586},
  year={2025}
}

@article{trott2010autodock,
  title={AutoDock Vina: improving the speed and accuracy of docking with a new scoring function, efficient optimization, and multithreading},
  author={Trott, Oleg and Olson, Arthur J},
  journal={Journal of computational chemistry},
  volume={31},
  number={2},
  pages={455--461},
  year={2010},
  publisher={Wiley Online Library}
}

@article{friesner2004glide,
  title={Glide: a new approach for rapid, accurate docking and scoring. 1. Method and assessment of docking accuracy},
  author={Friesner, Richard A and Banks, Jay L and Murphy, Robert B and Halgren, Thomas A and Klicic, Jasna J and Mainz, Daniel T and Repasky, Matthew P and Knoll, Eric H and Shelley, Mee and Perry, Jason K and others},
  journal={Journal of medicinal chemistry},
  volume={47},
  number={7},
  pages={1739--1749},
  year={2004},
  publisher={ACS Publications}
}

@article{verdonk2003improved,
  title={Improved protein--ligand docking using GOLD},
  author={Verdonk, Marcel L and Cole, Jason C and Hartshorn, Michael J and Murray, Christopher W and Taylor, Richard D},
  journal={Proteins: Structure, Function, and Bioinformatics},
  volume={52},
  number={4},
  pages={609--623},
  year={2003},
  publisher={Wiley Online Library}
}

@article{jain2003surflex,
  title={Surflex: fully automatic flexible molecular docking using a molecular similarity-based search engine},
  author={Jain, Ajay N},
  journal={Journal of medicinal chemistry},
  volume={46},
  number={4},
  pages={499--511},
  year={2003},
  publisher={ACS Publications}
}

@article{alhossary2015fast,
  title={Fast, accurate, and reliable molecular docking with QuickVina 2},
  author={Alhossary, Amr and Handoko, Stephanus Daniel and Mu, Yuguang and Kwoh, Chee-Keong},
  journal={Bioinformatics},
  volume={31},
  number={13},
  pages={2214--2216},
  year={2015},
  publisher={Oxford University Press}
}

@article{duffy2012early,
  title={Early phase drug discovery: cheminformatics and computational techniques in identifying lead series},
  author={Duffy, Bryan C and Zhu, Lei and Decornez, H{\'e}l{\`e}ne and Kitchen, Douglas B},
  journal={Bioorganic \& medicinal chemistry},
  volume={20},
  number={18},
  pages={5324--5342},
  year={2012},
  publisher={Elsevier}
}

@article{dahlin2014essential,
  title={The essential roles of chemistry in high-throughput screening triage},
  author={Dahlin, Jayme L and Walters, Michael A},
  journal={Future medicinal chemistry},
  volume={6},
  number={11},
  pages={1265--1290},
  year={2014},
  publisher={Taylor \& Francis}
}

@article{bedart2024emerging,
  title={Emerging structure-based computational methods to screen the exploding accessible chemical space},
  author={Bedart, Corentin and Simoben, Conrad Veranso and Schapira, Matthieu},
  journal={Current Opinion in Structural Biology},
  volume={86},
  pages={102812},
  year={2024},
  publisher={Elsevier}
}

@article{ivanenkov2019we,
  title={Are we opening the door to a new era of medicinal chemistry or being collapsed to a chemical singularity? Perspective},
  author={Ivanenkov, Yan A and Zagribelnyy, Bogdan A and Aladinskiy, Vladimir A},
  journal={Journal of medicinal chemistry},
  volume={62},
  number={22},
  pages={10026--10043},
  year={2019},
  publisher={ACS Publications}
}

@misc{nikolenko2026posecheckfast,
  author = {Nikolenko, Sergei},
  title = {Posecheck Fast},
  year = {2026},
  publisher = {GitHub},
  journal = {GitHub repository},
  howpublished = {\url{https://github.com/LigandPro/posecheck-fast}},
  commit = {7e502ec41750ddb154aeab5743da85aebbe6e9db}
}

@article{bruns2012rules,
  title={Rules for identifying potentially reactive or promiscuous compounds},
  author={Bruns, Robert F and Watson, Ian A},
  journal={Journal of medicinal chemistry},
  volume={55},
  number={22},
  pages={9763--9772},
  year={2012},
  publisher={ACS Publications}
}

@article{pak2025bento,
  title={Bento: Benchmarking Classical and AI Docking on Drug Design-Relevant Data},
  author={Pak, Marina A and Frolova, Daria and Nikolenko, Sergei A and Daulbaev, Talgat and Ryabchenko, Daria and Litvin, Anna and Gurevich, Pavel and Garifullin, Kamil and Shapeev, Alexander and Oseledets, Ivan and others},
  journal={bioRxiv},
  pages={2025--12},
  year={2025},
  publisher={Cold Spring Harbor Laboratory}
}

@article{vasta2022kras,
  title={KRAS is vulnerable to reversible switch-II pocket engagement in cells},
  author={Vasta, James D and Peacock, D Matthew and Zheng, Qinheng and Walker, Joel A and Zhang, Ziyang and Zimprich, Chad A and Thomas, Morgan R and Beck, Michael T and Binkowski, Brock F and Corona, Cesear R and others},
  journal={Nature chemical biology},
  volume={18},
  number={6},
  pages={596--604},
  year={2022},
  publisher={Nature Publishing Group US New York}
}

@misc{pdb7EW9,
  title= {GDP-bound KRAS G12D in complex with TH-Z816},
  author= {Shen, Panpan and Yang, Yu and Yang, Yunyun and Zhang, Lilan and Chen, Chun-Chi and Guo, Ray-Ting},
  year= {2021},
  publisher= {Protein Data Bank},
  doi= {10.2210/pdb7ew9/pdb},
  url= {https://doi.org/10.2210/pdb7ew9/pdb}
}

@article{mao2022kras,
  title={KRAS (G12D) can be targeted by potent inhibitors via formation of salt bridge},
  author={Mao, Zhongwei and Xiao, Hongying and Shen, Panpan and Yang, Yu and Xue, Jing and Yang, Yunyun and Shang, Yanguo and Zhang, Lilan and Li, Xin and Zhang, Yuying and others},
  journal={Cell discovery},
  volume={8},
  number={1},
  pages={5},
  year={2022},
  publisher={Springer Singapore Singapore}
}

@article{kessler2019drugging,
  title={Drugging an undruggable pocket on KRAS},
  author={Kessler, Dirk and Gmachl, Michael and Mantoulidis, Andreas and Martin, Laetitia J and Zoephel, Andreas and Mayer, Moriz and Gollner, Andreas and Covini, David and Fischer, Silke and Gerstberger, Thomas and others},
  journal={Proceedings of the National Academy of Sciences},
  volume={116},
  number={32},
  pages={15823--15829},
  year={2019},
  publisher={National Academy of Sciences}
}

@article{ivanenkov2023hitchhiker,
  title={The hitchhiker’s guide to deep learning driven generative chemistry},
  author={Ivanenkov, Yan and Zagribelnyy, Bogdan and Malyshev, Alex and Evteev, Sergei and Terentiev, Victor and Kamya, Petrina and Bezrukov, Dmitry and Aliper, Alex and Ren, Feng and Zhavoronkov, Alex},
  journal={ACS Medicinal Chemistry Letters},
  volume={14},
  number={7},
  pages={901--915},
  year={2023},
  publisher={ACS Publications}
}

@article{du2024machine,
  title={Machine learning-aided generative molecular design},
  author={Du, Yuanqi and Jamasb, Arian R and Guo, Jeff and Fu, Tianfan and Harris, Charles and Wang, Yingheng and Duan, Chenru and Li{\`o}, Pietro and Schwaller, Philippe and Blundell, Tom L},
  journal={Nature Machine Intelligence},
  volume={6},
  number={6},
  pages={589--604},
  year={2024},
  publisher={Nature Publishing Group UK London}
}

@article{tropsha2024integrating,
  title={Integrating QSAR modelling and deep learning in drug discovery: the emergence of deep QSAR},
  author={Tropsha, Alexander and Isayev, Olexandr and Varnek, Alexandre and Schneider, Gisbert and Cherkasov, Artem},
  journal={Nature Reviews Drug Discovery},
  volume={23},
  number={2},
  pages={141--155},
  year={2024},
  publisher={Nature Publishing Group UK London}
}

@article{preuer2018frechet,
  title={Fr{\'e}chet ChemNet distance: a metric for generative models for molecules in drug discovery},
  author={Preuer, Kristina and Renz, Philipp and Unterthiner, Thomas and Hochreiter, Sepp and Klambauer, Gunter},
  journal={Journal of chemical information and modeling},
  volume={58},
  number={9},
  pages={1736--1741},
  year={2018},
  publisher={ACS Publications}
}

@article{zhou2024artificial,
  title={An artificial intelligence accelerated virtual screening platform for drug discovery},
  author={Zhou, Guangfeng and Rusnac, Domnita-Valeria and Park, Hahnbeom and Canzani, Daniele and Nguyen, Hai Minh and Stewart, Lance and Bush, Matthew F and Nguyen, Phuong Tran and Wulff, Heike and Yarov-Yarovoy, Vladimir and others},
  journal={Nature Communications},
  volume={15},
  number={1},
  pages={7761},
  year={2024},
  publisher={Nature Publishing Group UK London}
}

@article{zhao2024science,
  title={The science and art of structure-based virtual screening},
  author={Zhao, Hongtao},
  journal={ACS Medicinal Chemistry Letters},
  volume={15},
  number={4},
  pages={436--440},
  year={2024},
  publisher={ACS Publications}
}

@article{huang2021therapeutics,
  title={Therapeutics data commons: Machine learning datasets and tasks for drug discovery and development},
  author={Huang, Kexin and Fu, Tianfan and Gao, Wenhao and Zhao, Yue and Roohani, Yusuf and Leskovec, Jure and Coley, Connor W and Xiao, Cao and Sun, Jimeng and Zitnik, Marinka},
  journal={arXiv preprint arXiv:2102.09548},
  year={2021}
}

@article{segler2018planning,
  title={Planning chemical syntheses with deep neural networks and symbolic AI},
  author={Segler, Marwin HS and Preuss, Mike and Waller, Mark P},
  journal={Nature},
  volume={555},
  number={7698},
  pages={604--610},
  year={2018},
  publisher={Nature Publishing Group UK London}
}

@article{swanson2024generative,
  title={Generative AI for designing and validating easily synthesizable and structurally novel antibiotics},
  author={Swanson, Kyle and Liu, Gary and Catacutan, Denise B and Arnold, Autumn and Zou, James and Stokes, Jonathan M},
  journal={Nature machine intelligence},
  volume={6},
  number={3},
  pages={338--353},
  year={2024},
  publisher={Nature Publishing Group UK London}
}

@inproceedings{lee2023exploring,
  title={Exploring chemical space with score-based out-of-distribution generation},
  author={Lee, Seul and Jo, Jaehyeong and Hwang, Sung Ju},
  booktitle={International Conference on Machine Learning},
  pages={18872--18892},
  year={2023},
  organization={PMLR}
}

@article{imrie2021generating,
  title={Generating property-matched decoy molecules using deep learning},
  author={Imrie, Fergus and Bradley, Anthony R and Deane, Charlotte M},
  journal={Bioinformatics},
  volume={37},
  number={15},
  pages={2134--2141},
  year={2021},
  publisher={Oxford University Press}
}

@article{jang2021drugs,
  title={Drugs repurposed for COVID-19 by virtual screening of 6,218 drugs and cell-based assay},
  author={Jang, Woo Dae and Jeon, Sangeun and Kim, Seungtaek and Lee, Sang Yup},
  journal={Proceedings of the National Academy of Sciences},
  volume={118},
  number={30},
  pages={e2024302118},
  year={2021},
  publisher={National Academy of Sciences}
}

@article{cai2025structure,
  title={Structure-Based Virtual Screening of Potential Inhibitors Targeting the Prolyl-tRNA Synthetase (PRS) in Eimeria tenella: Insights from Molecular Docking, ADMET Studies, and Molecular Dynamics Simulations},
  author={Cai, Haiming and Liao, Shenquan and Li, Juan and Lv, Minna and Lin, Xuhui and Song, Yongle and Chen, Xiangjie and Zhu, Yibin and Zhang, Jianfei and Qi, Nanshan and others},
  journal={Molecules},
  volume={30},
  number={4},
  pages={790},
  year={2025},
  publisher={MDPI}
}

@article{fischer2020potential,
  title={Potential inhibitors for novel coronavirus protease identified by virtual screening of 606 million compounds},
  author={Fischer, Andr{\'e} and Sellner, Manuel and Neranjan, Santhosh and Smie{\v{s}}ko, Martin and Lill, Markus A},
  journal={International journal of molecular sciences},
  volume={21},
  number={10},
  pages={3626},
  year={2020},
  publisher={MDPI}
}

@article{mysinger2012directory,
  title={Directory of useful decoys, enhanced (DUD-E): better ligands and decoys for better benchmarking},
  author={Mysinger, Michael M and Carchia, Michael and Irwin, John J and Shoichet, Brian K},
  journal={Journal of medicinal chemistry},
  volume={55},
  number={14},
  pages={6582},
  year={2012}
}

@article{salentin2015plip,
  title={PLIP: fully automated protein--ligand interaction profiler},
  author={Salentin, Sebastian and Schreiber, Sven and Haupt, V Joachim and Adasme, Melissa F and Schroeder, Michael},
  journal={Nucleic acids research},
  volume={43},
  number={W1},
  pages={W443--W447},
  year={2015},
  publisher={Oxford University Press}
}

@article{amaro2018ensemble,
  title={Ensemble docking in drug discovery},
  author={Amaro, Rommie E and Baudry, Jerome and Chodera, John and Demir, {\"O}zlem and McCammon, J Andrew and Miao, Yinglong and Smith, Jeremy C},
  journal={Biophysical journal},
  volume={114},
  number={10},
  pages={2271--2278},
  year={2018},
  publisher={Elsevier}
}

@article{totrov2008flexible,
  title={Flexible ligand docking to multiple receptor conformations: a practical alternative},
  author={Totrov, Maxim and Abagyan, Ruben},
  journal={Current opinion in structural biology},
  volume={18},
  number={2},
  pages={178--184},
  year={2008},
  publisher={Elsevier}
}

@article{capuzzi2017phantom,
  title={Phantom PAINS: problems with the utility of alerts for pan-assay interference compounds},
  author={Capuzzi, Stephen J and Muratov, Eugene N and Tropsha, Alexander},
  journal={Journal of chemical information and modeling},
  volume={57},
  number={3},
  pages={417},
  year={2017}
}

@inproceedings{adams2025shepherd,
  title={ShEPhERD: Diffusing shape, electrostatics, and pharmacophores for bioisosteric drug design},
  author={Adams, Keir and Abeywardane, Kento and Fromer, Jenna and Coley, Connor},
  booktitle={International Conference on Learning Representations},
  volume={2025},
  pages={32280--32342},
  year={2025}
}

@article{huth2005alarm,
  title={ALARM NMR: a rapid and robust experimental method to detect reactive false positives in biochemical screens},
  author={Huth, Jeffrey R and Mendoza, Renaldo and Olejniczak, Edward T and Johnson, Robert W and Cothron, Darlene A and Liu, Yaya and Lerner, Claude G and Chen, Jun and Hajduk, Philip J},
  journal={Journal of the American Chemical Society},
  volume={127},
  number={1},
  pages={217--224},
  year={2005},
  publisher={ACS Publications}
}

@article{benigni2008structure,
  title={Structure alerts for carcinogenicity, and the Salmonella assay system: a novel insight through the chemical relational databases technology},
  author={Benigni, Romualdo and Bossa, Cecilia},
  journal={Mutation Research/Reviews in Mutation Research},
  volume={659},
  number={3},
  pages={248--261},
  year={2008},
  publisher={Elsevier}
}

@article{brenke2016identification,
  title={Identification of small-molecule frequent hitters of glutathione S-transferase--glutathione interaction},
  author={Brenke, Jara K and Salmina, Elena S and Ringelstetter, Larissa and Dornauer, Scarlett and Kuzikov, Maria and Rothenaigner, Ina and Schorpp, Kenji and Giehler, Fabian and Gopalakrishnan, Jay and Kieser, Arnd and others},
  journal={Journal of biomolecular screening},
  volume={21},
  number={6},
  pages={596--607},
  year={2016},
  publisher={SAGE Publications Sage CA: Los Angeles, CA}
}

@article{schorpp2014identification,
  title={Identification of small-molecule frequent hitters from AlphaScreen high-throughput screens},
  author={Schorpp, Kenji and Rothenaigner, Ina and Salmina, Elena and Reinshagen, Jeanette and Low, Terence and Brenke, Jara K and Gopalakrishnan, Jay and Tetko, Igor V and Gul, Sheraz and Hadian, Kamyar},
  journal={Journal of biomolecular screening},
  volume={19},
  number={5},
  pages={715--726},
  year={2014},
  publisher={SAGE Publications Sage CA: Los Angeles, CA}
}

@article{benjamini1995controlling,
  title={Controlling the false discovery rate: a practical and powerful approach to multiple testing},
  author={Benjamini, Yoav and Hochberg, Yosef},
  journal={Journal of the Royal statistical society: series B (Methodological)},
  volume={57},
  number={1},
  pages={289--300},
  year={1995},
  publisher={Wiley Online Library}
}

\begin{biography}{}{The authors comprise a multidisciplinary team from Ligand Pro and the Applied AI Institute, with expertise in molecular generative modeling, cheminformatics, structure-based drug design, synthetic accessibility, and computational drug discovery.}
\end{biography}

% Supplementary Information after references (S-numbering for figures/tables/pages).
\clearpage
\setcounter{figure}{0}
\setcounter{table}{0}
\setcounter{section}{0}
\setcounter{subsection}{0}
\setcounter{subsubsection}{0}
\setcounter{page}{1}
\renewcommand{\thefigure}{S\arabic{figure}}
\renewcommand{\thetable}{S\arabic{table}}
\renewcommand{\figurename}{Figure}
\renewcommand{\tablename}{Table}
\renewcommand{\thesection}{\arabic{section}.}
\renewcommand{\thesubsection}{\thesection\arabic{subsection}.}
\renewcommand{\thesubsubsection}{\thesubsection\arabic{subsubsection}.}
\renewcommand{\thepage}{S-\arabic{page}}
\begin{appendices}

\onecolumn

\section*{Supplementary Information}
\section*{Table of Contents}

\begin{enumerate}[0.]
\item \hyperref[appendix:hedgehog-details]{Detailed molecular workflow of the \textsc{HEDGEHOG} pipeline} (Table~S1)
    \begingroup
    \renewcommand{\labelenumii}{\theenumi.\arabic{enumii}.}
    \begin{enumerate}[00.00.]
        \item \hyperref[appendix:molecules-preparation]{Molecular preprocessing definitions} (Table~S2)
        \item \hyperref[appendix:descriptors]{Physicochemical descriptors definitions and thresholds} (Table~S3)
        \item \hyperref[appendix:detailed-mol-flow-structfilters]{Structural filters definitions and thresholds} (Table~S4, Table~S5)
        \item \hyperref[appendix:detailed-mol-flow-retro]{Synthesis feasibility definitions and thresholds} (Table~S6, Table~S7, Fig.~S1)
        \item \hyperref[appendix:detailed-mol-flow-docking]{Docking definitions and thresholds} (Fig.~S2)
        \item \hyperref[appendix:detailed-mol-flow-3dfilters]{Three-dimensional filtration definitions and thresholds} (Table~S8, Fig.~S3)
        \item \hyperref[appendix:runtime]{Per-stage runtime} (Table~S9)
        \item \hyperref[appendix:models-overview]{Model Overview} (Table~S10)
        \item \hyperref[appendix:fixed-budget]{Fixed-budget stage-wise evaluation} (Table~S11, Table~S12, Table~S13, Table~S14, Table~S15, Table~S16, Table~S17, Table~S18, Fig.~S4, Fig.~S5)
    \end{enumerate}
    \endgroup

\item \hyperref[appendix:kras-instance]{KRAS G12D benchmark instance}
\item \hyperref[appendix:reinvent4]{REINVENT4 configurations}
\item \hyperref[appendix:validation]{Validation and controlled experiments}
    \begingroup
        \renewcommand{\labelenumii}{\theenumi.\arabic{enumii}.}
        \begin{enumerate}[00.00.]
            \item \hyperref[appendix:ref-validation]{Reference-collection and target validation} (Table~S19, Table~S20, Table~S21)
        \end{enumerate}
        \endgroup
\end{enumerate}

\newpage
\twocolumn

% SECTION 1
\section{Detailed molecular workflow of the \textsc{HEDGEHOG} pipeline} 
\label{appendix:hedgehog-details}

The code is available at \url{https://github.com/LigandPro/hedgehog}. Unless otherwise stated, all experiments use the core benchmark dependencies listed in Table~\ref{tab:detail-versions}.

\begin{table}[!ht]
    \centering
    \caption{Core \textsc{HEDGEHOG} benchmark dependencies. 
    Main domain-specific packages and tools used in the evaluation pipeline}
    \label{tab:detail-versions}
    \begin{tabular}{@{} >{\raggedright\arraybackslash}p{0.15\textwidth} >{\centering\arraybackslash}p{0.13\textwidth} >{\raggedright\arraybackslash}p{0.20\textwidth} @{}}
        \toprule
        \textbf{Package / Tool}                           & \textbf{Version} & \textbf{Pipeline Stage} \\
        \midrule
        RDKit~\citep{landrum2013rdkit}                    & $\geq$ 2022.3.0  & Stages 1-6 \\
        DataMol~\citep{hadrien_mary_2024_10535844}        & $\geq$ 0.8.0     & Stages 1, 3, 5 \\
        MedChem~\citep{emmanuel_noutahi_2025_14588938}    & $\geq$ 0.1.0     & Preprocessing \\
        SYBA~\citep{vorvsilak2020syba}                    & = 1.0.2.alpha    & Synthesis Feasibility \\
        AiZynthFinder~\citep{genheden2020aizynthfinder}   & $\geq$ 4.3.2     & Synthesis Feasibility \\
        smina~\citep{koes2013lessons}                     & Oct. 15 2019     & Docking \\
        GNINA~\citep{mcnutt2021gnina}                     & = 1.3.2          & Docking \\
        Matcha~\citep{frolova2025matcha}                  & = 2.0.0          & Docking \\
        PoseCheck Fast~\citep{nikolenko2026posecheckfast} & $\geq$ 0.1.6     & Three-Dimensional Filters \\
        ProLIF~\citep{bouysset2021prolif}                 & $\geq$ 2.1.0     & Three-Dimensional Filters \\
        \botrule
    \end{tabular}
\end{table}

A molecule is tested at stage $s+1$ only if it passes the active filters at stage $s$. Each analysis uses either the frozen generic profile or a target-adapted profile. The profile is fixed before evaluation. Target adaptation can widen descriptor limits and use less strict method-specific docking cutoffs. It does not automatically change the structural alert, synthesis score gates or three-dimensional (3D) filters.

% SECTION 1.1
\subsection{Molecular preprocessing definitions}\label{appendix:molecules-preparation}
Stage 1 standardizes all generated SMILES strings so that downstream calculations are performed on a consistent and comparable set of molecules. Different model architectures, training datasets and objectives, and sampling procedures can produce outputs in different formats. For that reason, \textsc{HEDGEHOG} first parses each SMILES with DataMol and RDKit, repairs valence issues when possible, and sanitizes the resulting molecule. Each molecule is then checked  for atom-type compatibility with a predefined restricted set of $11$ elements: H, C, N, O, F, P, S, Cl, Br, I, and Si. Although some approved drugs contain metals, boron, silicon, or other elements, this restricted atom set is appropriate for a general small molecule generation benchmark because it avoids model-dependent failures caused by uncommon chemistry.

Generated structures can contain multiple fragments, salts, or solvents. \textsc{HEDGEHOG} keeps the largest relevant fragment, disconnects metals, normalizes functional groups, reionizes the molecule, and removes salts and solvents. It also rejects radicals. Isotopes and formal charges are retained. Stereochemical labels are kept by default. Canonical isomeric representations are used to remove duplicate structures. Different strings for the same stereochemical graph are merged, while different stereoisomers remain separate. 

Preserving stereochemistry is important for downstream evaluation. First, structural filtering evaluates the number of stereocenters and undefined stereocenters. Removing stereochemical annotations can convert defined stereocenters into unspecified ones and thereby change structural filter decisions. Second, AiZynthFinder stock matching relies on complete InChIKeys which encode stereochemical information. Removing stereochemical information can change the resulting InChIKey and may therefore affect whether a stereochemically defined precursor is recognized as available in the stock. Finally, stereochemistry determines the 3D arrangement of substituents and can affect accessible docking poses and docking scores.

To quantify the effect of stereochemistry removal, we evaluated five protein-based generators, namely DiffSBDD~\citep{schneuing2024structure}, DrugFlow~\citep{schneuing2025multi}, Pocket2Mol~\citep{peng2022pocket2mol}, ResGen~\citep{zhang2023resgen}, and TargetDiff~\citep{guan20233d}, with native coordinate outputs. We sampled $100$ molecules from each model, giving $500$ molecules in total. Then evaluated sampled molecules with or without stereochemistry using generic configuration. Table~\ref{tab:stereo_nonstereo_res} reports the pass rates of each model. Each stage was evaluated independently on the same molecule subset.

\begin{table}[ht!]
  \caption{Independent stage-wise pass counts for $100$ sampled molecules from protein-based models, evaluated with the original stereochemical annotations and after their removal. Stages were evaluated independently}
  \label{tab:stereo_nonstereo_res}
  \centering
  \resizebox{\columnwidth}{!}{%
      \begin{tabular}{lccccc}
          \toprule
          \textbf{Model}        & \textbf{DiffSBDD} & \textbf{DrugFlow} & \textbf{Pocket2Mol} & \textbf{ResGen} & \textbf{TargetDiff} \\
          \midrule
          Stage~1, stereo         & $100$ ($100\%$) & $100$ ($100\%$)   & $100$ ($100\%$)     & $100$ ($100\%$) & $94$ ($94\%$) \\
          Stage~1, stereo removed & $100$ ($100\%$) & $100$ ($100\%$)   & $100$ ($100\%$)     & $100$ ($100\%$) & $94$ ($94\%$) \\
          Stage~2, stereo         & $68$ ($68\%$)   & $85$ ($85\%$)     & $61$ ($61\%$)       & $29$ ($29\%$)   & $61$ ($61\%$) \\
          Stage~2, stereo removed & $68$ ($68\%$)   & $85$ ($85\%$)     & $61$ ($61\%$)       & $29$ ($29\%$)   & $61$ ($61\%$) \\
          Stage~3, stereo         & $28$ ($28\%$)   & $49$ ($49\%$)     & $27$ ($27\%$)       & $38$ ($38\%$)   & $35$ ($35\%$) \\
          Stage~3, stereo removed & $15$ ($15\%$)   & $41$ ($41\%$)     & $21$ ($21\%$)       & $34$ ($34\%$)   & $3$ ($3\%$)   \\
          Stage~4, stereo         & $8$ ($8\%$)     & $15$ ($15\%$)     & $11$ ($11\%$)       & $15$ ($15\%$)   & $2$ ($2\%$) \\
          Stage~4, stereo removed & $6$ ($6\%$)     & $14$ ($14\%$)     & $8$ ($8\%$)         & $15$ ($15\%$)   & $2$ ($2\%$)   \\
          Stage~5, stereo         & $70$ ($70\%$)   & $95$ ($95\%$)     & $85$ ($85\%$)       & $63$ ($63\%$)   & $79$ ($79\%$) \\
          Stage~5, stereo removed & $67$ ($67\%$)   & $95$ ($95\%$)     & $84$ ($84\%$)       & $62$ ($62\%$)   & $77$ ($77\%$) \\
          Stage~6, stereo         & $4$ ($4\%$)     & $68$ ($68\%$)     & $18$ ($18\%$)       & $0$ ($0\%$)     & $38$ ($38\%$) \\
          Stage~6, stereo removed & $4$ ($4\%$)     & $68$ ($68\%$)     & $18$ ($18\%$)       & $0$ ($0\%$)     & $38$ ($38\%$) \\  
          % Stage~6, stereo, docked         & $35$ ($35\%$)   & $58$ ($58\%$)   & $27$ ($27\%$)   & $45$ ($45\%$)   & $44$ ($44\%$) \\
          % Stage~6, stereo removed, docked & $37$ ($35\%$)   & $63$ ($35\%$)   & $32$ ($35\%$)   & $47$ ($35\%$)   & $46$ ($35\%$) \\
          \botrule       
      \end{tabular}%
  }
\end{table}

Removal of stereochemical annotations changed the structural filter (Stage~3) decision for $63$ of $500$ paired molecules and reduced the structural filters pass count from $177$ to $114$. At Stage~4, the retrosynthetic outcome changed for eight molecules, reducing the solved count from $51$ to $45$. At Stage~5, the number passing all three docking score cutoffs decreased from $392$ to $385$. On the same native poses, Stage~6 pass counts were unchanged. The default protocol therefore keeps stereochemistry. Native versus redocked poses are compared in Section~\ref{appendix:detailed-mol-flow-3dfilters}

% SECTION 1.2
\subsection{Physicochemical descriptors definitions and thresholds}\label{appendix:descriptors}
\textsc{HEDGEHOG} computes $21$ physicochemical descriptors using RDKit, including one custom molecular complexity term, MCE-18~\citep{ivanenkov2019we}. All descriptors are two-dimensional (2D) and do not depend on 3D conformations.

The generic limits were calculated from fixed samples of $50,000$ ZINC-250k molecules (random state 42) and $50,000$ ChEMBL~34 molecules (random state 20260829). For each descriptor, values from two samples were pooled, and two-sided empirical intervals $[q_{\alpha},q_{1-\alpha}]$ were evaluated using the same tail probability $\alpha$ for all descriptors. We scanned $\alpha$ from $0.01$ to $0$ in steps of $10^{-4}$. We selected the narrowest interval that retained at least $95\%$ of each calibration set across all $21$ descriptors. The $0.29_{th}-99.71_{th}$ percentile interval retained $48,714$ ZINC-250k molecules ($97.428\%$) and $47,548$ ChEMBL~34 molecules ($95.096\%$). Its limits were rounded outward and frozen. The next larger grid value, $\alpha=0.0030$, gave a narrower interval that retained $48,609$ ZINC-250k molecules ($97.22\%$) but $47,256$ ChEMBL~34 molecules ($94.51\%$), and therefore did not meet the retention criterion. 

If a target reference set is available, a generic bound may be expanded, but not narrowed, to a predefined calibration retention level. Table~\ref{tab-descr-thresholds-revised} reports the frozen generic profile and the KRAS-95 profile used in the main comparison. Bold values are the endpoints widened on the $80\%$ KRAS~G12D calibration subset.

\begin{table}[!htbp]
    \centering
    \caption{Frozen generic descriptor limits calculated from the pooled ZINC-250k and ChEMBL~34 calibration samples, and KRAS-95 configuration thresholds }
    \label{tab-descr-thresholds-revised}
    \resizebox{\columnwidth}{!}{
        \begin{tabular}{lcc}
        \toprule
        \textbf{Metric}                        & \textbf{Generic bound} & \textbf{KRAS-95 bound} \\
        \midrule
        Total number of atoms                          & $[19,81]$      & $[19,\mathbf{104}]$     \\
        Number of heavy atoms                          & $[11,43]$      & $[11,\mathbf{51}]$      \\
        Number of heteroatoms                          & $[1,14]$       & $[1,14]$                \\
        Number of N atoms                              & $[0,9]$        & $[0,9]$                 \\
        Fraction of N atoms among heavy atoms          & $[0,0.334]$    & $[0,0.334]$             \\
        Fraction of N and S atoms among heavy atoms    & $[0,0.364]$    & $[0,0.364]$             \\
        Molecular weight                               & $[164,603]$    & $[164,\mathbf{717.32}]$ \\
        logP                                           & $[-2.2,5.5]$   & $[-2.2,\mathbf{7.05}]$  \\
        Estimated water solubility                     & $[-25.6,0.0]$  & $[\mathbf{-28.18},0.0]$ \\
        Ring sizes                                     & $[3,7]$        & $[3,7]$                 \\
        Number of rings                                & $[0,7]$        & $[0,\mathbf{10}]$       \\
        Number of aromatic rings                       & $[0,5]$        & $[0,\mathbf{6}]$        \\
        Number of fused aromatic rings                 & $[0,2]$        & $[0,\mathbf{3}]$        \\
        Number of rigid bonds                          & $[8,41]$       & $[8,\mathbf{51}]$       \\
        Number of rotatable bonds                      & $[0,10]$       & $[0,10]$                \\
        Hydrogen-bond donors                           & $[0,5]$        & $[0,5]$                 \\
        Hydrogen-bond acceptors                        & $[0,11]$       & $[0,11]$                \\
        Fraction of sp\textsuperscript{3}-hybridized C & $[0,1]$        & $[0,1]$                 \\
        MCE-18                                         & $[4,144]$      & $[4,\mathbf{236.89}]$   \\
        Topological polar surface area                 & $[7,183]$      & $[7,183]$               \\
        Quantitative estimate of drug-likeness         & $[0.17,0.942]$ & $[\mathbf{0.14},0.942]$ \\
        \botrule
    \end{tabular}
    }
\end{table}
 
Figure~\ref{fig:detail-dscrip} shows descriptor distributions from the output analysis. The dashed lines in that figure are the limits of the KRAS-95 bounds.

% SECTION 1.3
\subsection{Structural filters definitions and thresholds}\label{appendix:detailed-mol-flow-structfilters}
Stage~3 is a configurable triage. All structural modules can be calculated, but only the named subset in Table~\ref{tab:app-struct-filters-revised} is used for the frozen generic survival gate. 

The Common~Alerts collection comprises $23$ families drawn from ChEMBL and literature SMARTS sets curated in MedChem, including Alarm-NMR~\citep{huth2005alarm}; AlphaScreen-Hitters, Frequent-Hitter and HIS-Hitters~\citep{schorpp2014identification}; BMS~\citep{pearce2006empirical}; Dundee~\citep{brenk2008lessons}; GST-Hitters~\citep{brenke2016identification}; Benigni genotoxic and non-genotoxic carcinogenicity alerts~\citep{benigni2008structure}; Glaxo~\citep{hann1999strategic}; PAINS~\citep{baell2010new}; SureChEMBL and Toxicophore  The remaining families of filters, namely MLSMR, Inpharmatica, LINT, and additional ToxAlerts-derived sets such as Chelator, DNABinder, Electrophilic, Skin, and Toxicophore, LD50-Oral, LuciferaseInhibitor,  Reactive-Unstable-Toxic are derived from Common~Alerts catalog.

\begin{table}[!htbp]
    \centering
    \caption{Structural modules available in \textsc{HEDGEHOG}. All modules are calculated, but only modules marked "Yes" contribute to the frozen generic survival gate}
    \label{tab:app-struct-filters-revised}
        \begin{tabularx}{\columnwidth}{@{}
          >{\raggedright\arraybackslash}p{0.15\columnwidth}
          >{\centering\arraybackslash}p{0.07\columnwidth}
          >{\centering\arraybackslash}p{0.16\columnwidth}
          >{\raggedright\arraybackslash}X
        @{}}
            \toprule
            \textbf{Module}   & \textbf{Default} & \textbf{Setting} & \textbf{Description} \\
            \midrule
            Common Alerts           & Yes & PAINS                   & SMARTS-based structural PAINS alerts \\
            MolGraph Statistics     & Yes & $<5$                    & Graph-theoretic severity metrics, evaluated over levels 1-11; identify topological anomalies and molecular graph pathologies\\
            NIBR                    & Yes & $<10$                   & Novartis screening-deck structural alerts and developability concerns~\citep{schuffenhauer2020evolution}\\
            Lilly                   & Yes & $<160$ and hard rejects & Eli Lilly medicinal chemistry rules based on demerit scoring~\citep{bruns2012rules}\\
            Protecting Groups       & Yes & \cdash                  & Terminal protecting group motifs (Boc, Fmoc, and Cbz)\\
            Undefined stereocenters & Yes & $\leq 2$                & Caps underspecified stereochemistry; reduce ambiguity in 3D structures\\
            Molecular Complexity    & No  & \cdash                  & Complexity metrics; identify outliers relative to distributions from ZINC-15 \\
            Bredt                   & No  & \cdash                  & Bredt's-rule violations at bridgehead positions in constrained bicyclic systems~\citep{fawcett1950bredt}\\
            Ring Infraction         & No  & heterocycle size $> 4$  & Flags small rings with heteroatoms or non-single bonds \\
            Stereocenters           & No  & $\leq 4$                & Caps total stereocenter count  \\
            Halogenicity            & No  & F 6, Cl 3, Br 3         & Caps halogenated structures \\
            Symmetry                & No  & $\leq 0.8$              & Graph-automorphism symmetry score \\
            Other Common Alerts     & No  & Full collection         & Non-PAINS alert families are reported, including Dundee, BMS, and related sets \\
            \botrule
        \end{tabularx}
    \end{table}

To examine why diagnostic alert matches should not automatically be interpreted as exclusion criteria, we applied all $23$ Common~Alerts rule sets to oral FDA-approved drugs, $665$ of $673$ compounds matched at least one pattern, and every $629$ KRAS~G12D ligand matched at least one SMARTS pattern from the Genotoxic-Carcinogenicity collection. 

Among the enforced structural modules, Lilly~MedChem~Rules~\citep{bruns2012rules} made the largest contribution to attrition. We examined whether attrition produced by the Lilly~MedChem~Rules was determined primarily by categorical hard-reject rules or by the numerical demerit cutoff. Across the five reference subsets, immediate hard rejects accounted for $82.50-91.95\%$ of all Lilly exclusions (Table~\ref{tab:lilly_hard_rejects_reference}). The most frequent hard-reject categories included Michael acceptors, hydrazones, catechols, reactive carbonyl-containing motifs, and related structural liabilities. 

\begin{table}[ht!]
    \centering
    \caption{Contribution of immediate hard rejects to Lilly structural filters exclusions across the five reference molecular subsets. Percentages are calculated relative to all Lilly exclusions within each dataset}
    \label{tab:lilly_hard_rejects_reference}
    \resizebox{\columnwidth}{!}{%
    \begin{tabular}{lrrrrr}
        \toprule
                      & \textbf{ZINC-250k} & \textbf{ChEMBL~34}  & \textbf{MOSES}     & \textbf{PubChem}    & \textbf{GuacaMol} \\
        \midrule
        Lilly exclusions       & $6,827$   & $11,106$            & $1,876$            & $12,548$            & $13,456$ \\
        Immediate hard rejects & $6,017$   & $10,209$            & $1,725$            & $10,352$            & $11,687$ \\
        Hard rejects, \%       & $88.14$   & $91.92$             & $91.95$            & $82.50$             & $86.85$  \\
        \botrule
    \end{tabular}%
    }
\end{table}

As an additional positive control, we evaluated $673$ confirmed oral FDA-approved drugs. Of these compounds, $486$ passed the physicochemical descriptor stage and $368$ remained after the complete structural filters stage. The largest contribution to structural exclusions came from the Lilly and NIBR filters. In particular, $524$ out of $673$ ($77.86\%$) compounds passed the Lilly configuration, and only $22$ of $149$ Lilly exclusions resulted from accumulated demerits reaching the cutoff, whereas $127$ were rejected by native hard structural rules. Increasing the Lilly demerit cutoff from $160$ to $180$ changed the Lilly pass count only from $524$ ($77.86\%$) to $527$ ($78.31\%$), while the number of compounds retained after the complete structural filter stage remained unchanged at $368$.

% SECTION 1.4
\subsection{Synthesis feasibility definitions and thresholds}\label{appendix:detailed-mol-flow-retro}
Stage~4 comprises two sequential gates. First, \textsc{HEDGEHOG} calculates three synthesis-related scores. Molecules satisfying all three score thresholds are subsequently evaluated with AiZynthFinder. The default configuration includes three metrics. The Synthetic Accessibility score (SA score)~\citep{ertl2009estimation} is a fragment- and complexity-based heuristic that ranges molecules from one, for molecules expected to be easy to synthesize, to 10, indicating molecules that are difficult to synthesize. The cutoff of $4.5$ was selected based reported optimal SA score thresholds of $4.5$ and $4.4$~\citep{vorvsilak2020syba}. This setting is also consistent with the observation that most bioactive molecules have SA scores between $2.5$ and $4.5$~\citep{imrie2021generating}. Moreover, evaluated that $97.5\%$ of the full ZINC-250k dataset has SA score $ \leq 5$\citep{lee2023exploring}, and the corresponding proportion in our ZINC-250k subset was $97.37\%$. Independent evaluation of SA-score cutoffs on the reference molecular subsets is reported in Table~\ref{tab:analysis_sascore}.

\begin{table}[ht!]
    \centering
    \caption{Independent retention of reference molecular subsets at different SA-score cutoffs}
    \label{tab:analysis_sascore}
    \resizebox{\columnwidth}{!}{%
    \begin{tabular}{lcccc}
        \toprule
        \textbf{SA-score cutoff} & \textbf{ZINC-250k}   & \textbf{ChEMBL~34}   & \textbf{MOSES}       & \textbf{FDA-approved} \\
        \midrule
        $\leq 4.0$               & $42,676$ ($85.35\%$) & $46,880$ ($93.76\%$) & $49,684$ ($99.37\%$) & $542$ ($80.53\%$)    \\
        $\leq 4.5$               & $46,825$ ($93.65\%$) & $48,541$ ($97.08\%$) & $49,941$ ($99.88\%$) & $596$ ($88.56\%$)    \\
        $\leq 5.0$               & $48,687$ ($97.37\%$) & $49,337$ ($98.67\%$) & $49,987$ ($99.97\%$) & $637$ ($94.65\%$)    \\
        \botrule
    \end{tabular}%
    }
\end{table}

The Retrosynthetic Accessibility score (RA score)~\citep{thakkar2021retrosynthetic} is a machine-learned classifier that predicts whether an automated retrosynthesis planner is likely to identify a route to the target. We apply the natural decision cutoff of 0.5 for the predicted probability to find a synthesis route. SYBA~\citep{vorvsilak2020syba} is a fragment-based Bernoulli Naive Bayes classifier. Positive values indicate molecules expected to be easier to synthesize, and we therefore apply a threshold of 0.

The synthesis score thresholds were not adapted to KRAS~G12D inhibitors. Only $2$ of the $126$ holdout ligands ($1.59\%$) satisfied the SA, RA, and SYBA criteria simultaneously. The dominant score-level bottleneck was the SA score. The median was $5.23$ for the KRAS~G12D reference ligands, compared with $2.95$ for the oral FDA-approved drugs, $2.90$ for ZINC-250k, and $2.65$ for ChEMBL~34. The corresponding median molecular weights were $554.21$, $333.12$, $335.10$, and $362.17$~Da, respectively. SA score incorporates fragment contributions and structural complexity penalties. Therefore, the higher scores of the KRAS ligands are consistent with their substantially larger molecular size. We retained the fixed cutoff rather than adapting it to KRAS because it represents an explicit synthetic accessibility constraint.

AiZynthFinder was run using the USPTO expansion and filtering policies and a ZINC-derived building-block stock~\citep{sterling2015zinc}. Stock membership was determined by exact full-InChIKey matching. Consequently, charged and neutral forms of the same molecular graph were treated as distinct compounds. For each molecule, we searched the input representation and, where a valid structure could be generated, a stereochemistry-preserving neutralized representation as independent targets. Among the $629$ KRAS~G12D reference ligands, $285$ ($45.31\%$) were neutral and $344$ ($54.69\%$) carried a non-zero formal charge, including $334$ negatively and $10$ positively charged compounds. A stereochemistry-preserving neutralized representation was successfully generated for $343$ of the $344$ charged ligands. Consequently, the AiZynthFinder analysis comprised $972$ independently searched molecular representations with $629$ original inputs and $343$ additional neutralized variants.

\begin{figure}[!htbp]
    \centering
     \includegraphics[width=\linewidth]{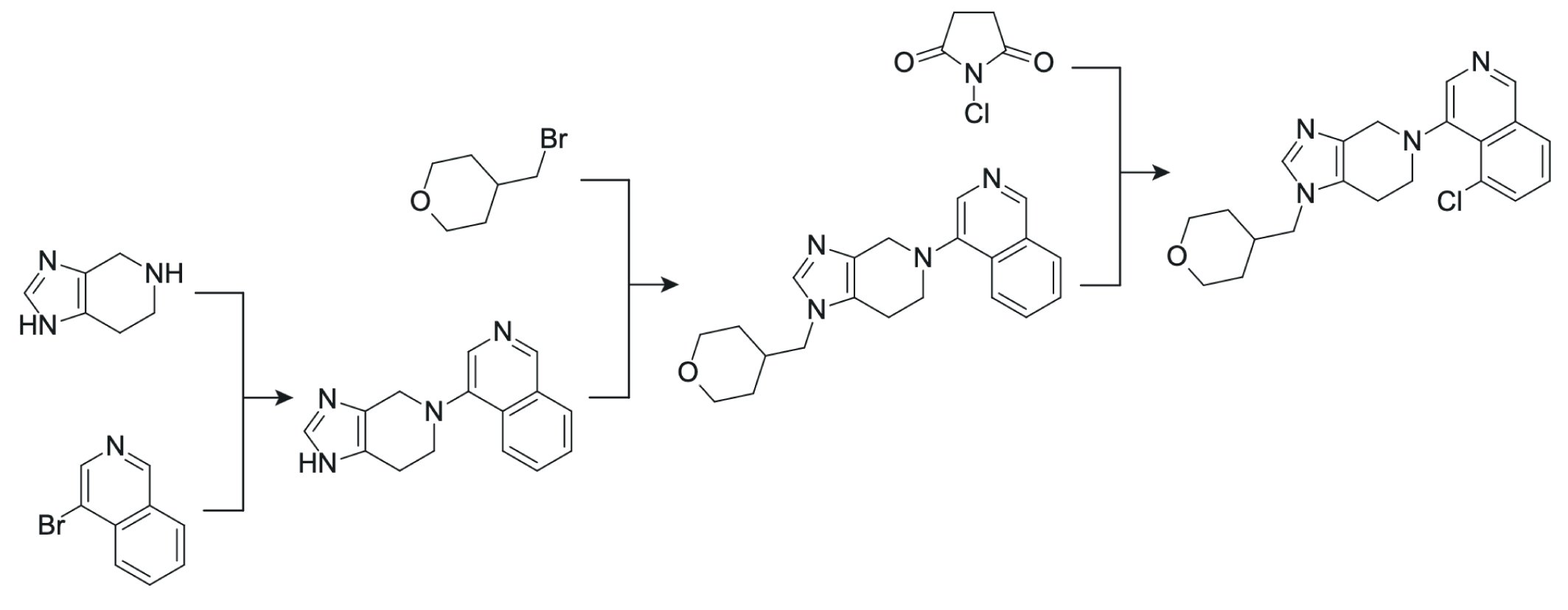}
     \caption{Example retrosynthetic route proposed by AiZynthFinder under the search configuration used in this work.}
     \label{fig-app-aizynthfinder-route}
\end{figure}

The default \textsc{HEDGEHOG} configuration used a maximum depth of $5$, a limit of $300$~seconds per molecule, and a maximum of $300$ iterations. We also ran expanded searches directly on all 629 KRAS ligands, with depth up to 15, time up to 20 minutes, and up to 4,000 iterations. A route was found for $230$ ligands ($36.6\%$) of KRAS~G12D reference compounds, and every successful route came from a neutral structure. Fig.~\ref{fig-app-aizynthfinder-route} shows an example retrosynthetic route proposed by AiZynthFinder under the search configuration used in this work. Our analysis showed that route completion depends on coverage of the selected ZINC stock. Unresolved routes frequently terminated at chemically simple precursors that were not recognized as available stock compounds, including water, hydrogen bromide, molecular bromine, and recurrent small organic intermediates (Table~\ref{tab:terminal_precursors}). 

\begin{table}[ht!]
    \centering
    \caption{Most frequent terminal precursors absent from the ZINC stock in unresolved expanded AiZynthFinder searches for the KRAS~G12D reference set}
    \label{tab:terminal_precursors}
    \resizebox{\columnwidth}{!}{%
    \begin{tabular}{lcc}
        \toprule
        \textbf{Terminal Precursor}      & \textbf{InChIKey}           & \textbf{Frequency} \\
        \midrule
        Br, Hydrogen bromide             & CPELXLSAUQHCOX-UHFFFAOYSA-N & $60$      \\
        COC(=O)c1cncc(F)c1N              & MRNNKEIHDABMRC-UHFFFAOYSA-N & $27$      \\
        CCOC(=O)[C@@]12CCCN1C[C@@H](O)C2 & BZISGTBTWVLKJE-WPRPVWTQSA-N & $18$      \\
        O, Water                         & XLYOFNOQVPJJNP-UHFFFAOYSA-N & $9$       \\
        BrBr, Bromine                    & GDTBXPJZTBHREO-UHFFFAOYSA-N & $4$       \\
        \botrule
    \end{tabular}%
    }
\end{table}

% SECTION 1.5
\subsection{Docking definitions and thresholds}\label{appendix:detailed-mol-flow-docking}
Stage~5 docks molecules with \textit{smina}, GNINA, and Matcha. Scores are calculated for every molecule that reaches this stage. For the main KRAS~G12D comparison, the docking region was defined from the 05C crystal ligand in PDB~7EW9. The \textit{smina} and GNINA autobox used 4~\AA{} padding. Each of the three methods used the generic cutoff of $-6.5$~kcal/mol. For KRAS-specific calibration, method-specific thresholds were estimated jointly on the calibration subset using a $95\%$ all-method retention objective. The resulting cutoffs were $-9.01$~kcal/mol for \textit{smina}, $-10.14$~kcal/mol for GNINA, and $-9.25$~kcal/mol for Matcha, retaining $478$ of $503$ calibration ligands and $122$ of $126$ holdout ligands. Because all three reference-derived cutoffs were more stringent than the generic cutoff, the operational cutoff remained $-6.5$~kcal/mol for each method. For each molecule, the best score returned by each method was retained. A molecule passed Stage~5 only if scores from all three methods were available and each score satisfied its method-specific cutoff, while a missing score was treated as a failure.

The three docking results are related. \textit{smina} is a Vina-family empirical method, and GNINA adds CNN scoring to a related framework. Matcha does not use the Vina, smina, or GNINA search procedure for pose generation. The ligand poses are generated by a three stage neural flow-matching model, which predicts ligand translation, rotation, and torsional degrees of freedom. GNINA is used only afterwards, for pose rescoring and, in our setup, for minimization and ranking of the generated poses. Importantly, this scoring step is not an intrinsic part of the Matcha pose generator and can be replaced by another scoring function or disabled.

\begin{figure}[!htbp]
    \centering
    \includegraphics[width=\linewidth]{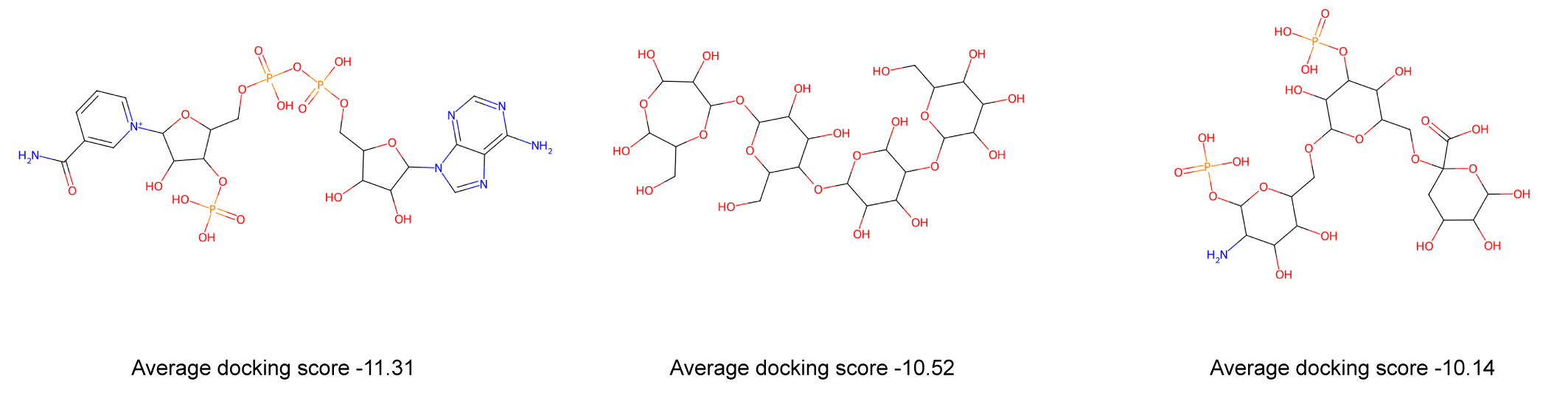}
     \caption{Examples of docking-score hacked molecules. 
     Average docking score is the mean of three docking tools, evaluated in the \textsc{HEDGEHOG} pipeline.}
     \label{fig:docking-hacked-mols}
\end{figure}

Some models appear to optimize docking scores at the expense of chemical plausibility, resulting in favorable docking scores but chemically poor molecules. To illustrate this, we sampled three illustrative molecules. These molecules passed Stage~1 only and were docked into the KRAS~G12D switch-II pocket under the same conditions as in the main pipeline. We report average docking score as the unweighted arithmetic mean of the best docking scores returned by \textit{smina}, GNINA, and Matcha. Figure~\ref{fig:docking-hacked-mols} shows that these molecules satisfy the docking scoring function while violating basic medicinal chemistry or synthesizability constraints, the SMILES structures from left to right are the followig:
\resizebox{\linewidth}{!}{\texttt{NC(=O)c1ccc[n+](C2OC(COP(=O)(O)OP(=O)(O)OCC3OC(n4cnc5c(N)ncnc54)C(O)C3O)C(OP(=O)(O)O)C2O)c1}},
\resizebox{\linewidth}{!}{\texttt{OCC1OC(OC2OC(CO)C(OC3OC(O)C(OC4OC(CO)C(O)C(O)C4O)C(O)C3O)C(O)C2O)C(O)C(O)OC1O}},
\resizebox{\linewidth}{!}{\texttt{NC1C(OP(=O)(O)O)OC(COC2OC(COC3(C(=O)O)CC(O)C(O)C(O)O3)C(O)C(OP(=O)(O)O)C2O)C(O)C1O}}.

These highlights that a model may appear strong under a docking-only metric by generating high scoring but chemically implausible candidates.

% SECTION 1.6
\subsection{Three-dimensional filtration definitions and thresholds}\label{appendix:detailed-mol-flow-3dfilters}

Stage~6 checks whether the pose is inside the search box, tests geometry with PoseCheck Fast~\citep{nikolenko2026posecheckfast}, calculates conformer deviation, and checks target-specific interactions. The source of the coordinates used at Stage~6 depends on the generator output. For SMILES and 2D-generator inputs, Stage~6 evaluates coordinates produced by an enabled docking stage. For 3D-generator inputs supplied as SDF files, the original generated coordinates are used as the primary coordinate source. Evaluating an additional set of redocked coordinates is optional and requires at least one configured docking tool.

To evaluate the effect of replacing native generated poses with redocked poses, we applied Stage~6 independently to the native coordinates and to the corresponding Matcha redocked coordinates of the $500$ molecules sampled from 5 models that natually outputs 3D coordinates. For each molecule, the lowest-scoring Matcha pose was selected as the redocked pose. The KRAS-specific Asp12 interaction requirement was enabled in both evaluations.

\begin{table}[ht!]
    \caption{Stage~6 pass counts for native generated poses and corresponding Matcha-redocked poses. Each model contributed $100$ sampled molecules}
    \label{tab:docking_native_best_res}
    \centering
    \resizebox{\columnwidth}{!}{%
        \begin{tabular}{lccccc}
            \toprule
            \textbf{Model}    & \textbf{DiffSBDD} & \textbf{DrugFlow} & \textbf{Pocket2Mol} & \textbf{ResGen} & \textbf{TargetDiff} \\
            \midrule
            Stage~6, native   & $4$ ($4\%$)       & $68$ ($68\%$)     & $18$ ($18\%$)       & $0$ ($0\%$)     & $38$ ($38\%$) \\
            Stage~6, redocked & $31$ ($31\%$)     & $45$ ($45\%$)     & $32$ ($32\%$)       & $8$ ($8\%$)     & $44$ ($44\%$) \\
            \botrule
        \end{tabular}%
    }
\end{table}

Absolute Stage~6 pass counts changed substantially for several models, but the ordinal ordering remained DrugFlow $>$ TargetDiff $>$ Pocket2Mol $>$ DiffSBDD $>$ ResGen under both coordinate protocols. Native and Matcha-redocked conformations produced different Stage~6 pass counts. Therefore, the default workflow uses the native coordinates of 3D generators as the primary Stage~6 input.

\begin{figure}[!htbp]
    \centering
     \includegraphics[width=\linewidth]{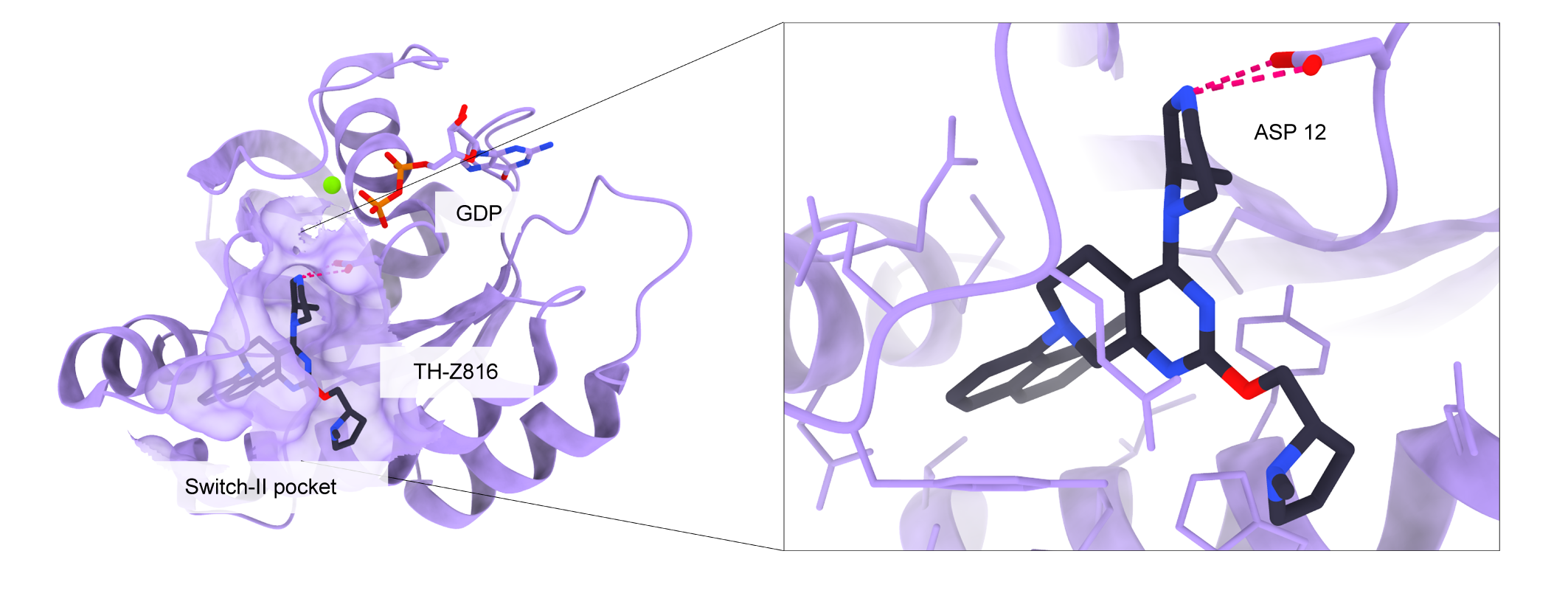}
     \caption{3D structure of KRAS-G12D in complex with inhibitor TH-Z816 (PDB ID: 7EW9). 
     \textbf{a}, KRAS-G12D bound to GDP, magnesium ion (green), and inhibitor TH-Z816 (CCD ID: 05C, dark grey) 
     in the switch-II pocket displayed as a surface. \textbf{b}, Key interaction between Asp 12 of KRAS-G12D and the piperazine moiety 
     of the inhibitor is shown with pink dashed lines. Side chains of residues within 5~\AA{} of the ligand are displayed.}
     \label{fig:asp12}
\end{figure}
    
% SECTION 1.7
\subsection{Per-stage runtime}\label{appendix:runtime}
The \textsc{HEDGEHOG} workflow was implemented as a staged, parallelized screening cascade. Molecules are filtered sequentially. Therefore, progressively fewer candidates reach more computationally demanding synthesis and docking stages. We extracted the wall-clock runtime and number of input molecules for each stage from the logs of all 66 canonical fixed-budget runs (Table~\ref{tab:timed}). Input molecules denotes the number of molecules entering each stage. Stages~5 and 6 were skipped in eight runs because no molecules remained after Stage~4.

\begin{table}[!htbp]
    \centering
    \caption{Recorded wall-clock runtimes across the 66 canonical fixed-budget runs. Values are mean $\pm$ sample standard deviation across runs in which the corresponding stage was executed}
    \label{tab:timed}
    \resizebox{\columnwidth}{!}{%
    \small
        \begin{tabular}{lccc}
        \toprule
        \textbf{Stage}          & \textbf{Input molecules} & \textbf{Time, s}    & \textbf{Median [IQR], s} \\
        \midrule
        Stage~1: Preprocessing         & $765.2 \pm 294.9$ & $3.2 \pm 0.9$       & $2.9$ [$2.5$--$3.9$] \\
        Stage~2: Descriptors           & $715.6 \pm 299.5$ & $12.3 \pm 2.8$      & $11.5$ [$10.9$--$12.4$] \\
        Stage~3: Structural Filters    & $571.7 \pm 299.7$ & $38.4 \pm 8.7$      & $37.8$ [$32.7$--$43.2$] \\
        Stage~4: Synthesis Feasibility & $357.5 \pm 250.2$ & $240.7 \pm 168.6$   & $217.1$ [$118.7$--$291.4$] \\
        Stage~5: Docking               & $179.7 \pm 205.8$ & $2270.8 \pm 5223.7$ & $717.4$ [$163.1$--$1957.6$] \\
        Stage~6: 3D Filters            & $124.6 \pm 135.5$ & $26.6 \pm 12.1$     & $25.2$ [$22.5$--$27.5$] \\
        \botrule
    \end{tabular}
    }
\end{table}

% SECTION 1.8
\subsection{Models overview}\label{appendix:models-overview}
We group molecular generators by strategy and architecture. These choices affect string validity, grammar errors, 3D geometry, and pocket alignment~\citep{david2020molecular,bilodeau2022generative}. Table~\ref{tab:generation-methods} lists the groups.

\begin{table}[!htbp]
    \small
    \caption{Taxonomy of molecular generators considered in our benchmark, by architecture (rows) and generative strategy (columns)}
    \label{tab:generation-methods}
    \begin{tabular}{@{}
      >{\raggedright\arraybackslash}p{0.24\linewidth}
      >{\raggedright\arraybackslash}p{0.20\linewidth}
      >{\raggedright\arraybackslash}p{0.20\linewidth}
      >{\raggedright\arraybackslash}p{0.25\linewidth}
  @{}}
    \toprule
    \textbf{Architecture / model class} & \textbf{Unconditional}  & \textbf{Ligand-based} & \textbf{Protein-based} \\
    \midrule

    Genetic algorithm                  & \cdash                   & MolFinder             & \cdash \\ \\

    Variational autoencoder            & HierGraphVAE \newline
                                         JT-VAE \newline
                                         MoLeR                     & GENTRL               & \cdash \\ \\

    Autoregressive                     & MolGPT                    & GCPG \newline
                                                                     PGMG \newline
                                                                     REINVENT4            & Dragonfly \newline
                                                                                            Pocket2Mol \newline
                                                                                            ResGen \\ \\

    Diffusion                          & E(3)DM \newline
                                         TGM-DLM                   & \cdash               & DiffSBDD \newline
                                                                                            ProtoBind-Diff\newline
                                                                                            TargetDiff \\ \\

    Flow matching                      & \cdash                    & \cdash               & DrugFlow \\
    \botrule
    \end{tabular}
\end{table}

\textbf{Genetic algorithm.} GA is a heuristic optimizer that evolves molecules via crossover and mutation operations. MolFinder~\citep{kwon2021molfinder} applies conformational space annealing to SMILES~\citep{weininger1988smiles} and requires no generative model pretraining.

\textbf{Variational autoencoder.} VAE models learn a latent distribution over chemical space with an encoder–decoder pair optimized through the evidence lower bound~\citep{kingma2013auto}. JT-VAE~\citep{jin2018junction}, HierGraphVAE~\citep{jin2020hierarchical}, and MoLeR~\citep{maziarz2021learning} operate on graphs with scaffold-aware decoders. GENTRL~\citep{zhavoronkov2019deep} is a VAE with a structured latent prior and reinforcement-learning fine-tuning.

\textbf{Autoregressive models.} String-based autoregressive models factorize sequence likelihood as $\prod_i P\left(t_i | t_{< i}\right)$. MolGPT~\citep{bagal2021molgpt} is a decoder-only Transformer. REINVENT4~\citep{loeffler2024reinvent} fine-tunes a SMILES prior into an agent via policy gradient with recurrent or transformer-based prior.

In structure-based design, 3D autoregressive models condition generation on a pocket $P$. Their conditional likelihood can be written as $p_{\theta}(M \mid P)=\prod_{t=1}^{T}p_{\theta}(z_t\mid z_{<t},P)$, where each step $z_t$ adds an atom, bond, or coordinate. Pocket2Mol~\citep{peng2022pocket2mol} and ResGen~\citep{zhang2023resgen} build 3D molecules inside the pocket. Dragonfly~\citep{atz2024prospective} is an interactome-based graph-to-sequence model. It can use ligand templates or 3D binding-site information and encodes pocket geometry with SE(3)- or E(3)-equivariant encoders.

Pharmacophore-conditioned models use a condition $c$ that contains 3D interaction features and their geometry. They can also use a latent variable $z$. One form of the conditional likelihood is $p_\theta(x\mid c)=\int p_\theta(x\mid c,z)\,p(z)\,dz$. This form allows one pharmacophore to match several ligands and one ligand to match several pharmacophores. PGMG~\citep{zhu2023pharmacophore} represents a pharmacophore as a fully connected graph. GCPG~\citep{zou2025structure} is a transformer encoder–decoder whose hidden state is modulated by pharmacophore embeddings.

\textbf{Diffusion models.} Diffusion models learn to approximate the reverse process of a predefined forward noising process~\citep{ho2020denoising}. E(3)DM~\citep{hoogeboom2022equivariant} is an E(3)-equivariant model that jointly denoises atom coordinates and types. DiffSBDD~\citep{schneuing2024structure} is an SE(3)-equivariant 3D-conditional model. TargetDiff~\citep{guan20233d} conditions the diffusion process on a protein binding site. TGM-DLM~\citep{gong2024text} is a text-guided diffusion language model over SMILES token embeddings. In our benchmark, it is used in a non-target-specific prompting setting and is therefore grouped with unconditional baselines. ProtoBind-Diff~\citep{mistryukova2025protobind} is a structure-free  diffusion language model that takes a protein amino-acid sequence as a condition.

\textbf{Flow matching.} Flow-matching models learn continuous-time velocity fields that transport a base distribution to the data distribution. DrugFlow~\citep{schneuing2025multi} is a pocket-conditioned ligand generation model with flow-based sampling.

% SECTION 1.9
\subsection{Fixed-budget stage-wise evaluation}\label{appendix:fixed-budget}
Invalid and duplicate outputs were not replaced. Besides default stractural filters, and synthesis feasibility thresholds, the main setup uses the frozen KRAS-95 descriptor profile, a $-6.5$~kcal/mol cutoff for each docking method, a 3.0~\AA{} conformer-RMSD cutoff, and the KRAS-specific Asp12 contact.

\textbf{Model implementation and sampling settings.} The model implementation and sampling settings are listed in Table~\ref{tab:model-implementation}.

\begin{table}[!htbp]
\centering
\caption{Implementation and sampling settings for the controlled experiment. The exact command and configuration files will be deposited with the stage-level outputs listed in the Data availability statement}
\label{tab:model-implementation}
\resizebox{\columnwidth}{!}{%
    \begin{tabular}{lllll}
        \toprule
        \textbf{Model} & \textbf{Type} & \textbf{Checkpoint/prior}    & \textbf{Sampling} & \textbf{Device} \\
        \midrule
        EDM            & UN            & generative\_model\_ema.npy   & batch 50 & A100 40GB \\
        HierGraphVAE   & UN            & chembl-pretrained/model.ckpt & latent sample; greedy; batch 50 & H100 PCIe \\
        JT-VAE         & UN            & model.iter-4                 & prior decode; batch 1 & H100 PCIe \\
        MoLeR          & UN            & molgen\_chkpt                & 6 workers & A100 40GB \\
        MolGPT         & UN            & gua\_tpsa\_logp\_sas.pt      & TPSA 40; logP 2; SAS 2; $T=1$ & H100 PCIe \\
        REINVENT4      & UN            & reinvent.prior               & unique false; randomized SMILES & A100 40GB \\
        ShEPhERD       & UN            & mosesaq                      & batch 10; 46 atoms & A100 40GB \\
        TGM-DLM        & UN            & PLAIN ema + Phase-2          & ChEBI prompts; top-$p=1$; batch 256 & H100 PCIe \\
        GCPG           & LB            & final\_model.pth             & batch 8 & A100 40GB \\
        GENTRL         & LB            & trained                      & batch 100 & A100 40GB \\
        MolFinder      & LB            & ---                          & one crossover per draw & CPU \\
        PGMG           & LB            & chembl fold0 epoch32         & batch 8 & A100 40GB \\
        % REINVENT4 (V)  & LB & reinvent.prior & 1,000; unique false; randomized SMILES & A100 40GB \\
        REINVENT4 (P)  & LB            & mol2mol medium similarity    & multinomial; $T=1$ & A100 40GB \\
        REINVENT4 (TL) & LB            & trained                      & multinomial; $T=1$ & A100 40GB \\
        DiffSBDD       & PB            & crossdocked fullatom cond    & batch 50; all fragments & A100 40GB \\
        Dragonfly      & PB            & config 991                   & $T=0.5$ & A100 40GB \\
        Dragonfly (b)  & PB            & config 701                   & $T=0.5$ & A100 40GB \\
        DrugFlow       & PB            & drugflow.ckpt                & batch 50 & A100 40GB \\
        Pocket2Mol     & PB            & pretrained Pocket2Mol.pt     & 100/loop; beam 100; max steps 100 & CPU \\
        ProtoBind-Diff & PB            & model.ckpt                   & 10 batches of 100 & A100 40GB \\
        ResGen         & PB            & val\_172.pt                  & beam 100; max steps 80 & A100 40GB \\
        TargetDiff     & PB            & pretrained diffusion.pt      & batch 50 & A100 40GB \\
        \botrule
    \end{tabular}}
\end{table}

Stage-wise cumulative survival for these runs is reported by generator class in Tables~\ref{tab:rerun1000_un}, \ref{tab:rerun1000_lb} \ref{tab:rerun1000_pb}. Each column is one model. Counts are molecules remaining after each successive gate, not independent per-stage tests.

\begin{table*}[ht!]
    \centering
    \caption{Sequential HEDGEHOG survival of unconditional molecular generators under a fixed-budget of $1,000$ raw generation attempts per seed. Values are mean $\pm$ sample standard deviation across three seeds}
    \label{tab:rerun1000_un}
    \resizebox{\textwidth}{!}{%
    \begin{tabular}{lcccccccc}
        \toprule
        \textbf{Stage}        & \textbf{EDM}     & \textbf{HierGraphVAE} & \textbf{JT-VAE}  & \textbf{MoLeR}   & \textbf{MolGPT}   & \textbf{REINVENT4} & \textbf{ShEPhERD} & \textbf{TGM-DLM} \\
        \midrule
        Initial               & $1000.0 \pm 0.0$ & $1000.0 \pm 0.0$      & $1000.0 \pm 0.0$ & $1000.0 \pm 0.0$ & $1000.0 \pm 0.0$  & $1000.0 \pm 0.0$   & $1000.0 \pm 0.0$  & $1000.0 \pm 0.0$ \\
        Preprocessing         & $422.3 \pm 8.1$  & $996.0 \pm 0.0$       & $997.7 \pm 1.2$  & $997.0 \pm 2.0$  & $968.3 \pm 6.4$   & $988.3 \pm 2.9$    & $627.0 \pm 8.0$   & $421.3 \pm 5.0$  \\
        Descriptors           & $219.0 \pm 10.6$ & $827.7 \pm 16.3$      & $964.3 \pm 5.1$  & $803.3 \pm 1.5$  & $937.7 \pm 5.9$   & $865.0 \pm 14.0$   & $529.7 \pm 8.0$   & $140.3 \pm 7.6$  \\
        Structural Filters    & $10.3 \pm 2.5$   & $603.7 \pm 15.3$      & $715.0 \pm 14.0$ & $518.0 \pm 22.9$ & $766.7 \pm 8.1$   & $600.3 \pm 17.0$   & $333.3 \pm 11.4$  & $52.7 \pm 1.2$   \\
        Synthesis Feasibility & $0.7 \pm 0.6$    & $461.3 \pm 25.4$      & $358.0 \pm 20.0$ & $337.7 \pm 19.3$ & $731.3 \pm 7.0$   & $352.0 \pm 29.1$   & $30.7 \pm 5.1$    & $20.3 \pm 1.2$   \\
        Docking               & $0.7 \pm 0.6$    & $310.7 \pm 13.1$      & $286.3 \pm 24.4$ & $238.3 \pm 24.6$ & $468.7 \pm 15.9$  & $249.3 \pm 16.7$   & $23.7 \pm 6.4$    & $13.7 \pm 0.6$   \\
        3D Filters            & $0.3 \pm 0.6$    & $91.7 \pm 9.5$        & $74.7 \pm 12.0$  & $92.7 \pm 9.7$   & $73.0 \pm 6.6$    & $102.0 \pm 5.0$    & $4.0 \pm 1.0$     & $3.0 \pm 0.0$    \\
        \botrule
     \end{tabular}
    }
\end{table*}

\begin{table*}[ht!]
    \centering
    \caption{Sequential HEDGEHOG survival of ligand-based molecular generators under a fixed generation budget of $1,000$ raw generation attempts per seed. Values are mean $\pm$ sample standard deviation across three seeds}
    \label{tab:rerun1000_lb}
    \resizebox{\textwidth}{!}{%
    \begin{tabular}{lccccccc}
        \toprule
        \textbf{Stage}        & \textbf{GCPG}    & \textbf{GENTRL}  & \textbf{MolFinder} & \textbf{PGMG}   & \textbf{REINVENT4 (P)} & \textbf{REINVENT4 (TL)}   \\
        \midrule
        Initial               & $1000.0 \pm 0.0$ & $1000.0 \pm 0.0$ & $1000.0 \pm 0.0$   & $1000.0 \pm 0.0$ & $1000.0 \pm 0.0$      & $1000.0 \pm 0.0$ \\
        Preprocessing         & $955.7 \pm 13.1$ & $431.0 \pm 8.7$  & $891.0 \pm 38.6$   & $949.7 \pm 5.5$  & $831.0 \pm 8.7$       & $37.0 \pm 3.0$   \\
        Descriptors           & $936.3 \pm 22.0$ & $383.0 \pm 14.1$ & $616.0 \pm 30.0$   & $654.3 \pm 26.3$ & $813.7 \pm 8.1$       & $37.0 \pm 3.0$   \\
        Structural Filters    & $335.7 \pm 54.9$ & $227.7 \pm 5.1$  & $281.3 \pm 25.4$   & $367.3 \pm 49.9$ & $805.3 \pm 6.8$       & $35.0 \pm 3.0$   \\
        Synthesis Feasibility & $11.0 \pm 1.7$   & $11.7 \pm 2.5$   & $114.3 \pm 16.2$   & $97.0 \pm 5.3$   & $115.0 \pm 11.1$      & $0.0 \pm 0.0$    \\
        Docking               & $5.7 \pm 2.1$    & $9.3 \pm 2.9$    & $97.3 \pm 12.9$    & $53.7 \pm 6.7$   & $91.0 \pm 12.1$       & $0.0 \pm 0.0$    \\
        3D Filters            & $3.3 \pm 1.5$    & $3.3 \pm 1.5$    & $12.0 \pm 0.0$     & $24.3 \pm 2.3$   & $61.3 \pm 11.9$       & $0.0 \pm 0.0$    \\
        \botrule
    \end{tabular}
    }
\end{table*}

\begin{table*}[ht!]
    \centering
    \caption{Sequential HEDGEHOG survival of protein-based molecular generators under a fixed generation budget of $1,000$ raw generation attempts per seed. Values are mean $\pm$ sample standard deviation across three seeds}
    \label{tab:rerun1000_pb}
    \resizebox{\textwidth}{!}{%
    \begin{tabular}{lcccccccc}
        \toprule
        \textbf{Stage}        & \textbf{DiffSBDD} & \textbf{Dragonfly} & \textbf{Dragonfly (b)} & \textbf{DrugFlow} & \textbf{Pocket2Mol}& \textbf{ProtoBind-Diff} & \textbf{ResGen}   & \textbf{TargetDiff}\\
        \midrule
        Initial               & $1000.0 \pm 0.0$  & $1000.0 \pm 0.0$   & $1000.0 \pm 0.0$       & $1000.0 \pm 0.0$  & $1000.0 \pm 0.0$   & $1000.0 \pm 0.0$        & $1000.0 \pm 0.0$  & $1000.0 \pm 0.0$  \\
        Preprocessing         & $980.3 \pm 7.1$   & $961.0 \pm 5.6$    & $512.7 \pm 27.6$       & $832.7 \pm 13.6$  & $544.7 \pm 17.1$   & $265.7 \pm 8.7$         & $237.0 \pm 111.3$ & $896.7 \pm 9.5$   \\
        Descriptors           & $720.7 \pm 24.1$  & $797.0 \pm 13.5$   & $319.7 \pm 15.0$       & $793.7 \pm 13.2$  & $495.3 \pm 25.8$   & $171.3 \pm 11.0$        & $67.0 \pm 44.6$   & $484.7 \pm 14.7$  \\
        Structural Filters    & $243.0 \pm 12.2$  & $685.0 \pm 18.1$   & $248.3 \pm 8.7$        & $439.0 \pm 11.5$  & $310.7 \pm 55.2$   & $59.7 \pm 4.5$          & $20.7 \pm 28.0$   & $206.0 \pm 9.8$   \\
        Synthesis Feasibility & $22.7 \pm 2.1$    & $491.0 \pm 17.3$   & $53.7 \pm 2.5$         & $111.7 \pm 12.4$  & $129.3 \pm 45.2$   & $22.3 \pm 2.3$          & $0.0 \pm 0.0$     & $1.7 \pm 1.5$     \\
        Docking               & $17.7 \pm 3.8$    & $288.3 \pm 15.5$   & $22.3 \pm 3.8$         & $102.7 \pm 9.0$   & $112.0 \pm 41.6$   & $16.0 \pm 6.0$          & $0.0 \pm 0.0$     & $1.3 \pm 1.5$     \\
        3D Filters            & $0.7 \pm 1.2$     & $127.0 \pm 18.1$   & $9.7 \pm 3.1$          & $58.0 \pm 8.5$    & $31.0 \pm 11.5$    & $6.0 \pm 2.6$           & $0.0 \pm 0.0$     & $0.3 \pm 0.6$     \\
        \botrule
    \end{tabular}%
    }
\end{table*}

\textbf{Generation efficiency.} Table~\ref{tab:generation_fixed_budget} reports wall-clock time and how many of the $1,000$ requested attempts were returned, chemically valid, and unique. Failed and duplicate outputs were not replaced. Raw validity is the number of RDKit-valid molecules divided by the number of observed generation attempts. Graph uniqueness uses canonical isomeric representations: alternative SMILES of the same stereochemically defined structure were collapsed, while distinct stereoisomers remained distinct. Uniqueness is then the number of unique valid structures divided by the number of valid molecules. Several models did not return $1,000$ outputs, including ShEPhERD, Pocket2Mol, ProtoBind-Diff, ResGen, and Dragonfly~(b). Validity and uniqueness also varied. TGM-DLM returned $1,000$ outputs but only $392.3\pm3.8$ unique valid graphs, and REINVENT4~(TL) returned $1,000$ valid molecules of which only $34.0\pm2.6$ were unique. 

\begin{table*}[ht!]
    \centering
    \caption{Fixed-budget generation statistics for the evaluated molecular generators. Each run comprised 1,000 requested raw generation attempts without replacement of failed outputs. Unique valid structures were defined as distinct canonical non-isomeric molecular graphs. Values are mean $\pm$ sample standard deviation across three seeds}
    \label{tab:generation_fixed_budget}
    \resizebox{\textwidth}{!}{%
    \begin{tabular}{lccccc}
        \toprule
        Model          & Time, s                & Returned          & Valid             & Unique valid      & Graph uniqueness  \\
        \midrule
        EDM            & $9857.28 \pm 45.18$    & $1000.0 \pm 0.0$  & $437.3 \pm 11.9$  & $437.3 \pm 11.9$  & $1.0 \pm 0.0$ \\
        HierGraphVAE   & $547.96 \pm 52.97$     & $1000.0 \pm 0.0$  & $1000.0 \pm 0.0$  & $996.3 \pm 0.6$   & $0.996 \pm 0.001$ \\
        JT-VAE         & $157.32 \pm 36.00$     & $1000.0 \pm 0.0$  & $997.7 \pm 1.2$   & $997.7 \pm 1.2$   & $1.0 \pm 0.0$ \\
        MoLeR          & $132.54 \pm 1.15$      & $1000.0 \pm 0.0$  & $1000.0 \pm 0.0$  & $1000.0 \pm 0.0$  & $1.0 \pm 0.0$ \\
        MolGPT         & $6.09 \pm 0.02$        & $1000.0 \pm 0.0$  & $984.7 \pm 4.7$   & $966.7 \pm 9.2$   & $0.982 \pm 0.007$ \\
        REINVENT4      & $9.13 \pm 0.63$        & $1000.0 \pm 0.0$  & $985.0 \pm 2.6$   & $985.0 \pm 2.6$   & $1.0 \pm 0.0$ \\
        ShEPhERD       & $15108.32 \pm 188.75$  & $641.3 \pm 8.1$   & $641.3 \pm 8.1$   & $641.3 \pm 8.1$   & $1.0 \pm 0.0$ \\
        TGM-DLM        & $898.72 \pm 12.80$     & $1000.0 \pm 0.0$  & $692.3 \pm 3.1$   & $392.3 \pm 3.8$   & $0.567 \pm 0.007$ \\
        \midrule 
        GCPG           & $53.28 \pm 0.29$       & $1000.0 \pm 0.0$  & $968.0 \pm 2.6$   & $949.3 \pm 9.9$   & $0.981 \pm 0.009$ \\
        GENTRL         & $6.83 \pm 0.33$        & $1000.0 \pm 0.0$  & $380.0 \pm 5.2$   & $380.0 \pm 5.2$   & $1.0 \pm 0.0$ \\
        MolFinder      & $1.22 \pm 0.43$        & $1000.0 \pm 0.0$  & $947.3 \pm 1.5$   & $941.7 \pm 0.6$   & $0.994 \pm 0.002$ \\
        PGMG           & $77.83 \pm 1.26$       & $1000.0 \pm 0.0$  & $934.0 \pm 13.1$  & $925.0 \pm 15.4$  & $0.990 \pm 0.003$ \\
        REINVENT4 (P)  & $21.73 \pm 1.87$       & $1000.0 \pm 0.0$  & $951.0 \pm 6.2$   & $784.3 \pm 8.4$   & $0.825 \pm 0.009$ \\
        REINVENT4 (TL) & $21.35 \pm 0.55$       & $1000.0 \pm 0.0$  & $1000.0 \pm 0.0$  & $34.0 \pm 2.6$    & $0.034 \pm 0.003$ \\
        \midrule
        DiffSBDD       & $1359.09 \pm 24.95$    & $1000.0 \pm 0.0$  & $980.3 \pm 7.1$   & $980.3 \pm 7.1$   & $1.0 \pm 0.0$ \\
        Dragonfly      & $69.24 \pm 1.82$       & $994.3 \pm 2.9$   & $994.3 \pm 2.9$   & $959.7 \pm 5.7$   & $0.965 \pm 0.004$ \\
        Dragonfly (b)  & $73.24 \pm 1.86$       & $589.0 \pm 21.7$  & $589.0 \pm 21.7$  & $512.7 \pm 27.6$  & $0.870 \pm 0.015$ \\
        DrugFlow       & $1814.61 \pm 6.20$     & $1000.0 \pm 0.0$  & $832.7 \pm 13.6$  & $832.7 \pm 13.6$  & $1.0 \pm 0.0$ \\
        Pocket2Mol     & $33522.91 \pm 6920.36$ & $556.0 \pm 18.2$  & $556.0 \pm 18.2$  & $544.7 \pm 17.1$  & $0.980 \pm 0.001$ \\
        ProtoBind-Diff & $302.03 \pm 18.46$     & $266.0 \pm 8.2$   & $266.0 \pm 8.2$   & $266.0 \pm 8.2$   & $1.0 \pm 0.0$ \\
        ResGen         & $487.04 \pm 103.11$    & $237.0 \pm 111.3$ & $237.0 \pm 111.3$ & $237.0 \pm 111.3$ & $1.0 \pm 0.0$ \\
        TargetDiff     & $19599.79 \pm 117.90$  & $1000.0 \pm 0.0$  & $896.7 \pm 9.5$   & $896.7 \pm 9.5$   & $1.0 \pm 0.0$ \\
        \bottomrule
    \end{tabular}%
    }
\end{table*}

\textbf{Conventional generative metrics.} Table~\ref{tab:conventional-metrics} reports uniqueness, KRAS novelty, internal diversity, KL similarity, and FCD for the $14$ target-conditioned models, computed on the same fixed-budget valid graphs used in the main comparison. These scores describe the generated set, not survival under \textsc{HEDGEHOG}. DiffSBDD had among the largest unique sets of $980.3\pm7.1$ and the highest IntDiv$_1$ of $0.884\pm0.001$, but almost no Stage~6 survivors (Table~\ref{tab:rerun1000_pb}). Dragonfly combined a large unique set with the highest Stage~6 count. REINVENT4~(TL) had the lowest novelty and diversity and no survivors, whereas REINVENT4~(P) had the most favorable KL similarity and FCD among ligand-based models and $61.3\pm11.9$ survivors (Table~\ref{tab:rerun1000_lb}). High conventional scores therefore do not indicate which generators remain after sequential filtration.

\begin{table*}[!htbp]
\centering
\caption{Conventional generative metrics for target-conditioned generators under a fixed-budget of $1,000$ requested generation attempts per seed. Metrics were calculated on distinct RDKit-valid canonical non-isomeric molecular graphs among returned outputs. Values are mean $\pm$ sample standard deviation across three seeds}
\label{tab:conventional-metrics}
\resizebox{\textwidth}{!}{%
    \begin{tabular}{lccccc}
        \toprule
        Model          & Unique valid    & KRAS novelty    & IntDiv$_1$ & KL similarity $\uparrow$ & FCD $\downarrow$ \\
        \midrule
        GCPG           & $949.3\pm9.9$   & $1.0\pm0.0$ & $0.823\pm0.005$ & $0.159\pm0.020$ & $21.30\pm1.15$ \\
        GENTRL         & $380.0\pm5.2$   & $1.0\pm0.0$ & $0.832\pm0.000$ & $0.209\pm0.008$ & $18.00\pm0.28$ \\
        MolFinder      & $941.7\pm0.6$   & $1.0\pm0.0$ & $0.859\pm0.002$ & $0.407\pm0.016$ & $21.59\pm0.64$ \\
        PGMG           & $925.0\pm15.4$  & $1.0\pm0.0$ & $0.812\pm0.003$ & $0.577\pm0.030$ & $27.40\pm0.43$ \\
        REINVENT4 (P)  & $784.3 \pm 8.4$ & $1.0\pm0.0$ & $0.677\pm0.001$ & $0.599\pm0.019$ & $12.15\pm0.34$ \\
        REINVENT4 (TL) & $34.0 \pm 2.6$  & $0.381\pm0.033$ & $0.361\pm0.008$ & $0.330\pm0.008$ & $11.74\pm0.29$ \\
        \midrule
        DiffSBDD       & $980.3\pm7.1$   & $1.0\pm0.0$ & $0.884\pm0.001$ & $0.297\pm0.006$ & $34.09\pm0.25$ \\
        Dragonfly      & $959.7 \pm 5.7$ & $1.0\pm0.0$ & $0.844\pm0.001$ & $0.470\pm0.002$ & $38.22\pm0.24$ \\
        Dragonfly (b)  & $512.7\pm27.6$  & $1.0\pm0.0$ & $0.783\pm0.002$ & $0.195\pm0.028$ & $35.83\pm0.29$ \\
        DrugFlow       & $832.7\pm13.6$  & $1.0\pm0.0$ & $0.856\pm0.001$ & $0.286\pm0.020$ & $29.11\pm0.23$ \\
        Pocket2Mol     & $544.7\pm17.1$  & $1.0\pm0.0$ & $0.861\pm0.003$ & $0.218\pm0.016$ & $35.60\pm1.10$ \\
        ProtoBind-Diff & $266.0\pm8.2$   & $1.0\pm0.0$ & $0.831\pm0.002$ & $0.510\pm0.005$ & $28.10\pm0.32$ \\
        ResGen         & $237.0\pm111.3$ & $1.0\pm0.0$ & $0.603\pm0.072$ & $0.374\pm0.149$ & $33.65\pm6.61$ \\
        TargetDiff     & $896.7\pm9.5$   & $1.0\pm0.0$ & $0.877\pm0.001$ & $0.284\pm0.008$ & $34.55\pm0.26$ \\
        \botrule
    \end{tabular}}
\end{table*}

\textbf{Similarity to the crystallographic ligand 05C.} Protein-based generators were conditioned on the KRAS~G12D switch-II pocket from PDB~7EW9, whose bound ligand is 05C (TH-Z816). Table~\ref{tab:pb_05c_novelty} therefore reports ECFP4 Tanimoto similarity of RDKit-valid outputs to 05C, together with graph and Bemis-Murcko scaffold novelty relative to 05C, using the same fixed-budget valid outputs as in the main comparison. Mean similarity remained low ($0.085-0.124$). No model produced a molecule with $T_c\geq0.3$ to 05C, and every valid output was novel at both the molecular-graph and scaffold level.

\begin{table*}[htbp]
    \centering
    \caption{Similarity and novelty of protein-based generator outputs relative to the crystallographic ligand 05C. Metrics were calculated for all RDKit-valid outputs obtained within the first $1,000$ requested raw generation attempts per seed. Values are mean $\pm$ sample standard deviation across three runs. Scaffold-based percentages were calculated for molecules containing a Bemis-Murcko scaffold}
    \label{tab:pb_05c_novelty}
    \resizebox{\textwidth}{!}{%
    \begin{tabular}{lcccccc}
    \toprule
    \textbf{Model} & \textbf{Mean $T_c$ to 05C} & \textbf{$T_c\geq0.3$, \%} & \textbf{$T_c\geq0.4$, \%} & \textbf{Molecular novelty, \%} & \textbf{Novel-scaffold molecules, \%} & \textbf{Novel unique scaffolds, \%} \\
    \midrule
    DiffSBDD       & $0.085 \pm 0.001$ & $0.0 \pm 0.0$ & $0.0 \pm 0.0$ & $100.0 \pm 0.0$ & $100.0 \pm 0.0$ & $100.0 \pm 0.0$ \\
    Dragonfly      & $0.105 \pm 0.000$ & $0.0 \pm 0.0$ & $0.0 \pm 0.0$ & $100.0 \pm 0.0$ & $100.0 \pm 0.0$ & $100.0 \pm 0.0$ \\
    Dragonfly (b)  & $0.117 \pm 0.000$ & $0.0 \pm 0.0$ & $0.0 \pm 0.0$ & $100.0 \pm 0.0$ & $100.0 \pm 0.0$ & $100.0 \pm 0.0$ \\
    DrugFlow       & $0.112 \pm 0.001$ & $0.0 \pm 0.0$ & $0.0 \pm 0.0$ & $100.0 \pm 0.0$ & $100.0 \pm 0.0$ & $100.0 \pm 0.0$ \\
    Pocket2Mol     & $0.096 \pm 0.003$ & $0.0 \pm 0.0$ & $0.0 \pm 0.0$ & $100.0 \pm 0.0$ & $100.0 \pm 0.0$ & $100.0 \pm 0.0$ \\
    ProtoBind-Diff & $0.124 \pm 0.002$ & $0.0 \pm 0.0$ & $0.0 \pm 0.0$ & $100.0 \pm 0.0$ & $100.0 \pm 0.0$ & $100.0 \pm 0.0$ \\
    ResGen         & $0.103 \pm 0.009$ & $0.0 \pm 0.0$ & $0.0 \pm 0.0$ & $100.0 \pm 0.0$ & $100.0 \pm 0.0$ & $100.0 \pm 0.0$ \\
    TargetDiff     & $0.087 \pm 0.001$ & $0.0 \pm 0.0$ & $0.0 \pm 0.0$ & $100.0 \pm 0.0$ & $100.0 \pm 0.0$ & $100.0 \pm 0.0$ \\
    \botrule
    \end{tabular}%
    }
\end{table*}

\begin{figure*}[!ht]
    \centering
     \includegraphics[width=\textwidth]{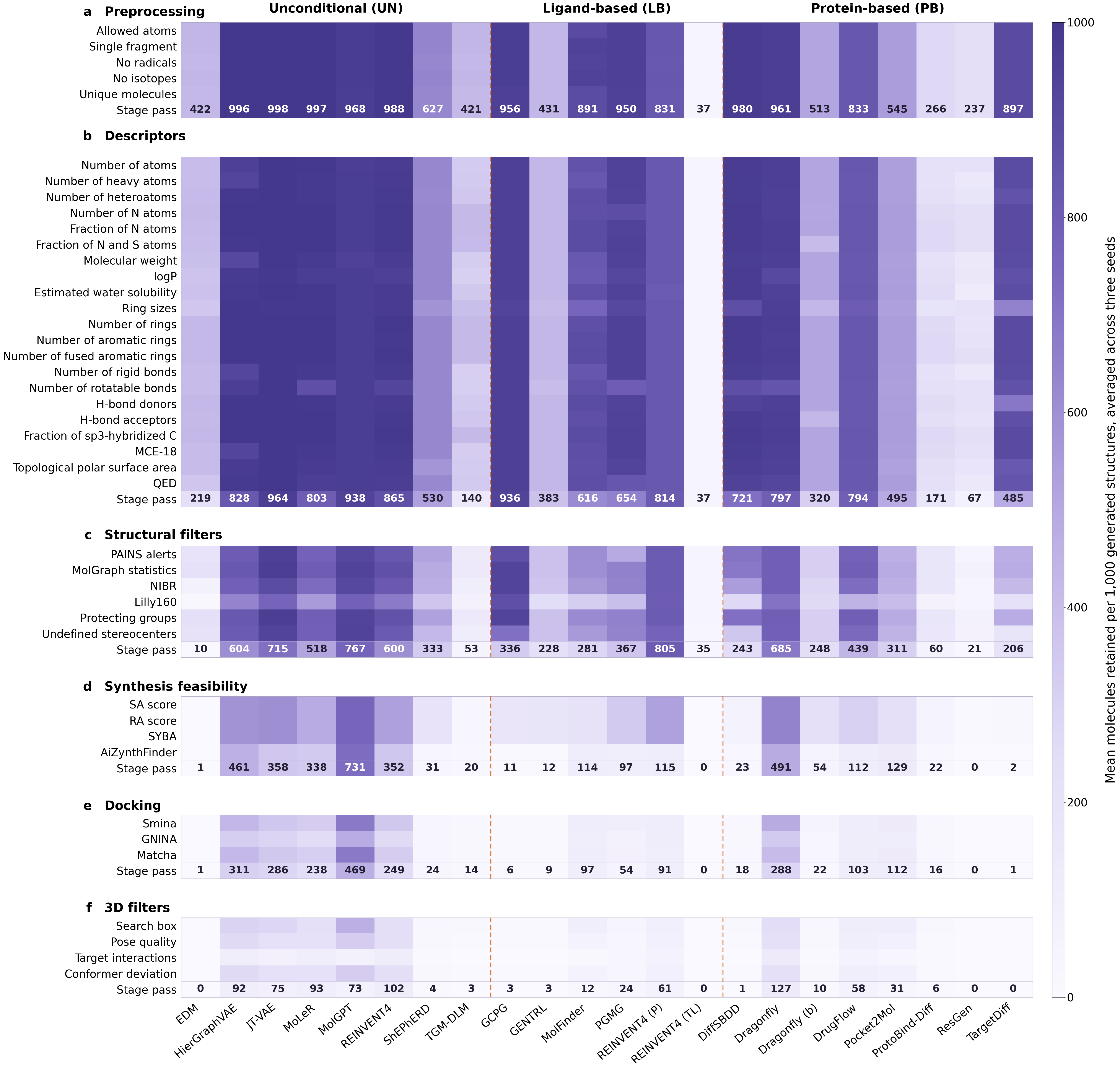}
     \caption{
        \textbf{\textsc{HEDGEHOG} molecule counts for each stage.}
        \textbf{a,} Preprocessing.
        \textbf{b,} Descriptors.
        \textbf{c,} Structural filters. 
        \textbf{d,} Synthesis feasibility.
        \textbf{e,} Docking.
        \textbf{f,} Three-dimensional filters.
        Columns represent individual generative models grouped as unconditional (UN), ligand-based (LB), or protein-based (PB). Dashed vertical lines separate the model classes. Rows represent the criteria or computational tools evaluated at each stage. Cell color indicates the mean number of molecules retained from $1,000$ generation attempts per seed, averaged across three seeds, using a common $0–1,000$ scale across all panels. The Stage pass rows show stage-level retention after all enabled criteria in the corresponding stage have been applied. Overlaid values are means rounded to the nearest whole molecule.
        }
     \label{fig:hedgehog_workflow}
\end{figure*}

\begin{figure*}[!ht]
    \centering
     \includegraphics[width=\textwidth]{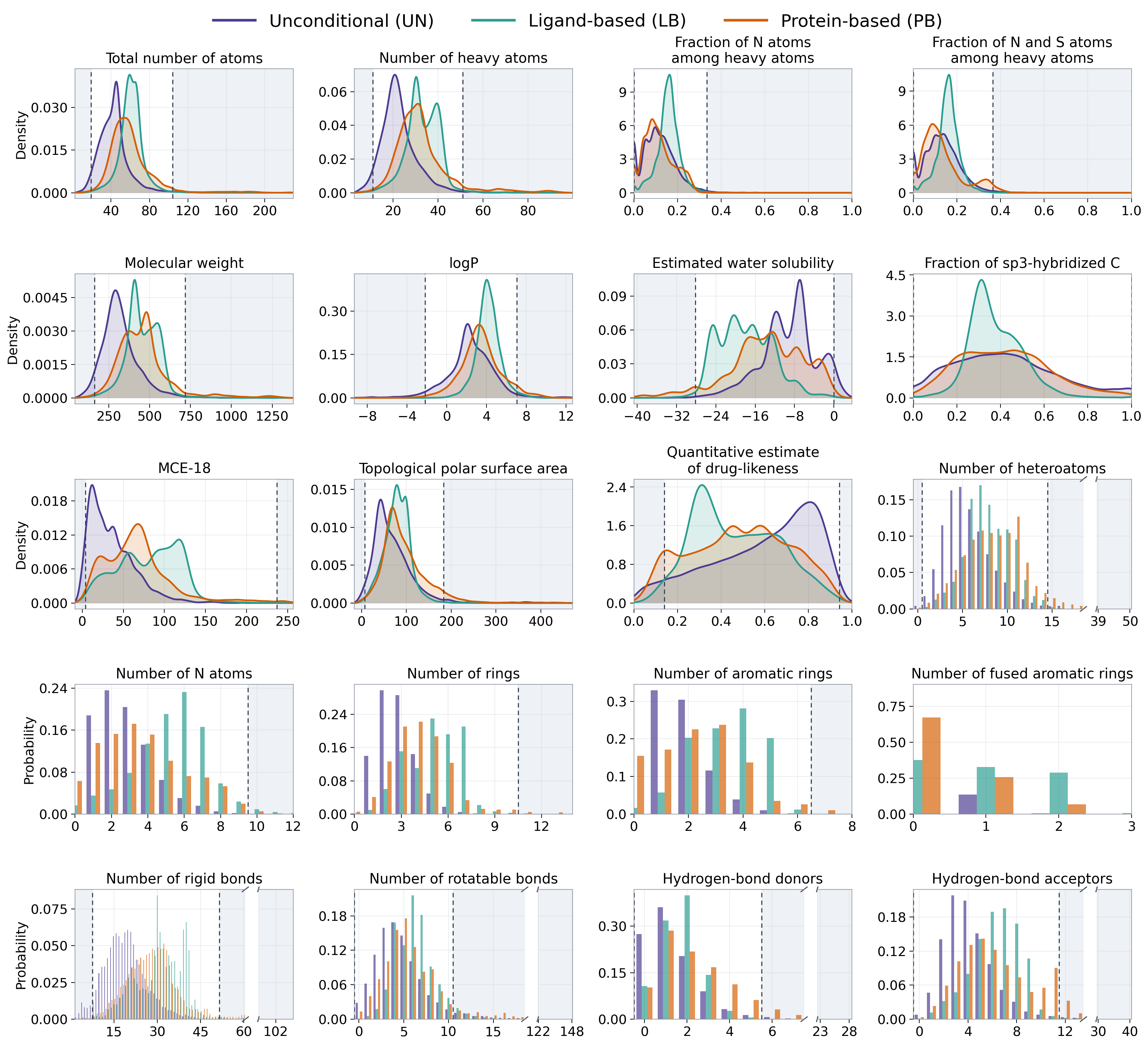}
     \caption{Physicochemical property distributions of molecules generated for KRAS. Distributions are shown for 22 models grouped into unconditional, ligand-based, and protein-based classes. Each model seed run contributes equally to its class-level distribution across three seeds. Continuous descriptors are represented by kernel density estimates, whereas integer-valued descriptors are shown as normalized frequencies. Dashed lines indicate the KRAS-95 descriptor configuration thresholds, and grey regions denote values outside these limits. Broken x axes separate the main distributions from distant extreme values.}
     \label{fig:detail-dscrip}
\end{figure*}

\textbf{Descriptor distributions under KRAS-95.} Figure~\ref{fig:hedgehog_workflow} counts how many molecules pass each descriptor filter. Figure~\ref{fig:detail-dscrip} shows the underlying property distributions for the same fixed-budget outputs, pooled by generator class. Dashed lines are the KRAS-95 limits from Table~\ref{tab-descr-thresholds-revised}. Grey regions lie outside those limits. Density inside one interval does not imply joint passage of all $21$ Stage~2 gates. Extreme tails are plotted on broken axes so that the main body of each distribution remains readable.

\clearpage

\textbf{Ligand-based novelty before and after filtration.} The main text reports pooled KRAS-relative novelty. Table~\ref{tab:lb_novelty_before_after_hedgehog} splits the same fixed-budget ligand-based runs by model and compares Stage~1 outputs with Stage~6 survivors. $T_c^{\mathrm{NN}}$ is ECFP4 Tanimoto similarity to the nearest KRAS~G12D reference ligand. GCPG, GENTRL, and PGMG remained far from the reference set at both time points. MolFinder was closer after Stage~1 ($16.63\pm1.07\%$ with $T_c^{\mathrm{NN}}\geq0.4$) but had no such molecules among survivors. REINVENT4~(TL) largely reproduced KRAS chemistry (mean $T_c^{\mathrm{NN}}=0.9378\pm0.0073$) and produced no Stage~6 survivors. REINVENT4~(P) accounted for all remaining ligand-based molecules with $T_c^{\mathrm{NN}}\geq0.4$ after Stage~6, and even for that model the fraction fell from $71.56\pm0.47\%$ after Stage~1 to $35.53\pm3.27\%$ among survivors. All Stage~6 survivors from models that reached this stage were novel at the graph and scaffold level.

\begin{table*}[htbp]
    \centering
    \caption{KRAS~G12D-reference novelty after Stage~1 and among final Stage~6 survivors for individual ligand-based models in the controlled fixed-budget experiment. Counts and metrics are mean $\pm$ sample standard deviation across three runs of $1,000$ requested raw generation attempts. Final novelty metrics are undefined for REINVENT4 (TL), which produced no Stage~6 survivors}
    \label{tab:lb_novelty_before_after_hedgehog}
    \resizebox{\textwidth}{!}{%
    \begin{tabular}{lcccccc}
    \toprule
    Metric & GCPG & GENTRL & MolFinder & PGMG & REINVENT4 (P) & REINVENT4 (TL) \\
    \midrule
    Stage~1 molecules, $N$ & $955.7 \pm 13.1$ & $431.0 \pm 8.7$ & $891.0 \pm 38.6$ & $949.7 \pm 5.5$ & $831.0 \pm 8.7$ & $37.0 \pm 3.0$ \\
    Stage~1 retention, \% & $95.57 \pm 1.31$ & $43.10 \pm 0.87$ & $89.10 \pm 3.86$ & $94.97 \pm 0.55$ & $83.10 \pm 0.87$ & $3.70 \pm 0.30$ \\
    Stage~6 survivors, $N$ & $3.3 \pm 1.5$ & $3.3 \pm 1.5$ & $12.0 \pm 0.0$ & $24.3 \pm 2.3$ & $61.3 \pm 11.9$ & $0.0 \pm 0.0$ \\
    Stage~6 relative to raw-attempt budget, \% & $0.33 \pm 0.15$ & $0.33 \pm 0.15$ & $1.20 \pm 0.00$ & $2.43 \pm 0.23$ & $6.13 \pm 1.19$ & $0.00 \pm 0.00$ \\
    Mean $T_c^{\mathrm{NN}}$ after Stage~1 & $0.1827 \pm 0.0071$ & $0.2302 \pm 0.0053$ & $0.2749 \pm 0.0061$ & $0.1820 \pm 0.0047$ & $0.4525 \pm 0.0012$ & $0.9378 \pm 0.0073$ \\
    Mean $T_c^{\mathrm{NN}}$ after Stage~6 & $0.2030 \pm 0.0228$ & $0.2394 \pm 0.0227$ & $0.2084 \pm 0.0105$ & $0.1975 \pm 0.0096$ & $0.3884 \pm 0.0052$ & $\mathrm{NA}$ \\
    $T_c^{\mathrm{NN}}\geq0.4$ after Stage~1, \% & $0.00 \pm 0.00$ & $1.94 \pm 0.69$ & $16.63 \pm 1.07$ & $0.00 \pm 0.00$ & $71.56 \pm 0.47$ & $100.00 \pm 0.00$ \\
    $T_c^{\mathrm{NN}}\geq0.4$ after Stage~6, \% & $0.00 \pm 0.00$ & $0.00 \pm 0.00$ & $0.00 \pm 0.00$ & $0.00 \pm 0.00$ & $35.53 \pm 3.27$ & $\mathrm{NA}$ \\
    Final molecular novelty, \% & $100.00 \pm 0.00$ & $100.00 \pm 0.00$ & $100.00 \pm 0.00$ & $100.00 \pm 0.00$ & $100.00 \pm 0.00$ & $\mathrm{NA}$ \\
    Final scaffold novelty, \% & $100.00 \pm 0.00$ & $100.00 \pm 0.00$ & $100.00 \pm 0.00$ & $100.00 \pm 0.00$ & $100.00 \pm 0.00$ & $\mathrm{NA}$ \\
    \botrule
    \end{tabular}%
    }
\end{table*}

% SECTION 3
\section{KRAS G12D benchmark instance}\label{appendix:kras-instance}

We prepared the KRAS~G12D receptor from PDB 7EW9 (pdb\_00007ew9~\citep{pdb7EW9}). We removed TH-Z816, GDP, Mg$^{2+}$, and crystallographic waters. The original PDB residue numbers were kept, and Asp12 was used for the case-study interaction rule. The receptor was used for \textit{smina}, GNINA, Matcha, and the Stage~6 checks. Boltz-2 is optional and does not affect primary survival.

We used KRAS G12D as the benchmark target for HEDGEHOG. KRAS G12D is the most prevalent KRAS oncogenic mutation~\citep{mao2022kras}, and for many years was considered undruggable. Unlike KRAS G12C, it lacks the Cys12 required for covalent binding, but at the same time the negatively charged Asp12 can be targeted via salt-bridge interaction, as demonstrated in the development of TH-Z816~\citep{kessler2019drugging}. Given its challenging nature and clinical significance, we considered KRAS G12D a relevant and mechanistically informative benchmark target for the HEDGEHOG evaluation pipeline~\citep{mao2022kras}.

For unconditional models, KRAS G12D is used for docking and post-docking 3D checks. For protein-based models, it also provides the target-conditioning structure. For ligand-based models, we used KRAS G12D inhibitors from~\citep{ghazi2025quantum}.

\section{REINVENT4 configurations}\label{appendix:reinvent4}
We evaluated REINVENT4 in one \textit{de novo} configuration and two ligand-conditioned configurations. REINVENT4 was used for this analysis because the same framework provides public priors, \textit{de novo} SMILES generation, molecule-based generation, and transfer-learning workflows. This allows prior choice to be varied while keeping the downstream HEDGEHOG evaluation fixed, isolating how prior choice and transfer-learning within models affect downstream utility. For the ligand-conditioned large-output runs, we used beam search with temperature set to $1.0$ and number of SMILES per input SMILES set to $500$. All REINVENT4 configurations were evaluated on one NVIDIA A100 40GB GPU. Prior checkpoints are listed in~\citep{loffler2025reinvent4priors} and are open-source.

The unconditional baseline was denoted as REINVENT4 (checkpoint, \texttt{reinvent.prior}) in sampling mode. No KRAS ligand SMILES were supplied to this run. REINVENT4 (P) used the provided \texttt{mol2mol\_medium\_similarity.prior}, trained with a medium Tanimoto-similarity objective. This public prior was trained on PubChem-derived molecular pairs~\citep{kim2023pubchem}. To avoid data leakage, we removed KRAS seed ligands with exact PubChem matches before sampling. This left $583$ seed molecules. For REINVENT4 (TL), we fine-tuned \texttt{reinvent.prior} on the same $629$ valid KRAS~G12D entries with the Mol2Mol pipeline. We used a similarity limit of $0.7$ and trained for $1,000$ epochs over $72$ hours. We selected the checkpoint with the lowest validation loss.

% SECTION 5
\section{Validation and controlled experiments}\label{appendix:validation}

% SECTION 5.1
\subsection{Reference-collection and target validation}\label{appendix:ref-validation}
Curated set of experimentally confirmed inactive compounds or decoys is not available for KRAS G12D. Because of that, we performed an additional evaluation using five targets from the DUD-E database~\citep{mysinger2012directory}, for which both experimentally active compounds and property matched decoys are available. For each target, the descriptors and docking thresholds were evaluated using active compounds and calibration technique described in Section. We report active (A) and decoy (D) retention rates in Table~\ref{tab:other_target_filtering_AD}. 

\begin{table}[ht!]
  \centering
  \caption{Sequential stage survival of DUD-E active compounds (A) and matched decoys (D) through the HEDGEHOG pipeline. Counts at each stage are conditional on survival of all preceding stages}
  \label{tab:other_target_filtering_AD}
  \resizebox{\columnwidth}{!}{%
  \begin{tabular}{lccccc}
      \toprule
      Stage                 & AMPC, A/D     & CXCR4, A/D    & INHA, A/D     & KITH, A/D     & PUR2, A/D \\
      \midrule
      Initial               & $48$ / $2847$ & $40$ / $3406$ & $43$ / $2300$ & $57$ / $2850$ & $50$ / $2694$ \\
      Preprocessing         & $48$ / $2361$ & $40$ / $1736$ & $43$ / $1286$ & $57$ / $2831$ & $50$ / $2306$ \\
      Descriptors           & $46$ / $672$  & $39$ / $80$   & $41$ / $341$  & $55$ / $1317$ & $49$ / $0$    \\
      Structural Filters    & $33$ / $456$  & $19$ / $54$   & $35$ / $224$  & $42$ / $938$  & $41$ / $0$    \\
      Synthesis Feasibility & $30$ / $429$  & $19$ / $34$   & $35$ / $208$  & $18$ / $868$  & $15$ / $0$    \\
      Docking               & $29$ / $325$  & $19$ / $28$   & $33$ / $175$  & $17$ / $531$  & $14$ / $0$    \\
      3D Filters            & $28$ / $277$  & $16$ / $8$    & $19$ / $109$  & $3$ / $24$    & $10$ / $0$    \\
      \botrule
  \end{tabular}%
  }
\end{table}

We additionally evaluated the target-aware docking gate independently on the complete active and decoy sets, because the target-agnostic stages are not intended to discriminate actives from decoys. Docking scores were aggregated at the source compound level using the all-pass rule with the respective cutoffs. Compounds lacking a score for any method were classified as failing. Sensitivity (S), specificity (Sp), false-negative rate (FNR), and enrichment factors (EF) for the docking stage are reported in Table~\ref{tab:targets_comparison}.

\begin{table}[ht!]
  \caption{Active-decoy discrimination by the target-aware docking score gate for five DUD-E targets}
  \label{tab:targets_comparison}
  \centering
  \resizebox{\columnwidth}{!}{%
  \begin{tabular}{lcccccc}
      \toprule
      Target & active & decoy & S, \%   & Sp, \%  & FNR, \% & EF  \\
      \midrule
      AMPC   & 48     & 2847  & $95.83$ & $25.85$ & $4.17$  & $1.29$ \\
      CXCR4  & 40     & 3406  & $92.50$ & $39.46$ & $7.50$  & $1.52$ \\
      INHA   & 43     & 2300  & $95.35$ & $37.35$ & $4.65$  & $1.51$ \\
      KITH   & 57     & 2850  & $92.98$ & $48.98$ & $7.02$  & $1.79$ \\
      PUR2   & 50     & 2694  & $96.00$ & $46.92$ & $4.00$  & $1.78$ \\
      \botrule
  \end{tabular}%
  }
\end{table}

The combined docking stage retained $92.50–96.00\%$ of known actives across the five targets, while specificity ranged from $25.85\%$ to $48.98\%$. Enrichment factors were modest, ranging from $1.29$ to $1.79$. The reference-based calibration is deliberately biased toward preserving sensitivity, with the generic docking threshold relaxed when necessary to retain the predefined fraction of known target ligands.

We additionally used the 629 KRAS~G12D reference ligands as an unrelated target control. KRAS~G12D inhibitors were evaluated against each of the five DUD-E targets using the corresponding target-aware docking and 3D filters. Table~\ref{tab:kras-unrelated-targets} shows how frequently compounds known for one target remain compatible with the target-specific criteria of an unrelated binding site.

\begin{table}[!htbp]
    \centering
    \caption{Retention of 629 known KRAS~G12D ligands with five unrelated DUD-E target settings. Percentages are relative to the initial KRAS set. No residue-specific interaction rule was used for these targets.}
    \label{tab:kras-unrelated-targets}
    \resizebox{\columnwidth}{!}{%
        \begin{tabular}{lccccc}
            \toprule
            Stage & AMPC & CXCR4 & INHA & KITH & PUR2 \\
            \midrule
            Docking & 625 (99.36\%) & 628 (99.84\%) & 77 (12.24\%) & 233 (37.04\%) & 375 (59.62\%) \\
            3D filters & 170 (27.03\%) & 326 (51.83\%) & 1 (0.16\%) & 0 (0.0\%) & 189 (30.05\%) \\
        \botrule
    \end{tabular}
    }
\end{table}

Although protein-ligand contacts can be detected automatically~\citep{salentin2015plip,bouysset2021prolif}, defining a biologically meaningful mandatory interaction is target-specific and generally requires comparison across multiple experimentally characterized complexes, recurrent interaction pattern analysis, and, where available, support from structure activity or mutagenesis data~\citep{totrov2008flexible,amaro2018ensemble}. We therefore did not introduce ad hoc residue-specific interaction constraints for the additional targets solely for the purpose of this validation. Their 3D evaluation used the pose- and geometry-based HEDGEHOG criteria.

\end{appendices}

\end{document}